\documentclass[10pt,twocolumn,letterpaper]{article}

\usepackage{cvpr}
\usepackage{times}
\usepackage{epsfig}
\usepackage{graphicx}
\usepackage{amsmath}
\usepackage{amssymb}
\usepackage{epstopdf}

\usepackage[pagebackref=false,breaklinks=true,letterpaper=true,colorlinks=true,bookmarks=false]{hyperref}

\cvprfinalcopy 

\def\cvprPaperID{1750} 

\ifcvprfinal\fi
\graphicspath{{figures/}} 
\begin{document}

\title{UntrimmedNets for Weakly Supervised Action Recognition and Detection}

\author{Limin Wang$^{1}$ \quad \quad Yuanjun Xiong$^{2}$ \quad \quad Dahua Lin$^{2}$ \quad \quad Luc Van Gool$^1$ \\
\small $^{1}$Computer Vision Laboratory, ETH Zurich, Switzerland \\
\small $^{2}$Department of Information Engineering, The Chinese University of Hong Kong, Hong Kong \\
}

\maketitle

\begin{abstract}
Current action recognition methods heavily rely on trimmed videos for model training. However, it is expensive and time-consuming to acquire a large-scale trimmed video dataset. This paper presents a new weakly supervised architecture, called {\em UntrimmedNet}, which is able to directly learn action recognition models from untrimmed videos without the requirement of temporal annotations of action instances. Our UntrimmedNet couples two important components, the {\em classification module} and the {\em selection module}, to learn the action models and reason about the temporal duration of action instances, respectively. These two components are implemented with feed-forward networks, and UntrimmedNet is therefore an end-to-end trainable architecture. We exploit the learned models for action recognition (WSR) and detection (WSD) on the untrimmed video datasets of THUMOS14 and ActivityNet. Although our UntrimmedNet only employs weak supervision, our method achieves performance superior or comparable to that of those strongly supervised approaches on these two datasets.~\footnote{The code and models are available at \url{https://github.com/wanglimin/UntrimmedNet}.}
\end{abstract}

\section{Introduction}
\begin{figure}
\includegraphics[width=\linewidth]{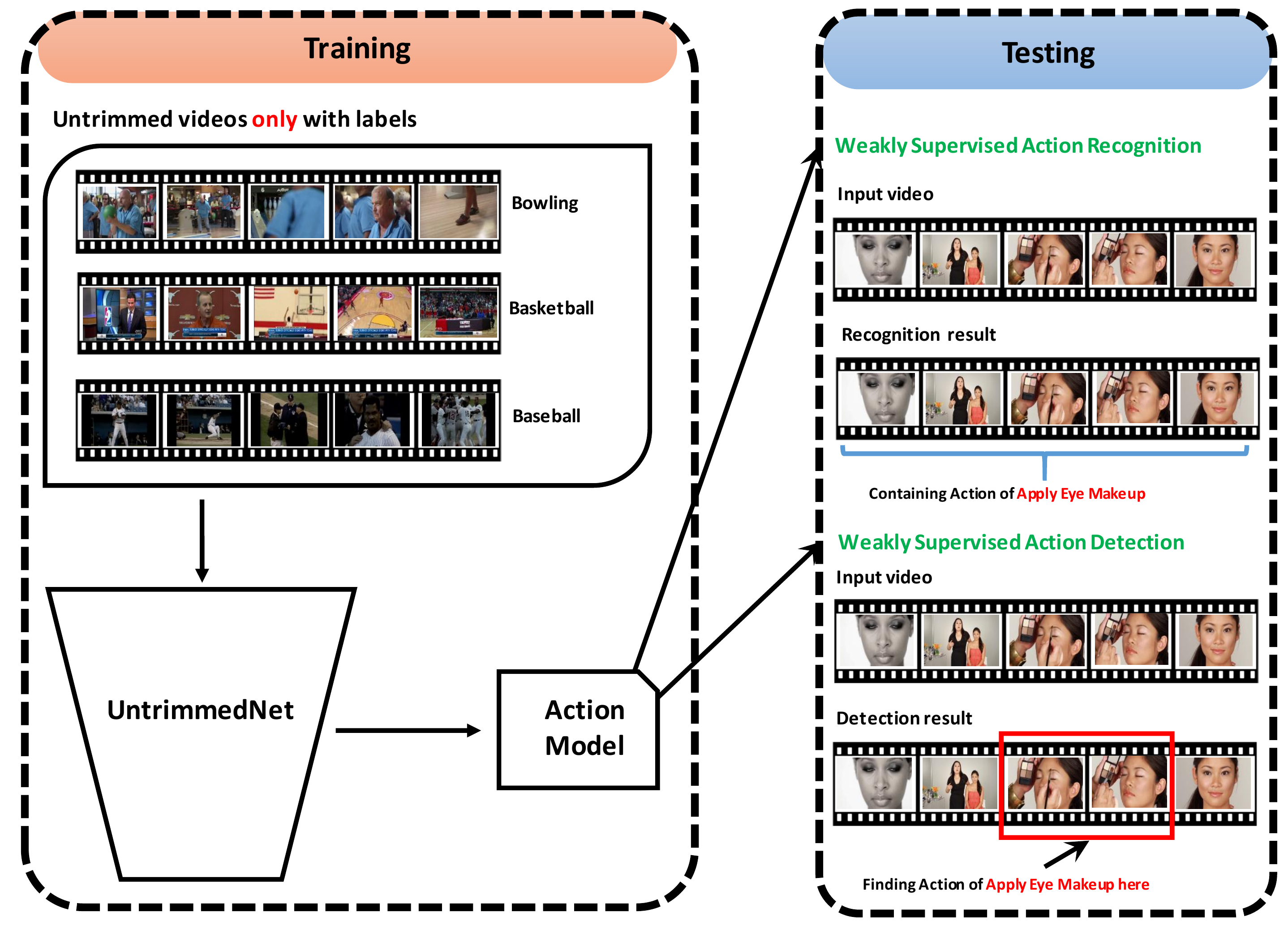}
\caption{{\bf Weakly supervised action recognition and detection}: during training phase, we simply have untrimmed videos without temporal annotation and we train action models from these untrimmed videos directly; during test phase, the learned action models could be applied to action recognition (WSR) and detection (WSD) in untrimmed videos.}
\label{fig:problem}
\end{figure}

Action recognition in videos has attracted extensive research attention in the past few years, and much progress has been made in computer vision community, on both aspects of hand-crafted representations~\cite{Laptev05,WangS13a,WangQT13,WangQT16} and deeply-learned representations~\cite{KarpathyTSLSF14,SimonyanZ14,TranBFTP15,TSN2016ECCV}. In general, action recognition is usually cast as a classification problem, where each action instance is manually trimmed from a long video sequence during the training phase, and the learned action model is exploited for action recognition in trimmed clips (e.g., HMDB51~\cite{KuehneJGPS11} and UCF101~\cite{Soomro12}) or untrimmed videos (e.g., THUMOS14~\cite{THUMOS14} and ActivityNet~\cite{HeilbronEGN15}). Although these precise temporal annotations could relieve the difficulty of learning action models, it may be difficult to adapt to large-scale action recognition in more realistic and challenging scenario due to several reasons. First, annotating temporal duration for each action instance is expensive and time-consuming. Meanwhile, huge numbers of videos on Youtube website are temporally untrimmed by nature, and trimming videos in such scale would be impractical. More importantly, unlike object boundary, there might even be no sensible definition about the exact temporal extent of actions~\cite{SatkinH10,SchindlerG08}. Thus, these temporal annotations may be subjective and not consistent across different persons. 

To overcome the above limitations of using trimmed videos for training, we introduce a more efficient setting of directly learning action recognition models from untrimmed videos, as shown in Figure~\ref{fig:problem}. In this new setting, only the video-level action label is available during training, and the goal is to learn the models from untrimmed videos, which could be applied to new videos to perform action recognition or detection. As we do not have precise temporal annotations of action instances in training, we call this new problem as {\em weakly supervised action recognition (WSR) and detection (WSD)}. 
Without the requirement of exact temporal annotations of action instances, the setup of WSR and WSD would greatly reduce the human efforts in building large-scale datasets. 
However, this weakly supervised setting also poses new challenges in that our learning algorithm needs to not only learn the visual patterns for each action class, but also automatically reason the temporal locations of possible action instances. Therefore, to deal with the problems of WSR and WSD, the designed method should consider these two aspects at the same time.

In this work, we address the challenges of the WSR and WSD problems by proposing a new end-to-end architecture, called {\em UntrimmedNet}. Without temporal annotations of action instances, our UntrimmedNet directly takes an untrimmed video as input and simply exploits its video-level label to learn the network weights. Considering the requirements mentioned above, in a nutshell, our UntrimmedNet is mainly composed of two components, namely a {\em classification module} and a {\em selection module}, which handle the problems of learning action models and detecting action instances, respectively. The outputs of the classification and selection modules are fused to yield the prediction results of untrimmed videos, which can be exploited to tune the UntrimmedNet parameters in an end-to-end manner.

Specifically, our UntrimmedNet starts with generating clip proposals, which may contain action instances, by using uniform or shot based sampling. Then, these clip proposals are fed into UntrimmedNet for feature extraction. Based on these clip-level representations, the classification module aims to predict the classification scores for each clip proposal, while the selection module tries to select or rank those clip proposals. In practice, the design of classification module is based on a standard Softmax classifier and the selection module is implemented with two alternative mechanisms: {\em hard selection} and {\em soft selection}. For hard selection, a top-$k$ pooling method is utilized to determine the most $k$ discriminative clips, and for soft selection, an attention weight is learned to rank the importance of different clips. Finally, the results of classification and selection modules are fused with an weighted summation multiplication to produce the untrimmed video-level prediction. With this video-level prediction and the global video label, we are able to jointly optimize the components of classification modules, selection modules, and feature extraction networks using the standard back propagation algorithm.

We perform experiments on two challenging untrimmed video datasets, namely THUMOS14~\cite{THUMOS14} and AcitivityNet~\cite{HeilbronEGN15}, to examine the UntrimmedNet on the tasks of weakly supervised action recognition (WSR) and detection (WSD). Although our UntrimmedNet does not employ the temporal annotations of action instances, it obtains superior performance for action recognition and comparable performance for action detection, when compared with the state-of-the-art methods that use strong supervision for training.

\section{Related Work}

{\bf Deep learning for action recognition.}
Since the breakthrough~\cite{KrizhevskySH12} in image classification with Convolutional Neural Networks (CNNs)~\cite{lecun-98} at ILSVRC 2012~\cite{RussakovskyDSKS15}, several works have been trying to design effective deep network  architectures for action recognition in videos~\cite{KarpathyTSLSF14,SimonyanZ14,TranBFTP15,FeichtenhoferPZ16,TSN2016ECCV,Wang0T15}. Karpathy \emph{et al.}~\cite{KarpathyTSLSF14} first tested deep networks on a large-scale dataset (Sports-1M) and achieved lower performance than traditional features~\cite{WangS13a}. Simonyan \emph{et al.}~\cite{SimonyanZ14} designed two stream CNNs containing spatial and temporal nets by explicitly exploiting pre-trained models and optical flow calculation. Tran \emph{et al.}~\cite{TranBFTP15} investigated 3D CNNs~\cite{JiXYY13} on the realistic and large-scale video datasets. Meanwhile, several works~\cite{Ng15,varol,DonahueJ2015,TSN2016ECCV} tried to model long-term temporal information for action understanding. Ng \emph{et al.}~\cite{Ng15} and Donahue \emph{et al.}~\cite{DonahueJ2015} utilized recurrent neural networks (LSTM) to capture the long range dynamics for action recognition. Wang \emph{et al.}~\cite{TSN2016ECCV} designed a sparse sampling strategy to model the entire video information with average aggregation. In addition, several deep learning methods have been proposed for action proposal generation and detection~\cite{GkioxariM15,Wang0TG16,MaSS16,EscorciaHNG16,GemertJGS15,ShouWC16}. Our UntrimmedNets differ to those deep networks in that the UntrimmedNets take the untrimmed videos as inputs and only require weak supervision to guide model training, while those previous architectures all uses the trimmed clips for training.

\begin{figure*}
\includegraphics[width=\textwidth]{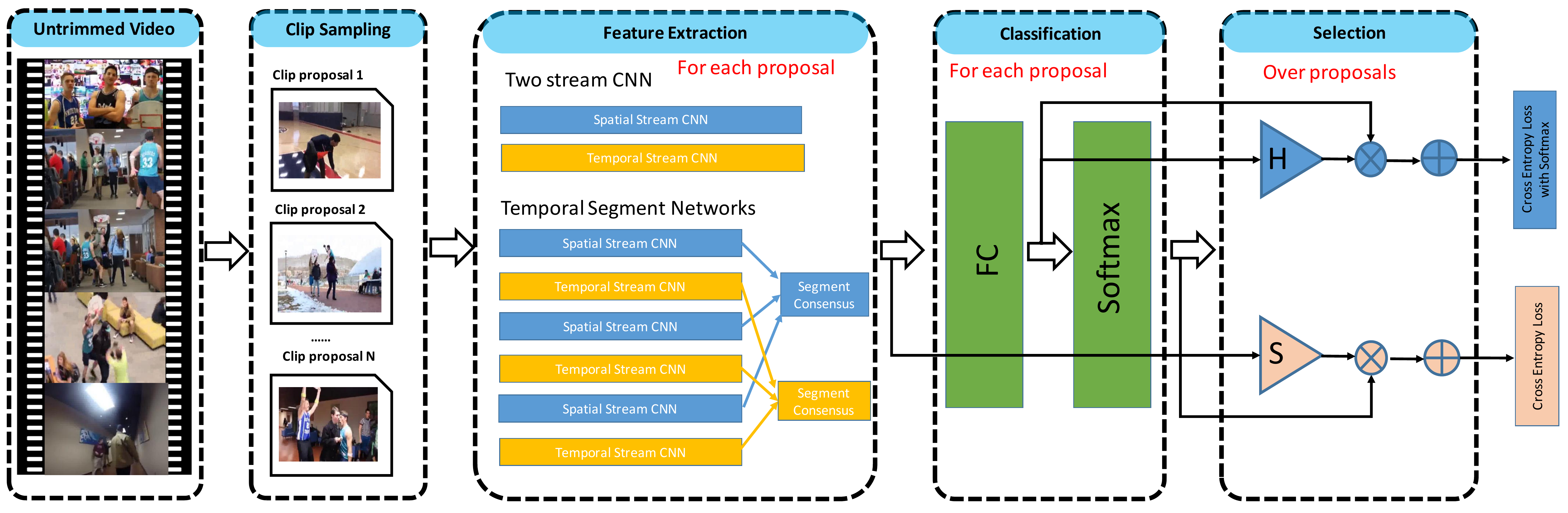}
\caption{{\bf Pipeline of learning from untrimmed videos}: our UntrimmedNets start with clip proposal generation, where we sample a set of short clips from the continuous untrimmed videos. Then, these clip proposals are separately fed into pre-trained networks for feature extraction. After this, a classification module is designed to perform action recognition for each clip proposal independently, and a selection module is proposed to detect or rank important clip proposals. Finally, the outputs of classification module and selection module are combined to yield the video-level prediction.}
\label{fig:pipeline}
\end{figure*}

{\bf Weakly supervised learning in videos.}
Weakly supervised learning was extensively studied in object recognition and detection~\cite{BilenV16,CinbisVS17,DurandTC16,OquabBLS15}, and there were several works adapting this method to learn action models from videos~\cite{LaptevMSR08,DuchenneLSBP09,BojanowskiBLPSS13,BojanowskiLBLPSS14,HuangFN16,KuehneRG16,gan2016webly,gan2016you}. The {\bf first} type of weak supervision is movie script, which provides uncertain temporal annotations of action instances. For example, Laptev \emph{et al.}~\cite{LaptevMSR08} proposed to learn action models from movie scripts for action recognition, and Duchennel \emph{et al.}~\cite{DuchenneLSBP09} tried to localize action instances in movies with the help of scripts. Compared with our work, movie script supervision show two differences: (1) movie scripts are usually aligned with frames and so they can provide approximate temporal annotations of instance, while our weak supervision does not provide any temporal information about action instances, (2) movie script supervision only applies to movie videos while our method applies to all kinds of videos. The {\bf second} type of weak supervision is a ordered list of action classes occurring in the videos. For instance, Bojanowski \emph{et al.}~\cite{BojanowskiLBLPSS14} proposed a discriminative clustering method for weakly supervised action labeling, and Huang \emph{et al.}~\cite{HuangFN16} adapted the framework of Connectionist Temporal Classification~\cite{GravesFGS06} from speech recognition to weakly supervised action labeling. Our UntrimmedNet differs from them in that our weak supervision contains no any order information on the containing action instances.

\section{Learning from Untrimmed Videos}
In this section we introduce the pipeline of learning from untrimmed videos. First, we describe the methods of generating clip proposals for UntrimmedNets. Second, we give a detailed description on the architecture design of UntrimmedNet. Finally, we present the learning algorithm to tune the parameters of UntrimmedNet in an end-to-end manner.

\subsection{Clip sampling}
\label{sec:sampling}

An action instance usually describes the continuous and coherent motion pattern with a specific intention, which may last for a few seconds and contain no shot changes. However, an untrimmed video often exhibits extremely complex motion dynamics, action instances may only occupy small portions of it. Therefore, our UntrimmedNet starts with generating short clips from the untrimmed videos, which could serve as action proposals for UntrimmedNet training.

Formally, given an untrimmed video $V$ with the duration of $T$ frames, our method generates a set of clip proposals $C = \{c_i\}_{i=1}^N$, where $N$ is the number of proposals and $c_i = (b_i, e_i)$ denote the beginning and ending location of the $i^{th}$ proposal $c_i$. We design two simple yet effective methods to generate proposals: {\em uniform sampling} and {\em shot-based sampling}.

{\bf Uniform sampling.} Under the assumption that an action instance may have a relatively short duration, we propose to divide the long video into $N$ clips with equal duration, i.e., $b_i = \frac{i-1}{N} T + 1 $ and $e_i = \frac{i}{N} T$. This sampling method ignores the continuous and consistent properties of action instances and is prone to generating imprecise proposals.

{\bf Shot-based sampling.} It is expected each action instance focuses on the consistent motion within a single shot. We present a sampling method based on shot change detection. Specifically, we extract the HOG features for each frame and calculate the HOG feature difference between adjacent frames. Then, we use the absolute value of this difference to measure the change of visual content and if it is larger than a threshold, a shot change would be detected. After this, in each shot, we propose to sample shot clips of fixed duration of $K$ frames in a sequential manner ($K$ set to 300 in practice), which helps to break down shots with very long durations. Suppose we have a shot denoted by $s_i = (s_i^b, s_i^e)$, where $(s^b_i, s^e_i)$ represents the beginning and ending locations of this shot, we produce proposals from this shot as $C(s_i) = \{ (s_i^b + (i-1) \times K, s_i^b + i \times K)\}_{i: s^b_i+  i \times K < s_i^e}$. Finally, we merge all these clip proposals from different shots for UntrimmedNet training.

\subsection{UntrimmedNets}

As shown in Figure~\ref{fig:pipeline}, the architecture of UntrimmedNet is composed of a feature extraction module, a classification module, and a selection module. These different components are all designed to be differentiable and render the UntrimmedNet to be trainable in an end-to-end manner.

{\bf Feature extraction module.} After proposal generation, these shot clips are fed into deep networks for feature extraction. These feature representations are utilized to describe the clip visual content and passed to the next layers for action recognition. Formally, given a video $V$ with a set of clip proposals $C=\{c_i\}_{i=1}^N$, we extract the representation as $\phi(V; c) \in \mathbb{R}^D$ for each clip proposal $c$. Our UntrimmedNet is a general framework for weakly supervised action recognition and detection, and does not depend on the choice of feature extraction network. In the experiments, we try out two architectures: Two-Stream CNN~\cite{SimonyanZ14} with deeper architecture~\cite{IoffeS15} and Temporal Segment Network~\cite{TSN2016ECCV} with the same architecture. More details will be described in Section~\ref{sec:exp}.

{\bf Classification module.} In the classification module, we aim to classify each clip proposal $c$ into the predefined action categories based on the extracted features $\phi(c)$. Suppose we have $C$ action classes, we learn a linear mapping $\mathbf{W}^c \in \mathbb{R}^{C \times D}$ to transform the feature representation $\phi(c)$ into a $C$-dimensional score vector $\mathbf{x}^c(c)$, i.e., $\mathbf{x}^c(c) = \mathbf{W}^c \phi(c)$, where $C$ is the number of action categories and $\mathbf{W}^c$ are the model parameters. This score vector can be also passed through a softmax layer as follows:
\begin{equation}
  \bar{x}^c_i (c) = \frac{\exp(x^c_i(c))}{\sum_{k=1}^C \exp(x^c_k(c))},
  \label{equ:softmax1}
\end{equation}
where $x^c_i(c)$ denotes the $i^{th}$ dimension of $\mathbf{x}^c(c)$. For clarity, we use the notation $\mathbf{x}^c(c)$ to denote the {\em original} classification score of clip proposal $c$ and $\bar{\mathbf{x}}^c(c)$ to represent the {\em softmax} classification score. There is a slight difference between those two types of classification scores. The original classification score $\bar{\mathbf{x}}^c(c)$ encodes the raw class activation and its response is able to reflect the degree of containing a specific action class. In the case of containing no action instance, its value could be very small for all classes. However, the softmax classification score $\bar{\mathbf{x}}^c(c)$ undergoes the normalization operation, turning its sum into 1. If there is an action instance in this clip, this softmax score could encode information of action class distribution. But for the case of background clips, this normalization operation may amplify noisy activations and its response may not encode the visual information correctly.

{\bf Selection module.} The selection module aims to select those clip proposals of most probably containing action instances. Here we design two kinds of selection mechanisms for this goal: {\em hard selection} based on the principle of multiple instance learning (MIL)~\cite{DietterichLL97} and {\em soft selection} based on the attention-based modeling~\cite{MnihHGK14,XuBKCCSZB15}. As we shall see in experiments, those two selection method can both well handle the problem of weakly supervised learning.

In the hard selection method, we try to identify a subset of $k$ clip proposals (instances) for each action class. Inspired by the idea of multiple instance learning, we choose top $k$ instances with the highest classification scores and then average among these selected instances. It should be noted that here we use the original classification score as its value is able to correctly reflect the likelihood of containing certain action instances.  Formally, let us use $x^s_i(c_j) = \delta(j\in S_i^k)$ to encode the selection choice for class $i$ and instance $c_j$, where $S_i^k$ is the set of indices of clip proposals with the highest $k$ classification scores for class $i$.

In the soft selection method, we want to combine the classification scores of all clip proposals and learn an importance weight to rank different clip proposals. Intuitively, these clip proposals are not all relevant to the action class and we could learn an attention weight to highlight the discriminative clip proposals and suppress the background clip proposals. Formally, for each clip proposal, we learn this attention weight based on the feature representation $\phi(c)$ with a linear transformation, i.e., $x^s(c) = \mathbf{w}^{sT} \phi(c)$, where $\mathbf{w}^s \in \mathcal{R}^D$ is the model parameter. Then the attention weights of different clip proposals are passed through a softmax layer and compared with each other as follows:
\begin{equation}
\bar{x}^s(c_i) = \frac{\exp(x^s(c_i))}{\sum_{n=1}^N \exp(x^s(c_n))},
\label{equ:softmax2}
\end{equation}
where $x^s(c)$ denotes the {\em original} selection score of clip proposal $c$ and $\bar{x}^s(c)$ is the {\em softmax} selection score. It should be noted that, in the classification module, the softmax operation (Eq. (\ref{equ:softmax1})) is applied to the classification scores of different action classes, for each clip proposal separately, while in the selection module, this operation (Eq. (\ref{equ:softmax2})) is performed across different clip proposals. In spite of sharing a similar mathematical formulation, these two softmax layers are designed for the purpose of classification and selection, respectively.

{\bf Video prediction.} Finally, we are able to produce the prediction score $\bar{\mathbf{x}}^p(V) $ for the untrimmed video $V$ by combining the classification and selection scores. Specifically, for hard selection, we simply average the selected top-$k$ instances as follows:
\begin{equation}
\begin{split}
x^p_i(V) & = \sum_{n=1}^N x^s_i(c_n) x^c_i(c_n), \\ 
\bar{x}^p_i(V) & = \frac{\exp(x^r_i(V))}{\sum_{k=1}^C \exp(x^r_k(V))}, \\
\end{split}
\end{equation}
where $x^s(c_n)$ and $x^c(c_n)$ are the hard selection indicator and classification score for clip proposal $c_n$, respectively. As our hard selection module is based on the original classification score, we need to perform a softmax operation to normalize the aggregated video-level score.

In the case of soft selection, as we have learned an attention weight to rank those clip proposals, we simply employ a weighted summation to combine the scores of the classification and selection modules, as follows:
\begin{equation}\label{eq:weighted_combine}
\bar{\mathbf{x}}^p(V) = \sum_{n=1}^N \bar{x}^s(c_n) \bar{\mathbf{x}}^c(c_n).
\end{equation}
Here, different from hard selection, we use the softmax classification score for each clip proposal, as this normalized score would make attention weight learning easier and more stable. Note that Eq.~(\ref{eq:weighted_combine}) forms a convex combination of probability vectors. Hence no further normalization is required.

\subsection{Training}

After the introduction of UntrimmedNet architecture in the previous subsection, we turn to discuss how to optimize the model parameters. The components of feature extraction, classification module, and selection module are implemented with feed-forward neural networks that are all differentiable with model parameters. Therefore, following training methods of strongly supervised architecture (e.g., Two-Stream CNNs), we employ the standard back propagation method with cross-entropy loss:
\begin{equation}
\ell(\mathbf{w}) = \sum_{i=1}^M \sum_{k=1}^C y_{ik} \log \bar{x}^p_k(V_i),
\end{equation}
where $y_{ik}$ is set to 1 if video $V_i$ contains action instances of $k^{th}$ category, and to 0 otherwise, $M$ is the number of training videos. A weight decay rate of 0.0005 is enforced during the training. In the case of video containing action instances from multiple classes, we first normalize the label vector $\mathbf{y}$ with its $\ell_1$-norm~\cite{WeiXHNDZY14}, i.e. $\bar{\mathbf{y}} = \mathbf{y} / \|\mathbf{y}\|_1$, and then use this normalized label vector $\bar{\mathbf{y}}$ to calculate cross-entropy loss.

\section{Action Recognition and Detection}
\label{sec:rd}

Having introduced UntrimmedNet for directly learning from untrimmed videos, we now turn to describing how to exploit these learned models for action recognition and detection in untrimmed videos. 

{\bf Action recognition.} As our UntrimmedNets are built on the two stream CNNs~\cite{SimonyanZ14} or temporal segment networks~\cite{TSN2016ECCV}, the learned models can be viewed as snippet-level classifiers. Following the recognition pipeline of previous methods~\cite{SimonyanZ14,TSN2016ECCV,xiong2016cuhk}, we perform snippet-wise evaluation for action recognition in untrimmed videos. In practice, we sample a single frame (or 5 frame stacking of optical flow) every 30 frames. The recognition scores of sampled frames are aggregated with top-$k$ pooling (k set to 20) or weighted sum to yield the final video-level prediction.

{\bf Action detection.} Our UntrimmedNet with soft selection module not only delivers a recognition score, but also outputs an attention weight for each snippet. Naturally, this attention weight could be exploited for action detection (temporal localization) in untrimmed videos. For more precise localization, we perform test every 15 frames and keep the prediction score and attention weight for each frame. Based on the attention weight, we remove background  by thresholding (set to 0.0001) on it. Finally, after removing background, we produce the final detection results by thresholding (set to 0.5) on the classification score.

\section{Experiments}
\label{sec:exp}

In this section we describe the experimental results of our method. First, we introduce the evaluation datasets and the implementation details of our UntrimmedNets. Then, we perform exploration studies to determine important configurations of our approach. Finally, we examine our method on weakly supervised action recognition (WSR) and action detection (WSD), and compare with the state-of-the-art methods.

\subsection{Datasets}
We evaluate our UntrimmedNet on two large datasets, namely THUMOS14~\cite{THUMOS14} and ActivityNet~\cite{HeilbronEGN15}. These two datasets are suitable to evaluate our method as they provide the original untrimmed videos. It should be noted that these two datasets also have temporal annotations of action instances for training data, but {\bf we do not use these temporal annotations when training our UntrimmedNets}. 

The {\bf THUMOS14} dataset has 101 classes for action recognition and 20 classes for action detection. It is composed of four parts: training data, validation data, testing data, and background data. To verify the effectiveness of our UntrimmedNet on learning from untrimmed videos, we mainly use the validation data (1,010 videos) to train our models and the test data (1,574 videos) to evaluate their performance. The {\bf ActivityNet} dataset is a recently introduced benchmark for action recognition and detection in untrimmed videos. We use the ActivityNet release 1.2 for our experiments. In this release, the ActivityNet consists of 4,819 videos for training, 2,383 videos for validation, and 2,480 videos for testing, of 100 activity classes. We perform two kinds of experiments: 1) learning UntrimmedNets on the training data and testing it on the validation data, 2) learning UntrimmedNets on the combination of training and validation data and submitting testing results to the evaluation server. The {\bf evaluation metric} is based on mean average precision (mAP) for action recognition on these two datasets. For action detection, we follow the standard evaluation metric by reporting mAP values for different intersection over union (IoU) values on the dataset of THUMOS14.

\subsection{Implementation details}

We use the video extension version~\cite{TSN2016ECCV} of the Caffe toolbox~\cite{JiaSDKLGGD14} to implement the UntrimmedNet. We choose two successful deep architectures for feature extraction in our UntrimmedNet, namely Two Stream CNNs~\cite{SimonyanZ14} and Temporal Segment Network~\cite{TSN2016ECCV}. The two networks are both based on two stream inputs (RGB and Optical Flow) and Temporal Segment Network is equipped with segmental modeling (3 segments) to capture long-range temporal information. Following the Temporal Segment Network, the input to the spatial stream is 1 RGB frame and the temporal stream takes 5-frame stacks of TVL1 optical flow. We choose the Inception architecture~\cite{IoffeS15} with Batch Normalization for the UntrimmedNet design and we initialize UntrimmedNet parameters of both streams with pre-trained models from ImageNet~\cite{DengDSLL009} with the method introduced in ~\cite{TSN2016ECCV}. The UntrimmedNet parameters are optimized with the mini-batch stochastic gradient algorithm, where the batch size is set to 256 and the momentum to 0.9. The initial learning rate is set to 0.001 for the spatial stream and decreases every 4,000 iterations by a factor of 10, and the whole training stops at $10,000$ iterations. For the temporal stream, we set the initial learning rate to 0.005, which is decreased every 6,000 iterations by a factor of 10, and it stops training at $18,000$ iterations. As the training set size of THUMOS14 and ActivityNet is relatively small, we use high dropout ratios (0.8 for the spatial stream and 0.7 for the temporal stream) and common data augmentation techniques including cropping augmentation and scale jittering.

\subsection{Exploration studies}

In this subsection, we focus on the exploration studies to determine the important setups of UntrimmedNet. Specifically, we perform investigation on the THUMOS14 dataset, where we train the UntrimmedNet on the validation data and conduct evaluation on the testing data. In all these experiments, we report performance of both hard selection and soft selection

\begin{figure}[t]
\centering
\includegraphics[width=0.495\linewidth]{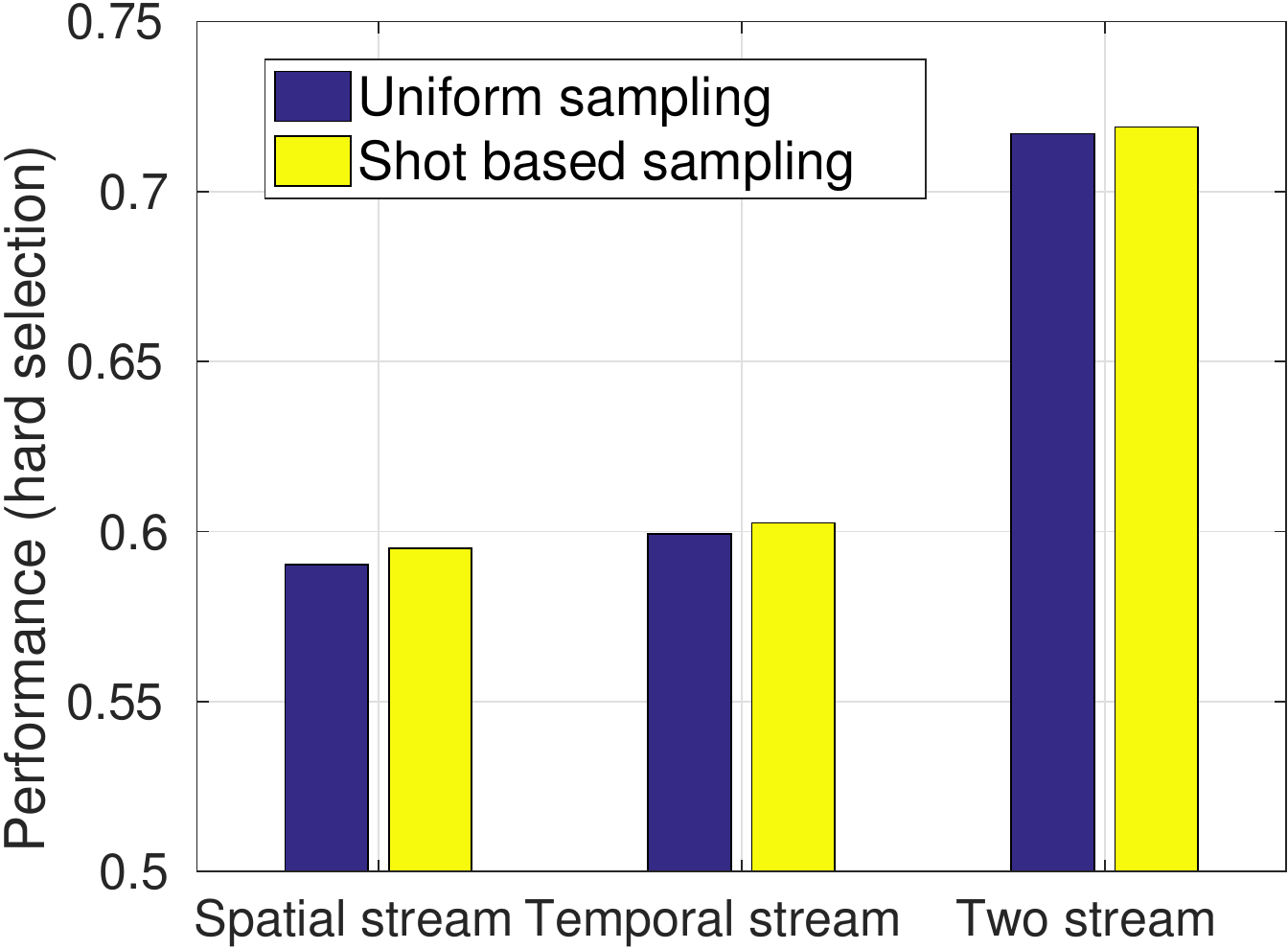}
\includegraphics[width=0.495\linewidth]{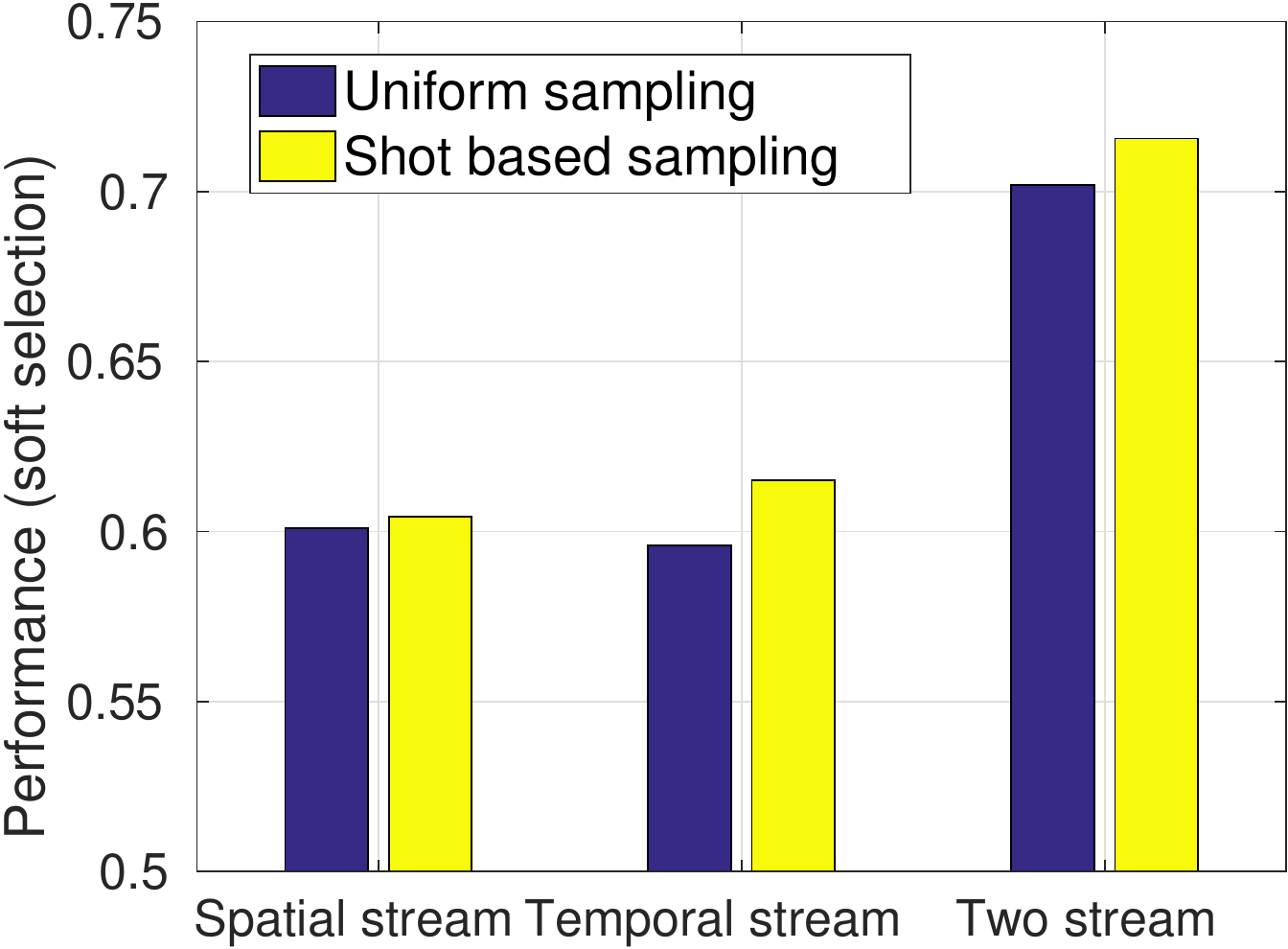}
\caption{Comparison of different clip proposal sampling methods on the THUMOS14 dataset.}
\label{fig:sampling}
\end{figure}

{\bf Clip sampling.} We design two simple sampling method in Section~\ref{sec:sampling}. We start our experiments by comparing these two proposal sampling methods. In this study, we use the two stream CNNs for feature extraction in the UntrimmedNet and seven clips are randomly sampled from each video. The numerical results are summarized in Figure~\ref{fig:sampling}. From the results, we see that both sampling methods can give good performance for UntrimmedNet training and the shot based sampling is able to yield better performance (71.6\% vs. 70.2\% for the soft selection module). We ascribe the better performance of shot based sampling to the fact that shot detection is able to automatically detect the action boundary and is more natural for video temporal segmentation than uniform segmentation. In the remaining experiments, we choose the shot based proposal sampling by default.

\begin{figure}[t]
\centering
\includegraphics[width=0.49\linewidth]{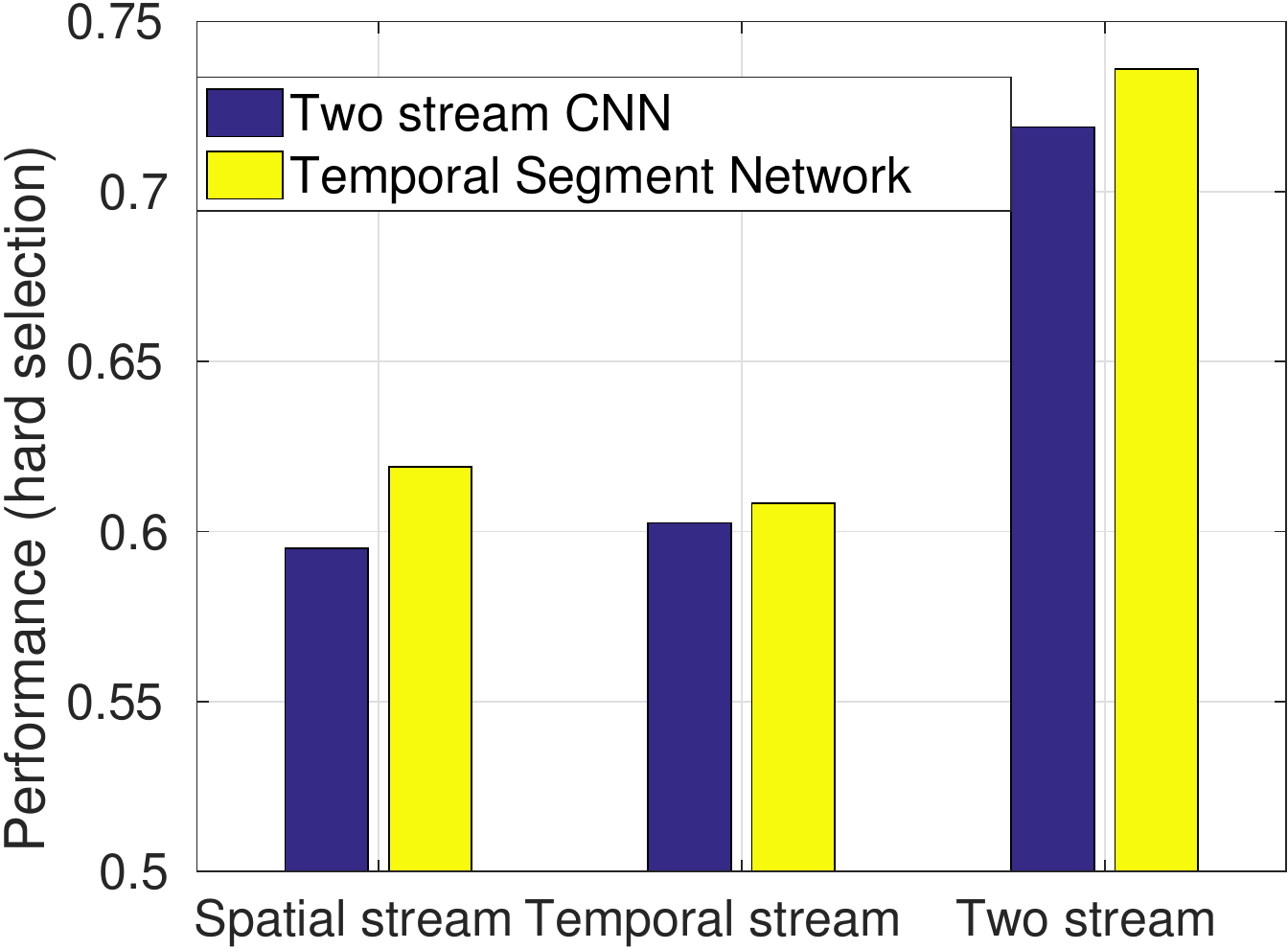}
\includegraphics[width=0.49\linewidth]{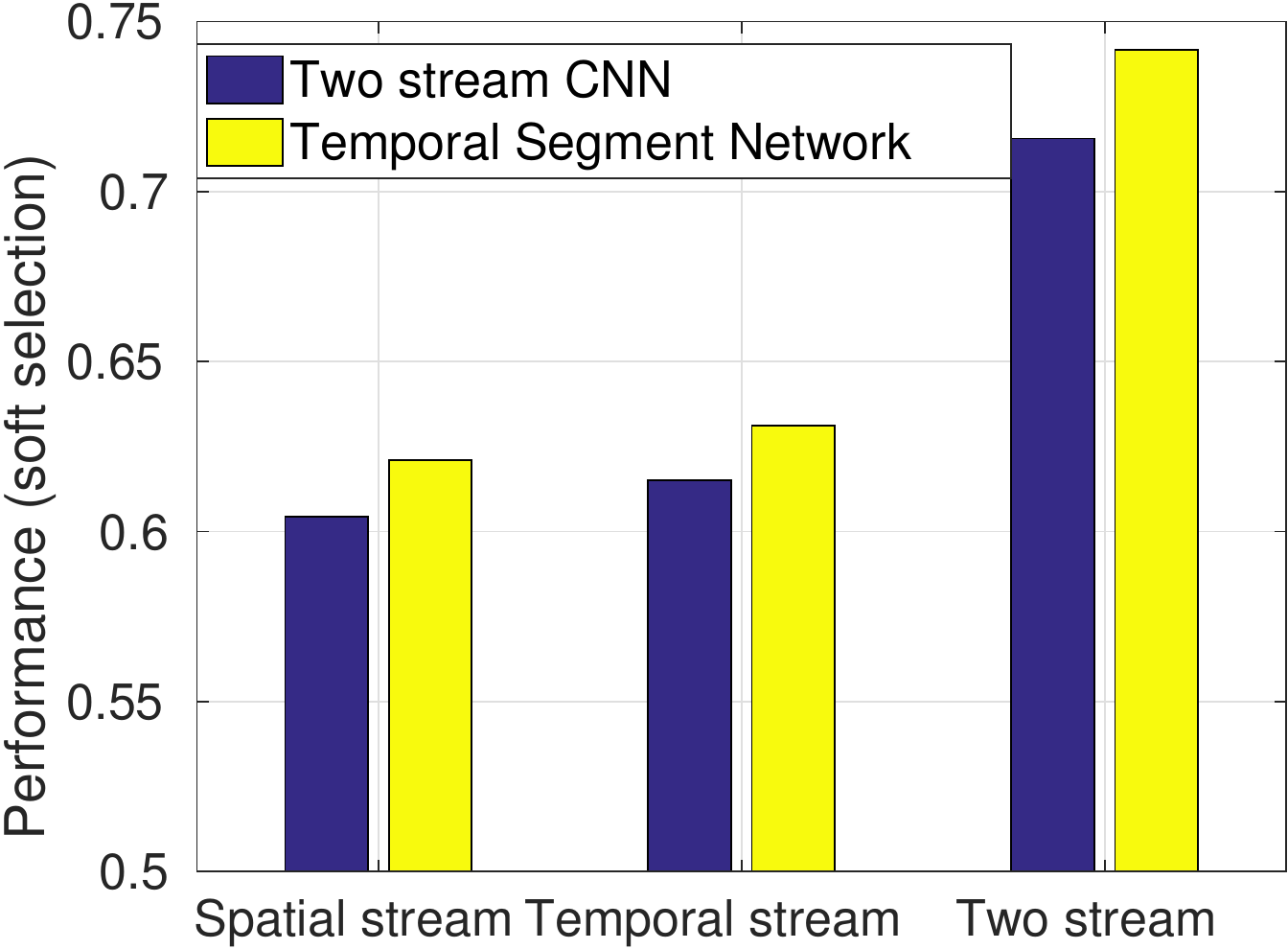}
\caption{Comparison of different architectures for feature extraction on the THUMOS14 dataset.}
\label{fig:feature}
\end{figure}

{\bf Feature extraction.} An important component in our UntrimmedNet is feature extraction as the classification and selection modules both depend on feature representations. In this experiment, we choose two networks, namely two stream CNNs~\cite{SimonyanZ14} and temporal segment networks~\cite{TSN2016ECCV}, and sample seven clip proposals per video during the training phase. The experimental results are reported in Figure~\ref{fig:feature}, and we observe that the temporal segment networks consistently outperform the original two stream CNNs for both hard and soft selection modules, due to their long-term modeling over the entire clip (74.2\% vs. 71.6\% for the soft selection module). Therefore, we choose the temporal segment networks for feature extraction in the remaining experiments.

\begin{figure}
\centering
\includegraphics[width=0.49\linewidth]{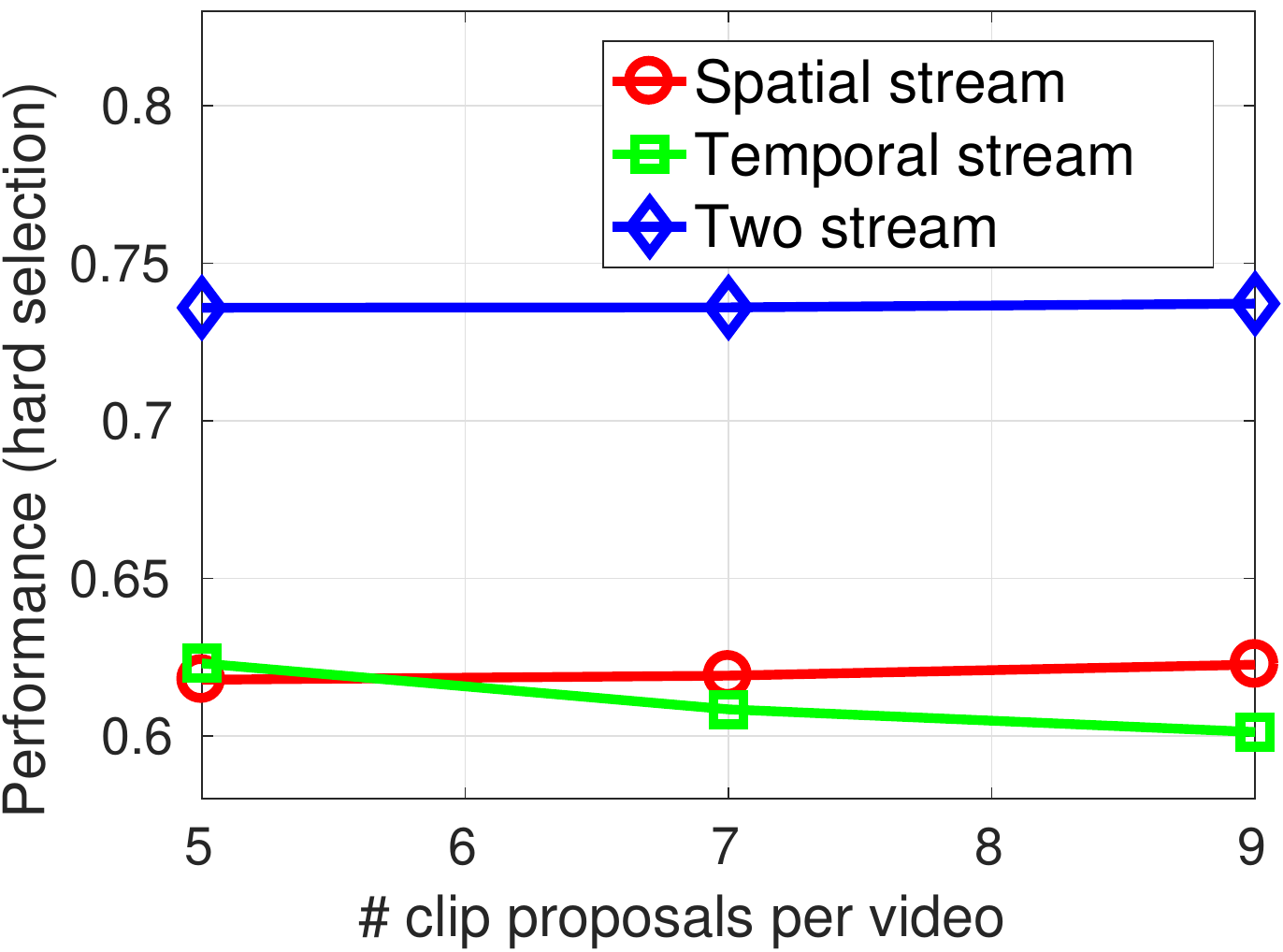}
\includegraphics[width=0.49\linewidth]{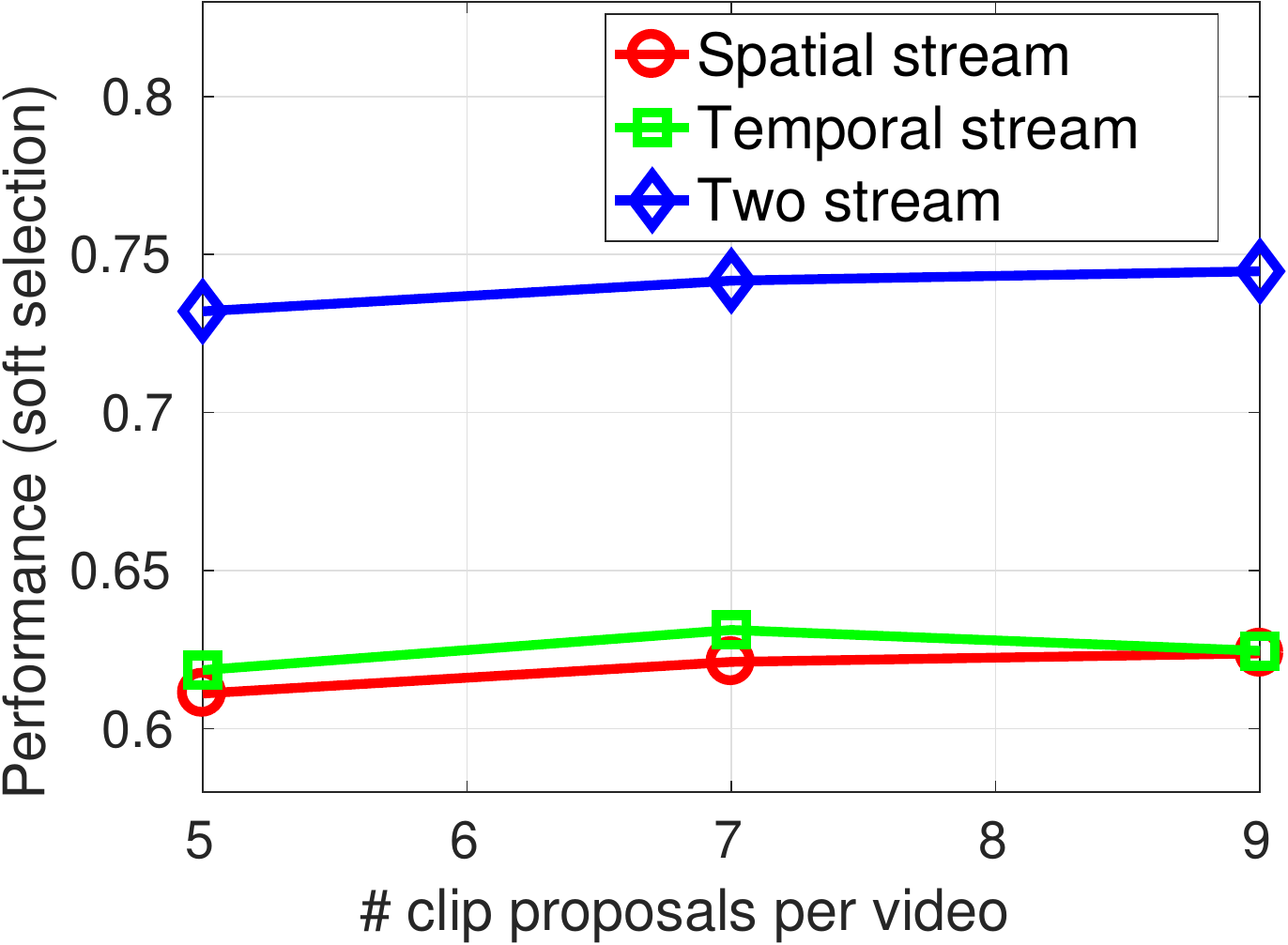}
\caption{Performance of UntrimmedNets with different numbers of clip proposal per video on the THUMOS14 dataset.}
\label{fig:proposal_num}
\end{figure}

{\bf Number of proposals.} Another important parameter in the design of UntrimmedNet is the number of clip proposals sampled from each video. As the GPU memory is limited, we need to strike a balance between the number of sampled clip proposals per video and the number of videos per batch. According to our experiment, on average, we generate 40 clip proposals for each video on the THUMOS14 dataset and 20 clip proposals for each video on the ActivityNet dataset. In our experiment, we set the number of sampled clip proposals per video to 5, 7, 9. In the hard selection module, we set the parameter $k$ in top-$k$ pooling as $\lfloor \frac{N}{2} \rfloor$, where $N$ is the number of sampled clip proposals. The experimental results are summarized in Figure~\ref{fig:proposal_num} and we see that for separate streams, the performance slightly varies when the number of sampled proposals changes, but the performance of two stream networks is quite stable for the hard selection module. For the soft selection module, the values 7 and 9 show a small advantage over 5 and therefore, to keep a balance between accuracy and efficiency, we fix the number of sampled proposal to 7 in the remaining experiments.

\subsection{Evaluation on WSR}
\begin{table}
\centering
\resizebox{\linewidth}{!}{
  \begin{tabular}{|l|c|c|c|}
    \hline
     Method & THUMOS14 & ActivityNet (a) & ActivityNet (b) \\
    \hline
    \hline
    TSN (3 seg)~\cite{TSN2016ECCV} & 67.7\% & 85.0\% & 88.5\%\\
    TSN (21 seg) &  68.5\% & 86.3\% & 90.5\% \\
    \hline
    UntrimmedNet (hard) & 73.6\% & {\bf 87.7\%} & {\bf 91.3\%} \\
    UntrimmedNet (soft)  & {\bf 74.2\%} & 86.9\% & 90.9\% \\
    \hline
  \end{tabular}
  }
  \vspace{1mm}
  \caption{Effectiveness of selection module on the problem of weakly supervised action recognition (WSR). On the THUMOS14 dataset, we train UntrimmedNet on the validation data and evaluate on the test data. For the setting (a) of ActivityNet, we train UntrimmedNet on the training videos and test on the validation videos. For the setting (b) of AcitivtyNet, we train on the train+val videos and evaluate on the test server. ``hard'' and ``soft'' in UntrimmedNet rows refer to hard and soft selection modules.}
  \label{tbl:result1}
\end{table}

\begin{table}
\centering
\resizebox{\linewidth}{!}{
\begin{tabular}{|lr|lr|}
\hline
\multicolumn{2}{|c|}{THUMOS14} & \multicolumn{2}{|c|}{ActivityNet} \\
\hline
\hline
iDT+FV~\cite{WangS13a} & 63.1\% & iDT+FV~\cite{WangS13a} & 66.5\%$^*$ \\
Two Stream~\cite{SimonyanZ14} & 66.1\% & Two Stream~\cite{SimonyanZ14} & 71.9\%$^*$  \\
EMV+RGB~\cite{ZhangWWQW16} & 61.5\% & C3D~\cite{TranBFTP15} & 74.1\%$^*$  \\
Objects+Motion~\cite{JainGS15} & 71.6\% & Depth2Action~\cite{ZhuN16} & 78.1\%$^*$  \\
TSN (3 seg)~\cite{TSN2016ECCV} & 78.5\% & TSN (3 seg)~\cite{TSN2016ECCV} & 88.8\%$^*$ \\
\hline
\hline
UntrimmedNet (hard) & 81.2\% & UntrimmedNet (hard) &   {\bf 91.3\%} \\
UntrimmedNet (soft) & {\bf 82.2\%} & UntrimmedNet (soft)  & 90.9\% \\
\hline
\end{tabular}
}
\vspace{1mm}
\caption{Comparison of our UntrimmedNet with other state-of-the-art methods on the datasets of THUMOS14 and AcitivtyNet (v1.2) for action recognition. For ActivityNet, we train the models on train+val videos and evaluate on the test server. $^*$ indicates using strong supervision for training. }
  \label{tbl:result_sota}
\end{table}

After the exploration study on different configurations, we turn to the investigation of UntrimmedNet on the problem of weakly supervised action recognition (WSR) on the datasets of THUMOS14 and ActivityNet in this subsection. 

{\bf Effectiveness of selection module.} We first examine the effectiveness of leveraging selection modules in UntrimmedNets for learning from untrimmed videos. In order to study the setting of learning from untrimmed videos, we use the validation data for training on the THUMOS14 dataset, and use the untrimmed videos without temporal annotations for training on the ActivityNet dataset.

We choose two baseline methods to compare: the standard temporal segment network with the average aggregation function (TSN), which is the state-of-the-art action recognition method, and TSN with more segments, which uses more segments during training. The numerical results are summarized in Table~\ref{tbl:result1}. From these results, we first see that our UntrimmedNet equipped with a hard or soft selection module outperforms the original TSN frameworks on both datasets. Furthermore, for the sake of a fair comparison with our UntrimmedNet, we increase the segment number of TSN to 21, which is equal to the number of segments in our UntrimmedNet ($3 \times 7$), and we see that increasing the segment numbers indeed contributes to improving the recognition performance. But the performance of TSN with 21 segments is still below that of our UntrimmedNet, which indicates that explicitly designing selection modules for learning from untrimmed videos is effective.

{\bf Comparison with the state of of the art.} After a separate study on the effectiveness of selection modules on WSR, we now compare the UntrimmedNet with other state-of-the-art methods on those two challenging datasets. To get a fair comparison with other methods, we use the training and validation videos to learn UntrimmedNets on the THUMOS14 dataset. As its training data (UCF101) is already trimmed, we simply use the whole video clips as proposals to train our UntrimmedNet. On the dataset of ActivityNet, we combine the training and validation videos to train our models and report the performance on the testing videos. It is worth noting that other methods all use strong supervision (i.e. temporal annotation and video labels), while our UntrimmedNet only uses weak supervision (i.e. only video labels)

We compare with several previous successful action recognition methods, which previously achieved the state-of-the-art performance on these two datasets, including improved trajectories (iDT+FV)~\cite{WangS13a}, two stream CNNs~\cite{SimonyanZ14}, 3D convolutional networks (C3D)~\cite{TranBFTP15}, temporal segment networks (TSN)~\cite{TSN2016ECCV}, Object+Motion~\cite{JainGS15}, and Depth2Action~\cite{ZhuN16}. The numerical results are summarized in Table~\ref{tbl:result_sota}. We see that our UntrimmedNets outperform all these previous methods. Our best performance is 3.7\% above that of other methods on the THUMOS14 dataset and 2.5\% on the ActivityNet dataset. This superior performance of UntrimmedNet justifies the importance of jointly learning classification and selection modules. Furthermore, we are only using weak supervision and have obtained better performance than those methods relying on strong supervision, which could be explained by the fact that our UntrimmedNet could well utilize useful context information in the whole untrimmed videos rather than only learning from trimmed activity clips.

\begin{table}[t]
\centering
\resizebox{1.\linewidth}{!}{
\begin{tabular}{l|ccccc}
\hline
IoU ($\alpha$) & $\alpha$= 0.5 & $\alpha$ = 0.4 & $\alpha$ = 0.3 & $\alpha$ = 0.2 & $\alpha$ = 0.1 \\
\hline
Oneata {\em et al.}~\cite{OneataVS14}$^*$  & 14.4 & 20.8 & 27.0 & 33.6 & 36.6 \\
Richard {\em et al.}~\cite{Richard_2016_CVPR}$^*$  & 15.2 & 23.2 & 30.0 & 35.7 & 39.7 \\
Shou {\em et al.}~\cite{ShouWC16}$^*$ & 19.0 & 28.7 & 36.3 & 43.5 & 47.7 \\
Yeung {\em et al.}~\cite{YeungRMF16}$^*$  & 17.1 & 26.4 & 36.0 & 44.0 & 48.9 \\
Yuan {\em et al.}~\cite{Yuan_2016_CVPR}$^*$ & 18.8 & 26.1 & 33.6 & 42.6 & 51.4 \\
\hline 
UntrimmedNet (soft) & 13.7 & 21.1 & 28.2 & 37.7 & 44.4 \\
\hline
\end{tabular}
}
\vspace{1mm}
\caption{Comparison of our UntrimmedNet with other state-of-the-art methods on the datasets of THUMOS14 for action detection. $^*$ indicates using strong supervision for training.}
\label{tbl:det2}
\end{table}

\begin{figure*}[t]
\center
 \vspace{-1.5mm}
  \includegraphics[width=0.121\linewidth]{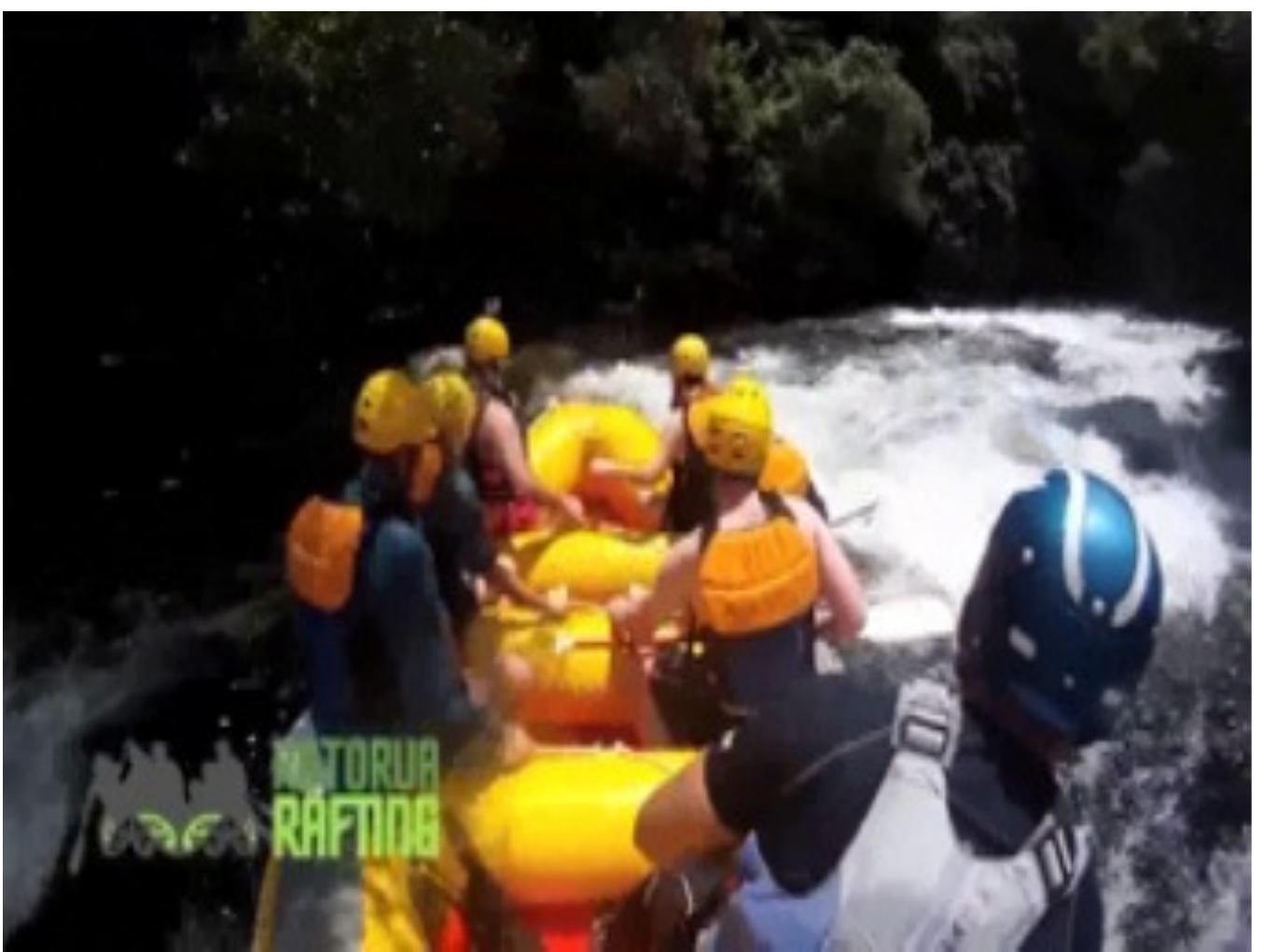}
  \hspace{-1.5mm}
  \includegraphics[width=0.121\linewidth]{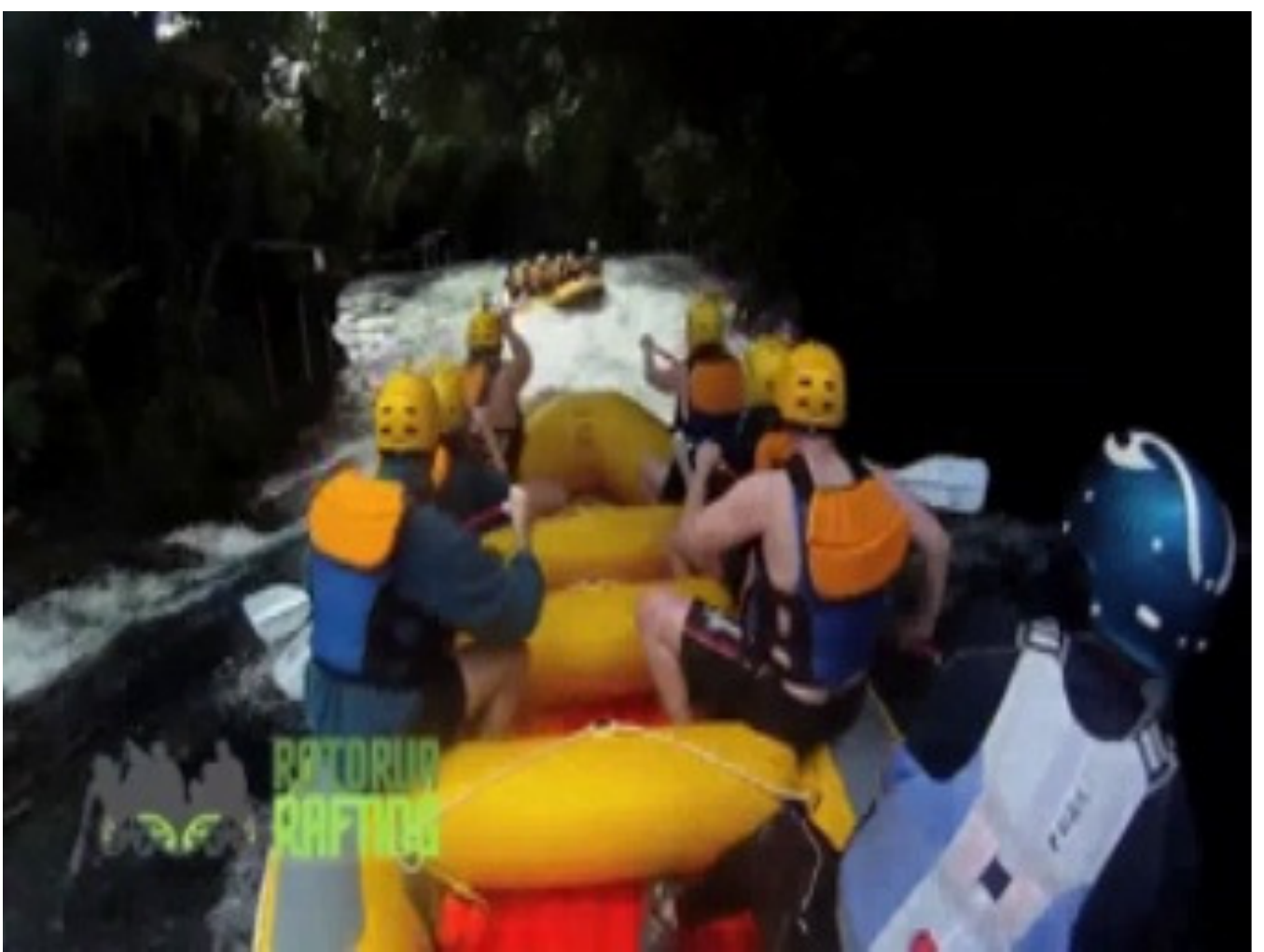}
  \hspace{-1.5mm}
  \includegraphics[width=0.121\linewidth]{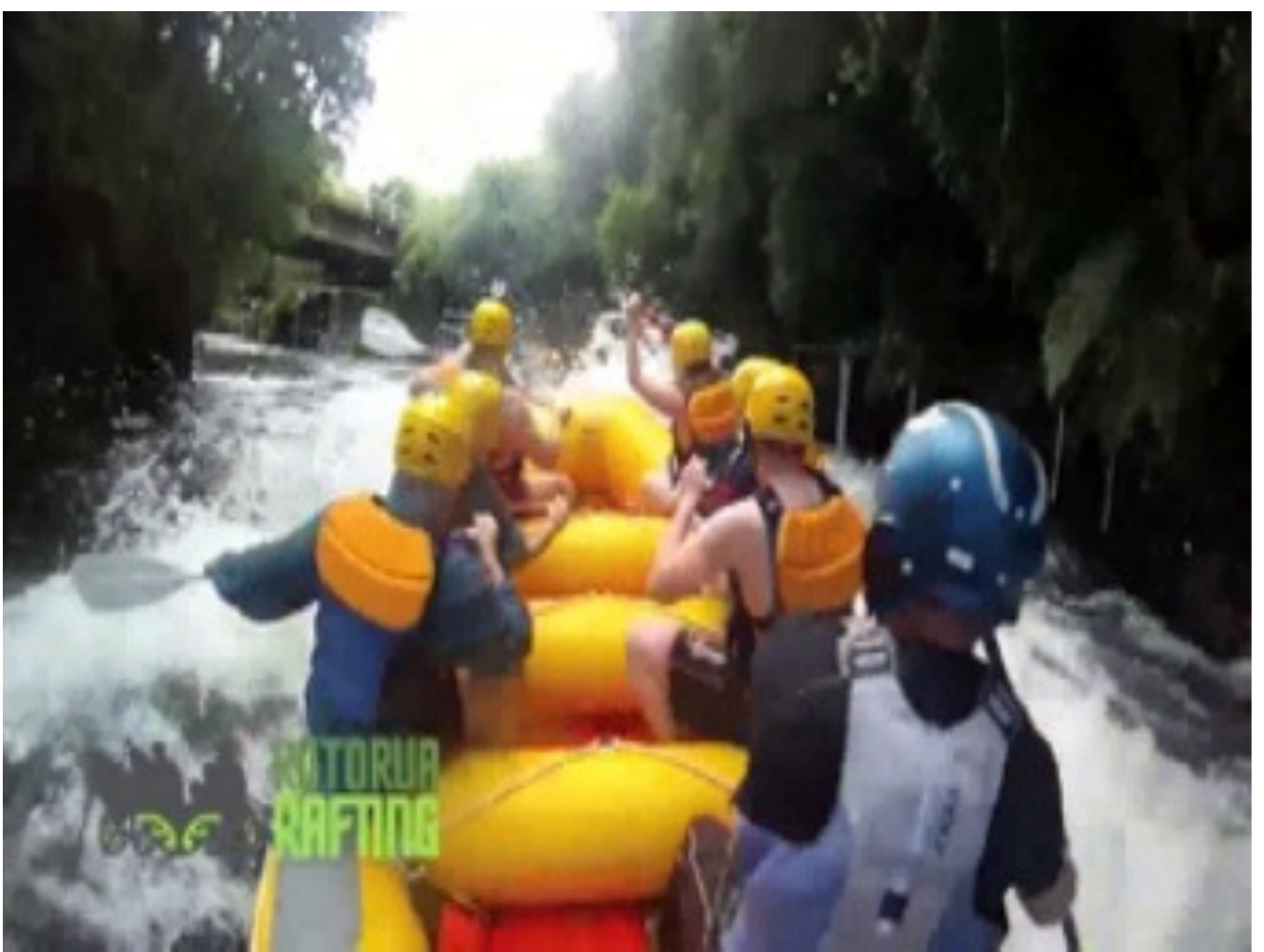}
  \hspace{-1.5mm}
  \includegraphics[width=0.121\linewidth]{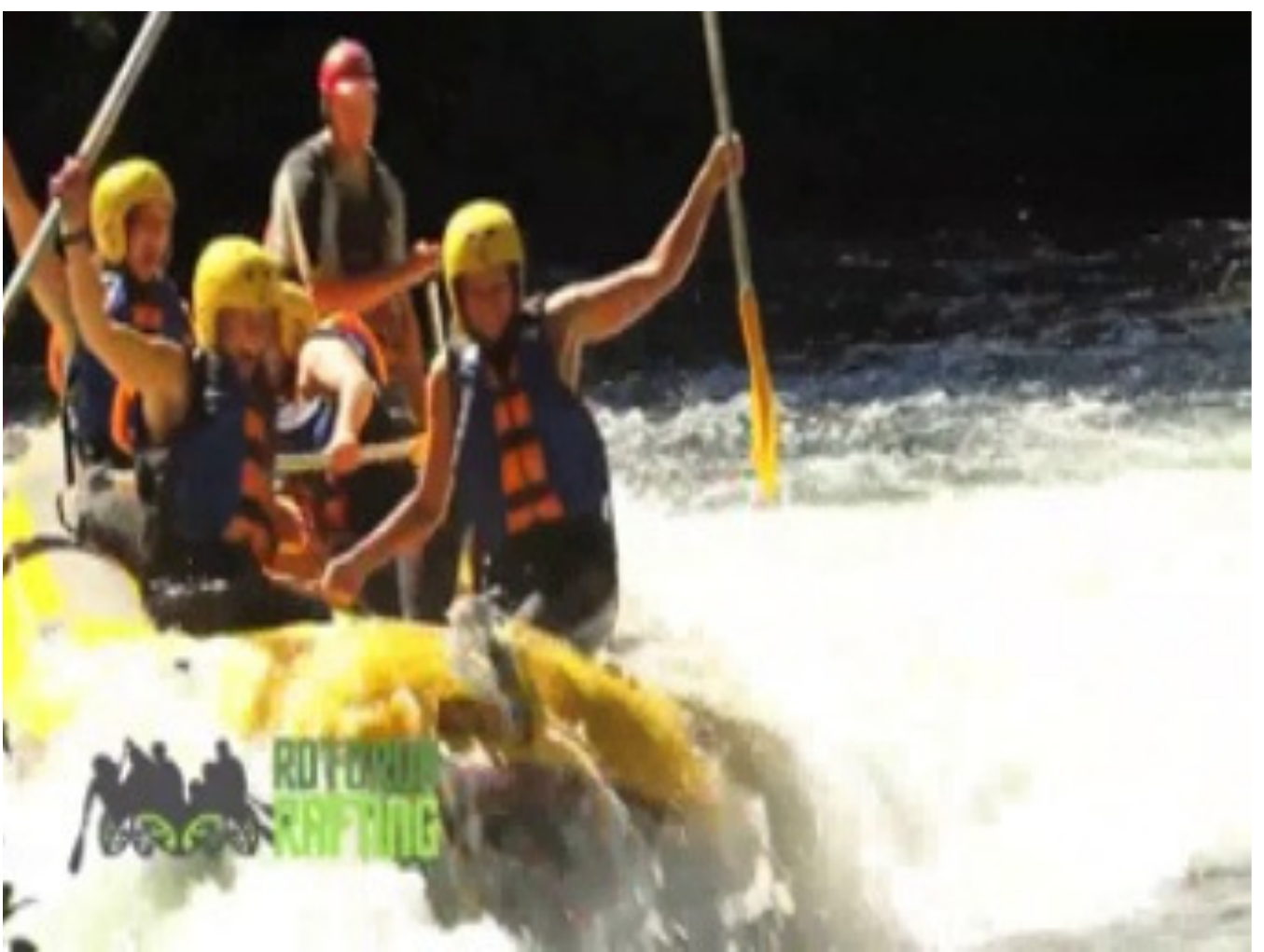}
  \hspace{-1.5mm}
  \includegraphics[width=0.121\linewidth]{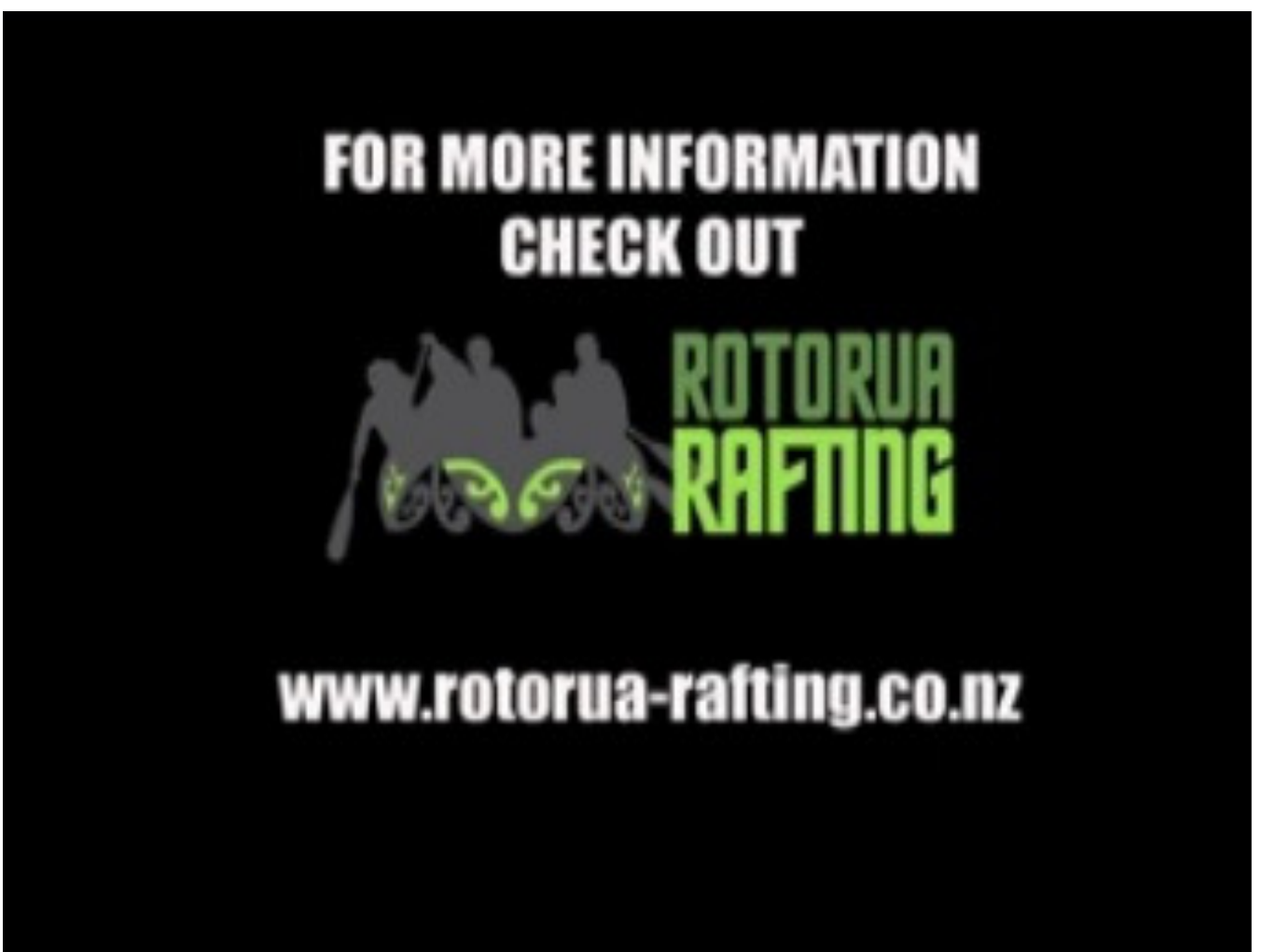}
  \hspace{-1.5mm}
  \includegraphics[width=0.121\linewidth]{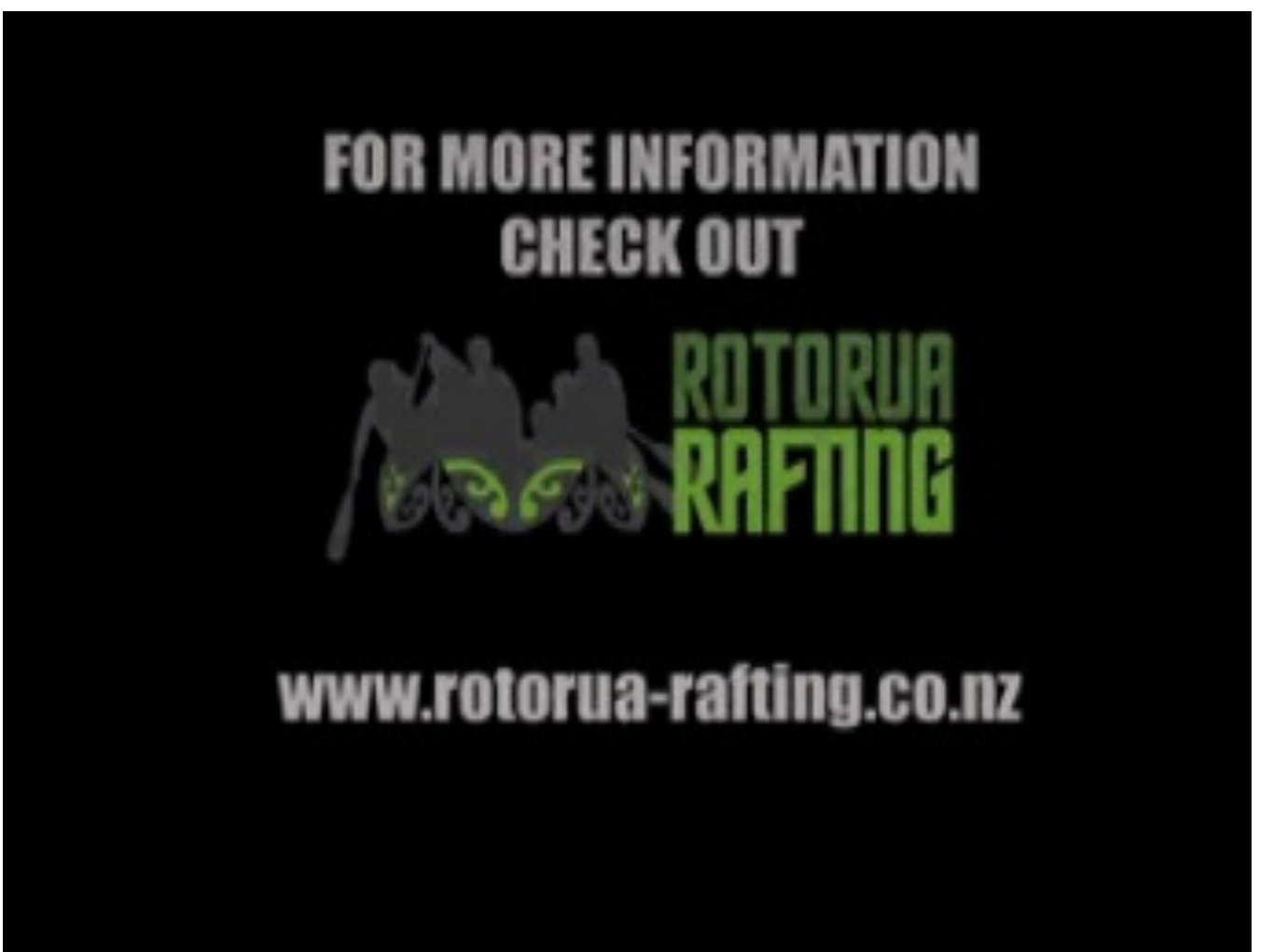}
  \hspace{-1.5mm}
  \includegraphics[width=0.121\linewidth]{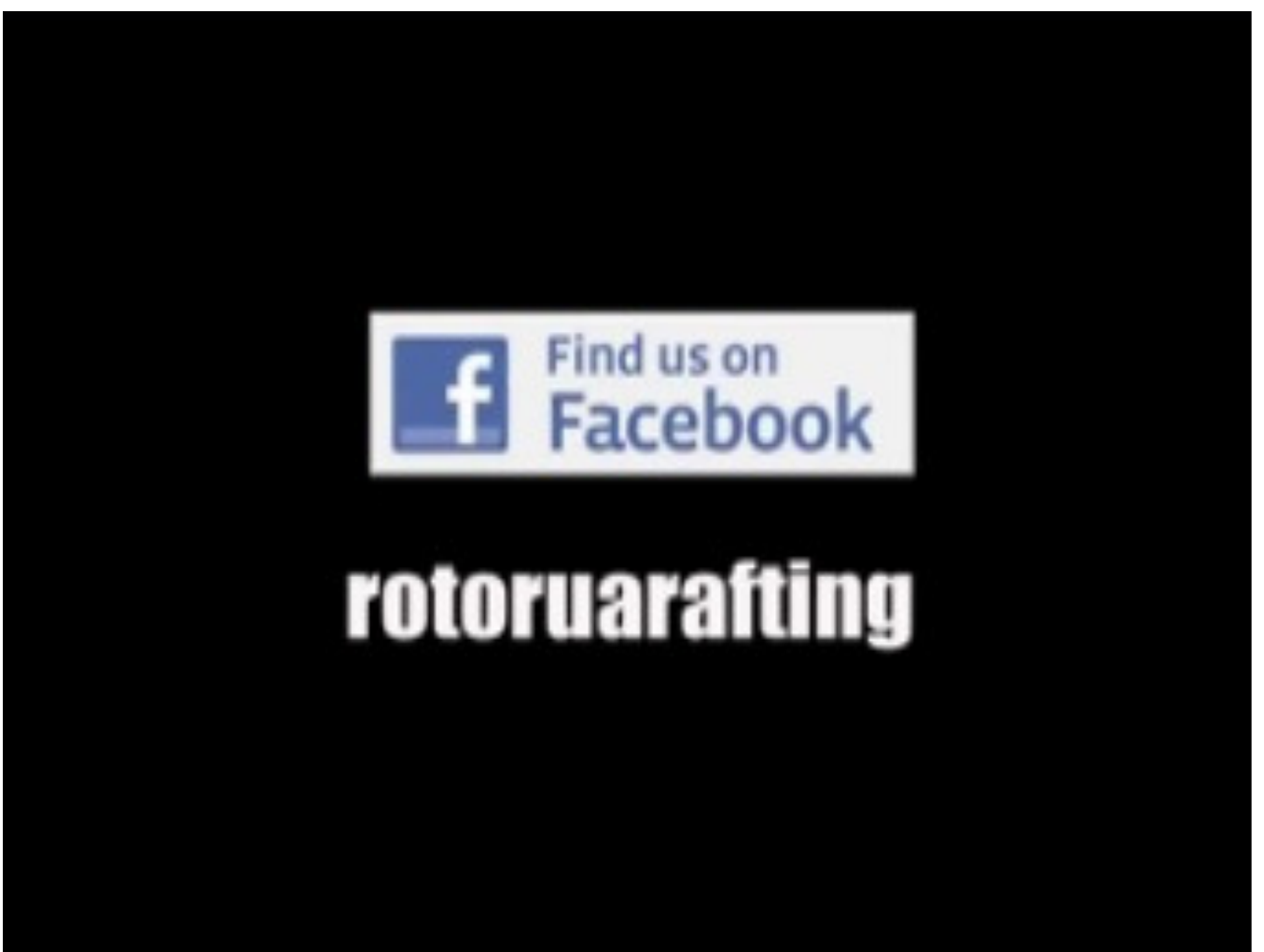}
  \hspace{-1.5mm}
  \includegraphics[width=0.121\linewidth]{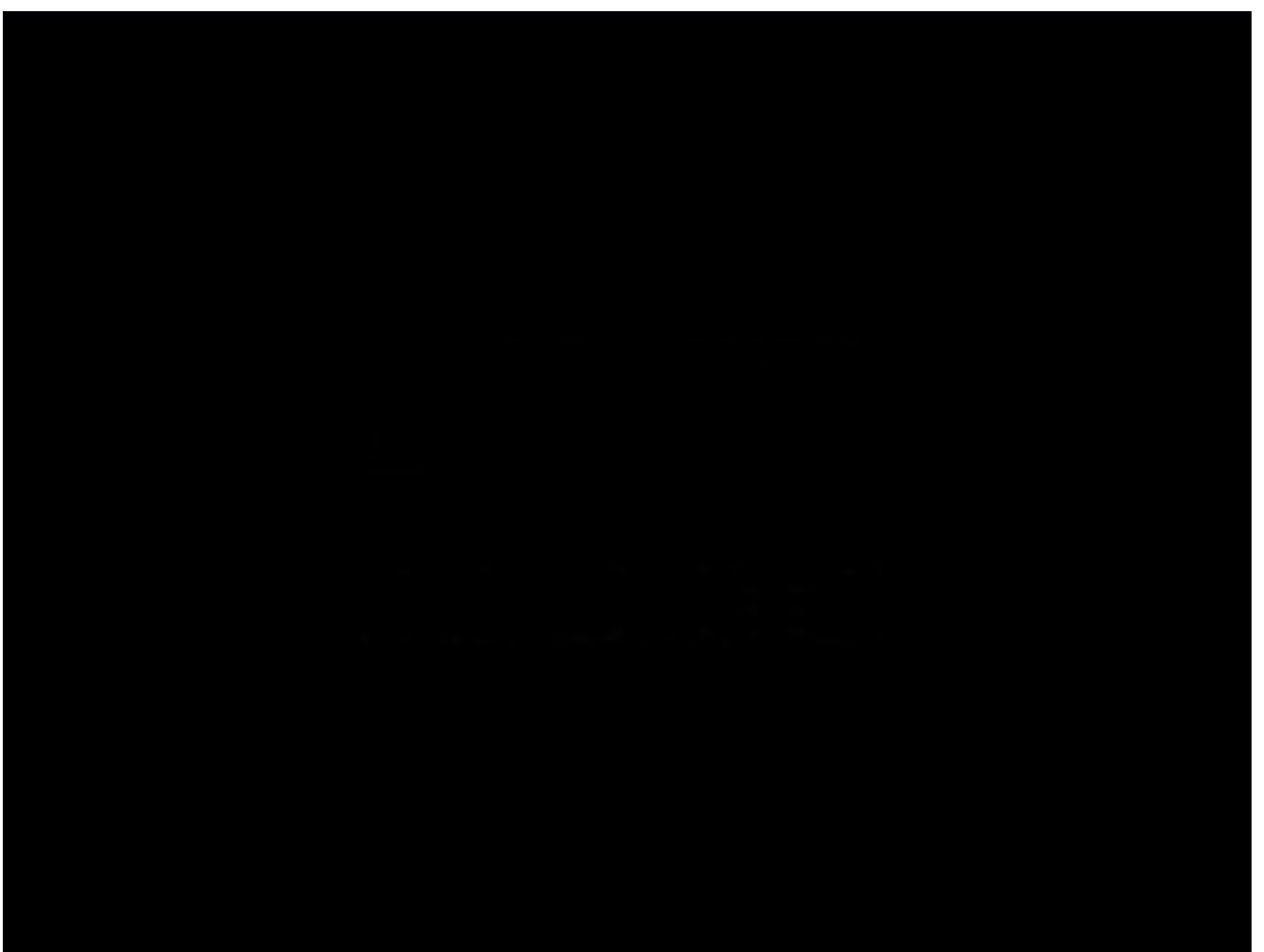}

  \includegraphics[width=0.121\linewidth]{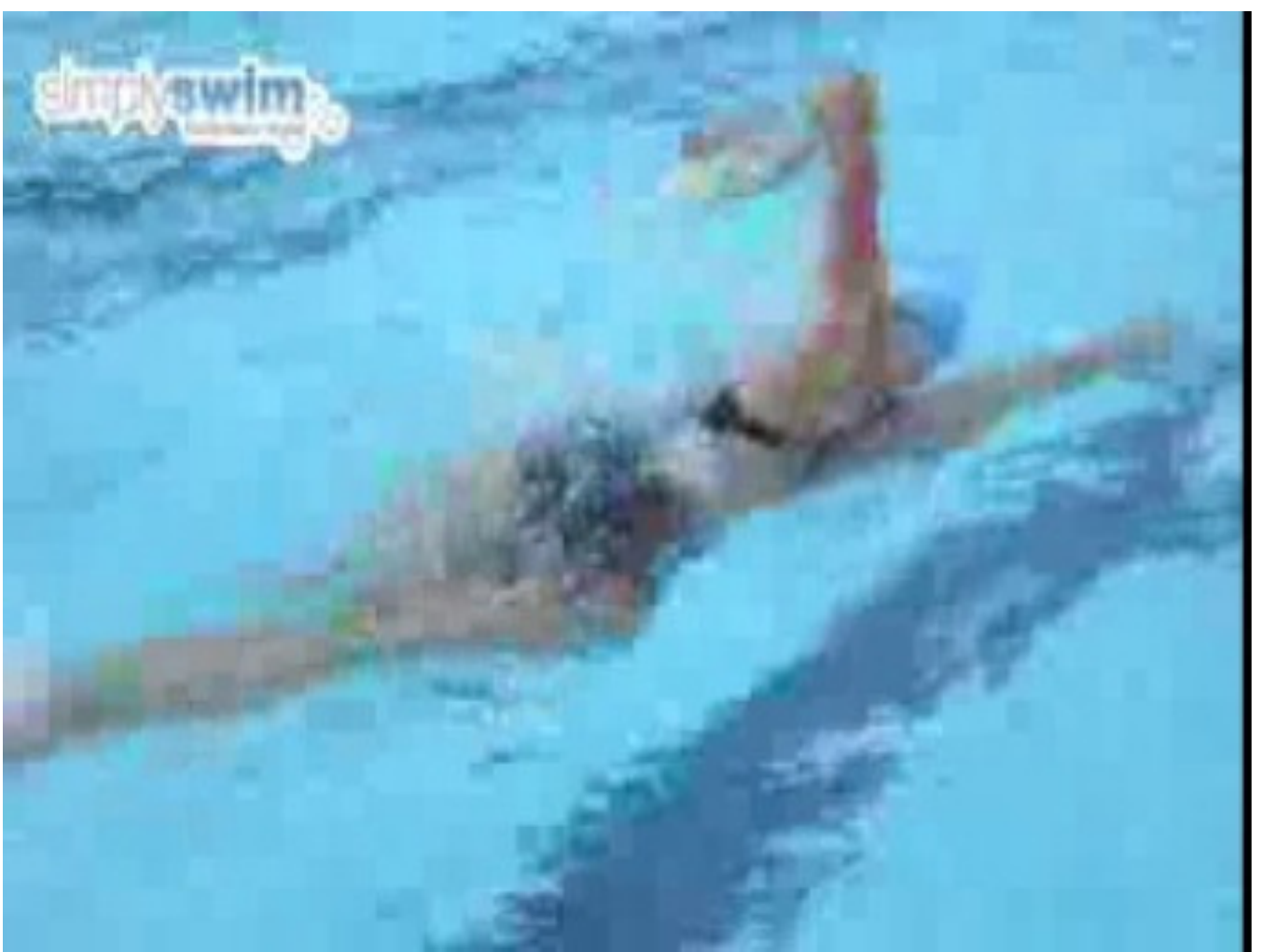}
  \hspace{-1.5mm}
  \includegraphics[width=0.121\linewidth]{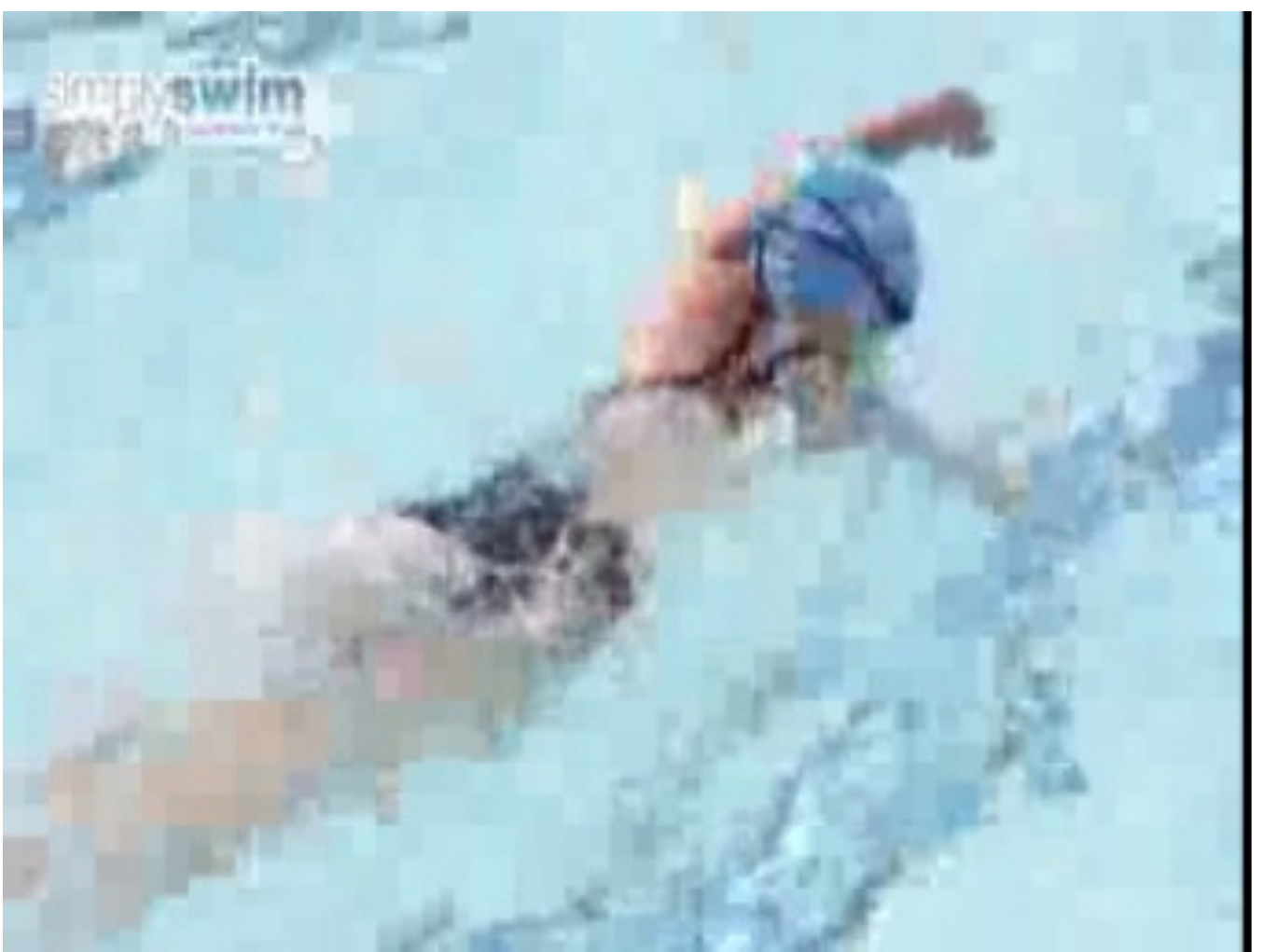}
  \hspace{-1.5mm}
  \includegraphics[width=0.121\linewidth]{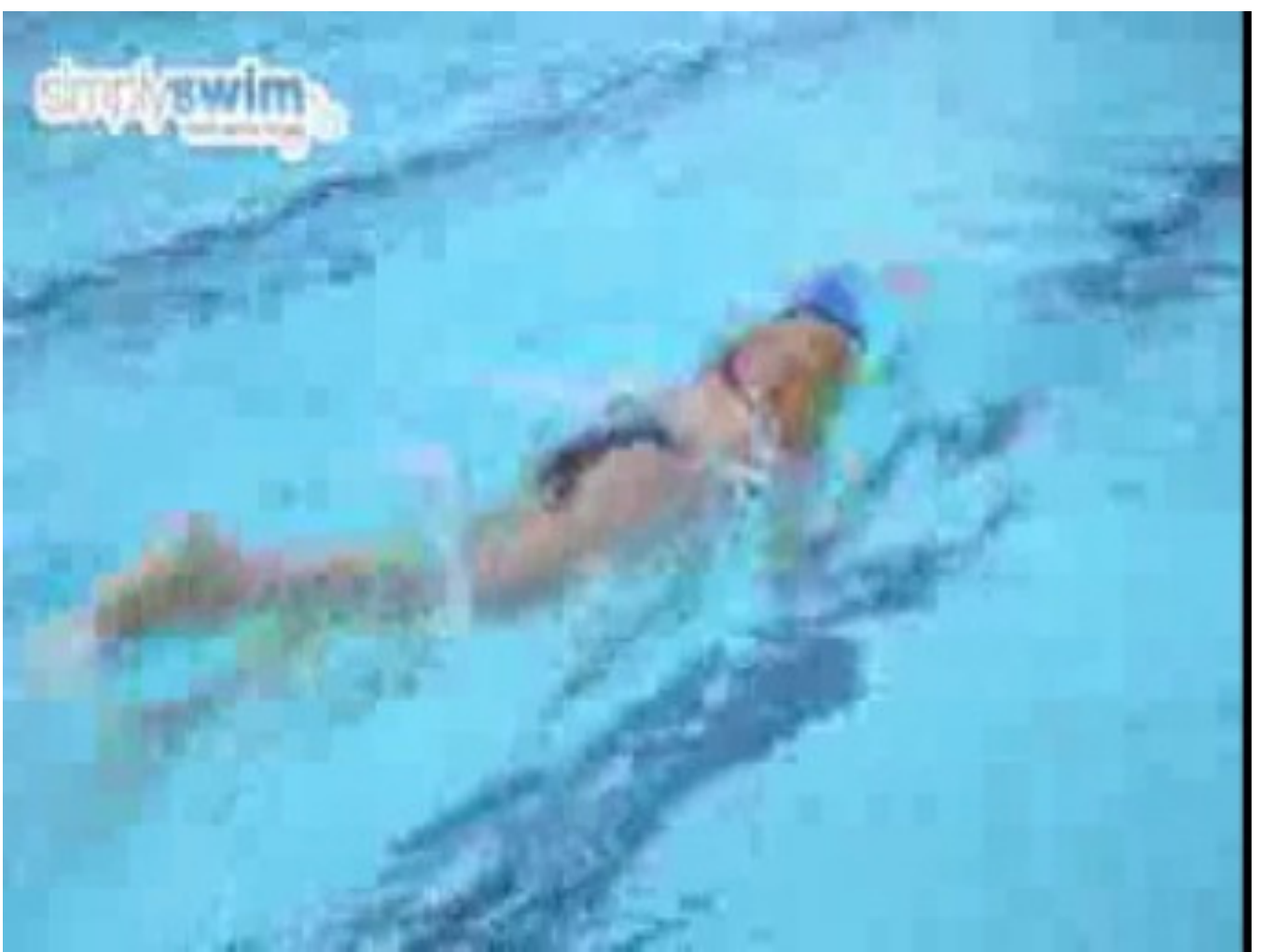}
  \hspace{-1.5mm}
  \includegraphics[width=0.121\linewidth]{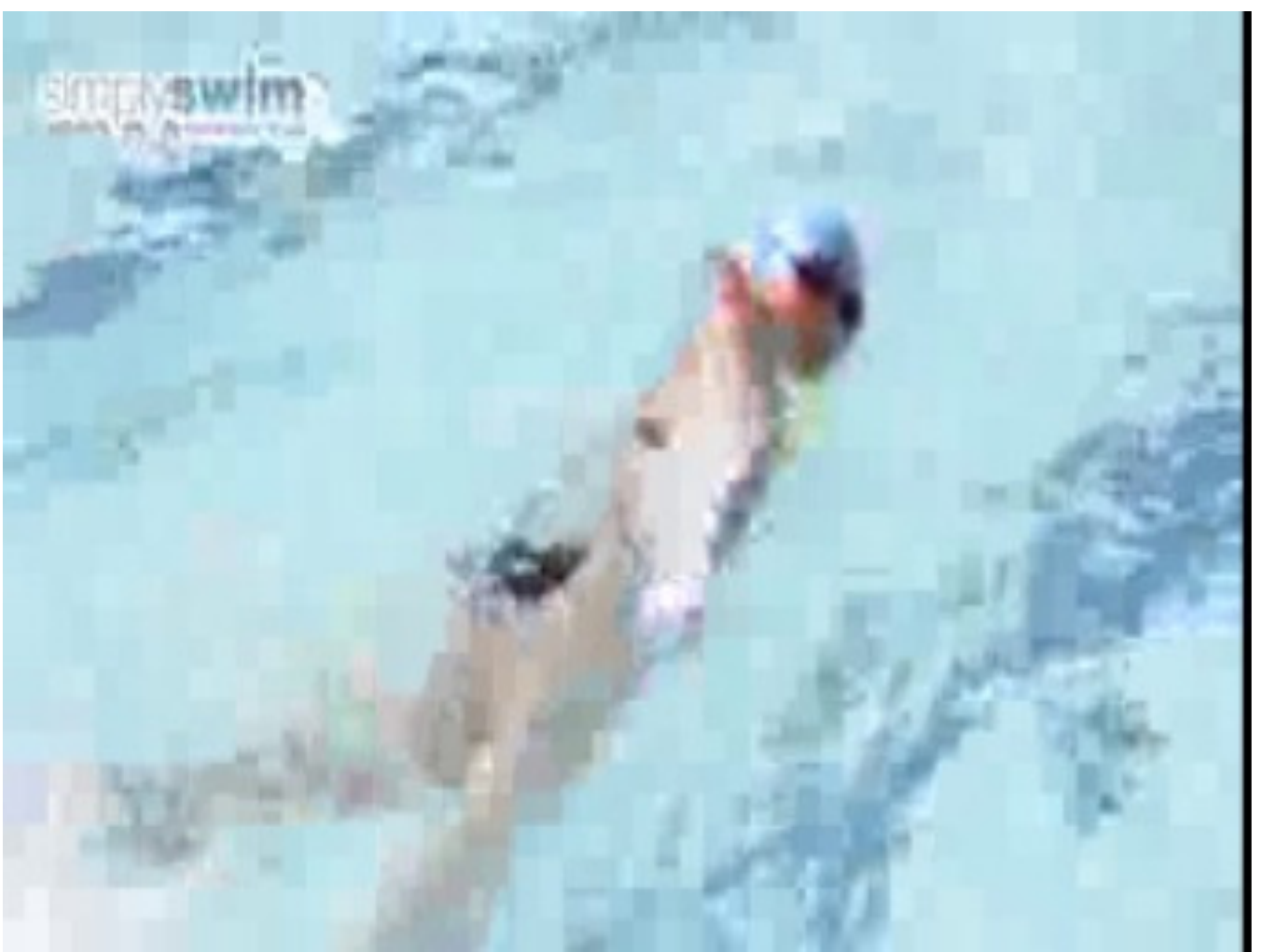}
  \hspace{-1.5mm}
  \includegraphics[width=0.121\linewidth]{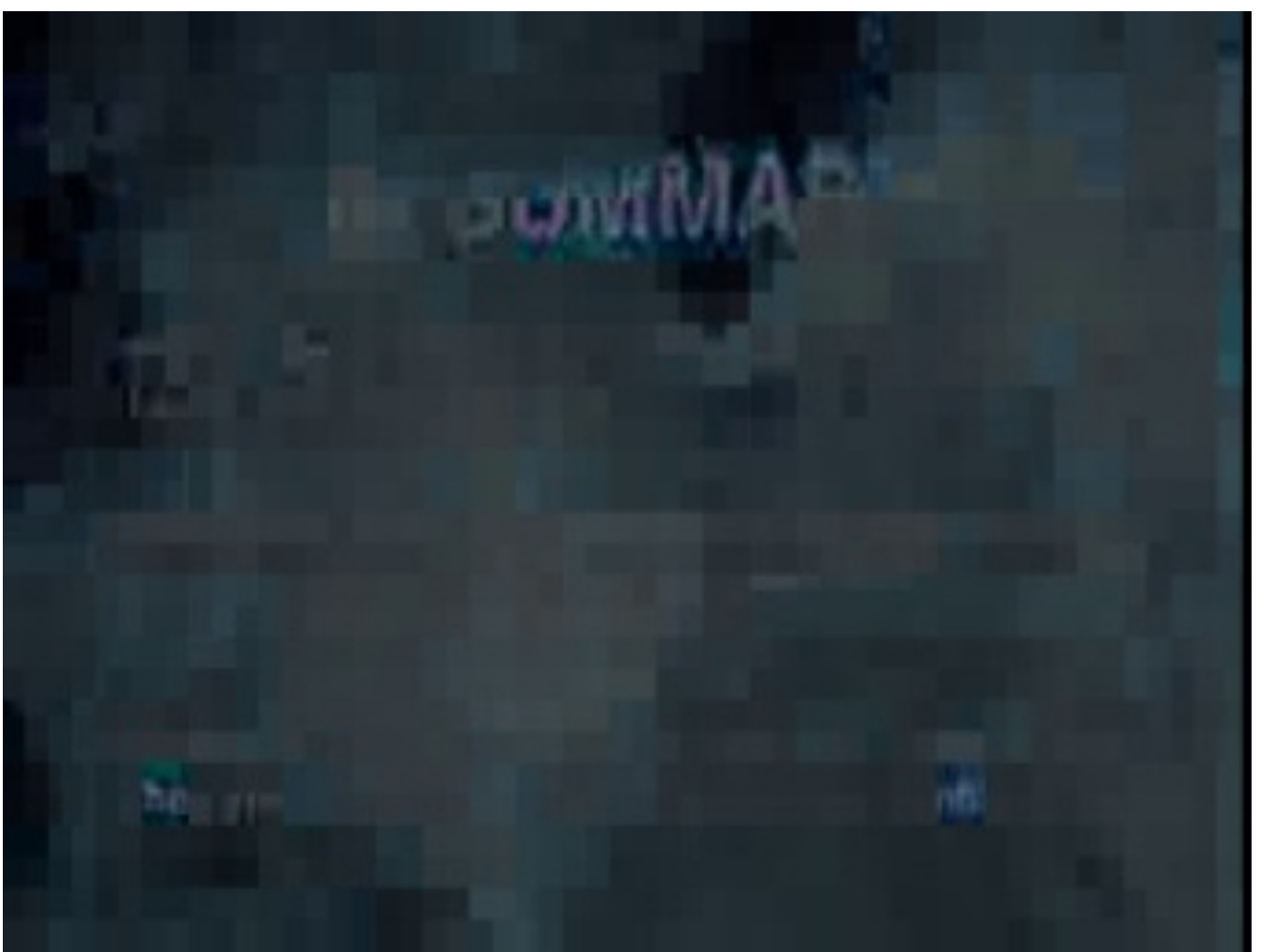}
  \hspace{-1.5mm}
  \includegraphics[width=0.121\linewidth]{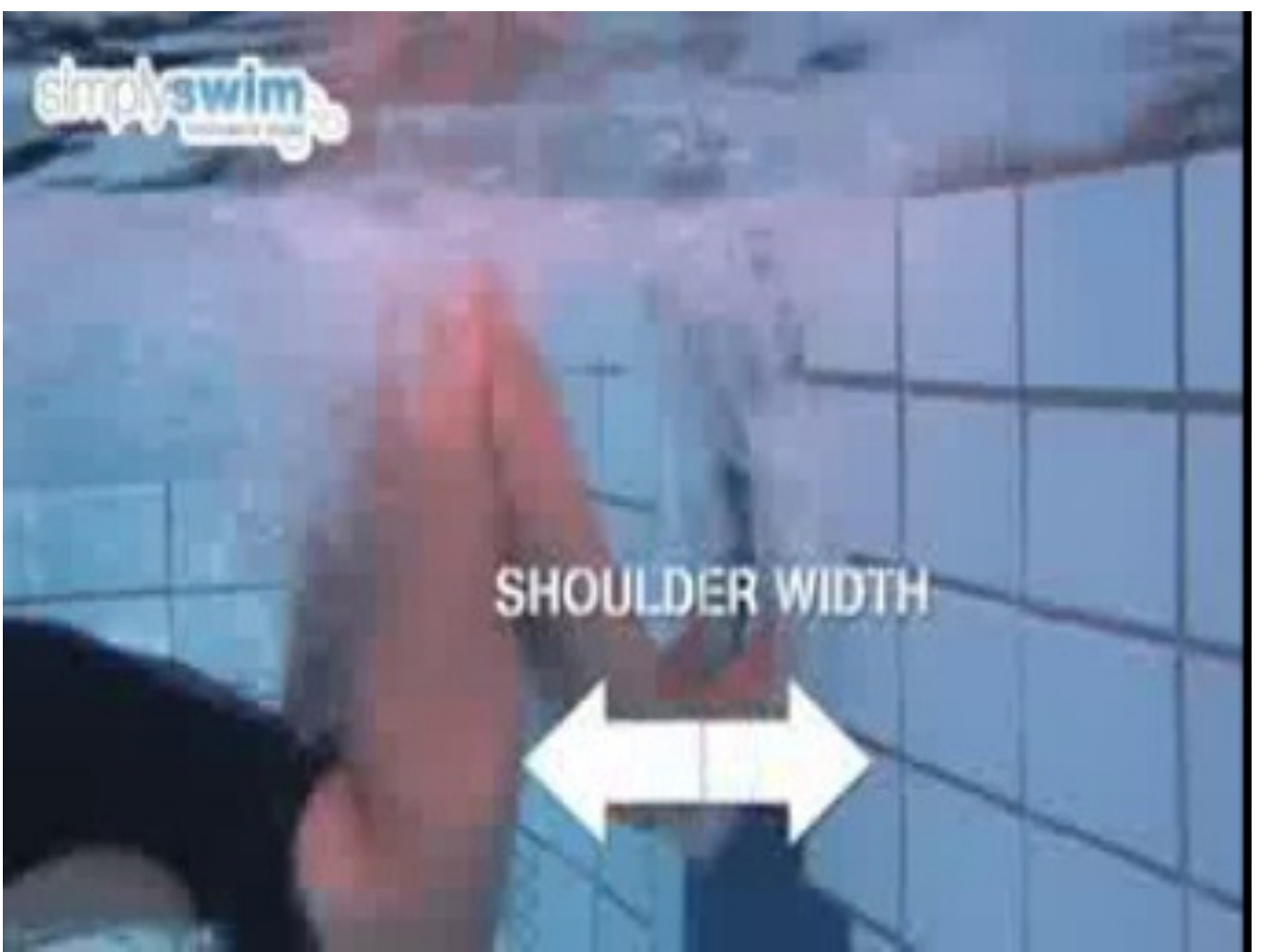}
  \hspace{-1.5mm}
  \includegraphics[width=0.121\linewidth]{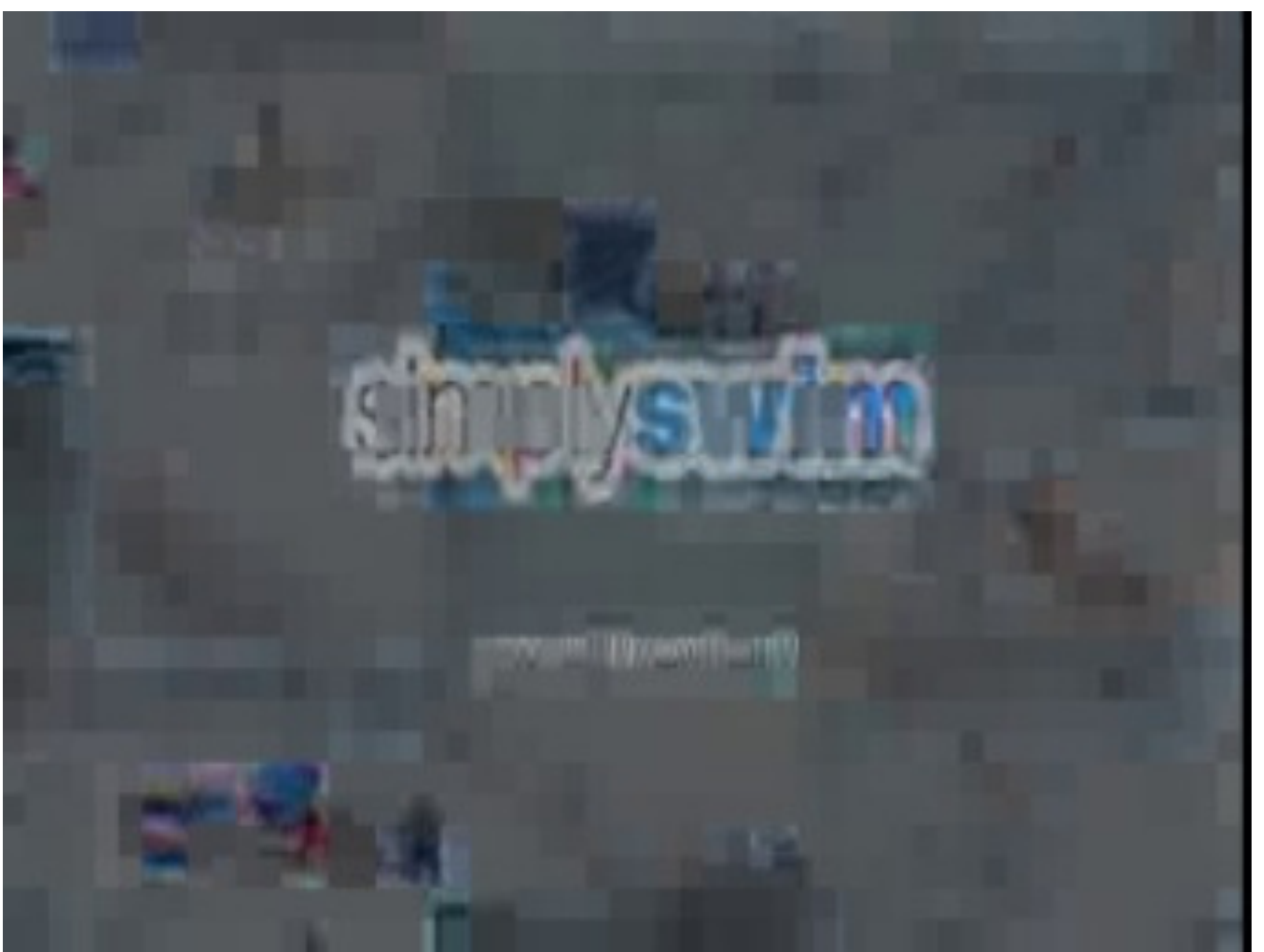}
  \hspace{-1.5mm}
  \includegraphics[width=0.121\linewidth]{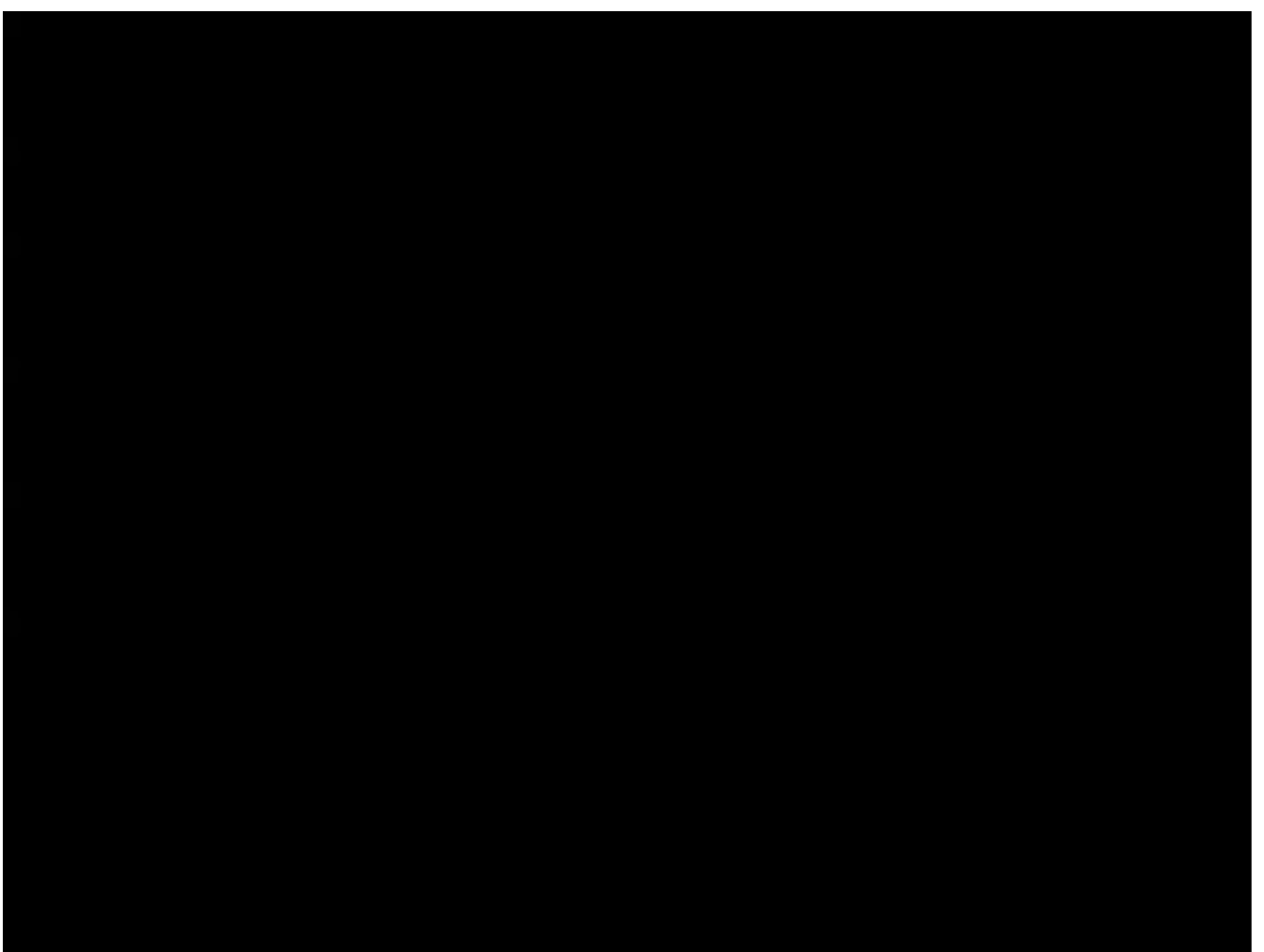}

  \includegraphics[width=0.121\linewidth]{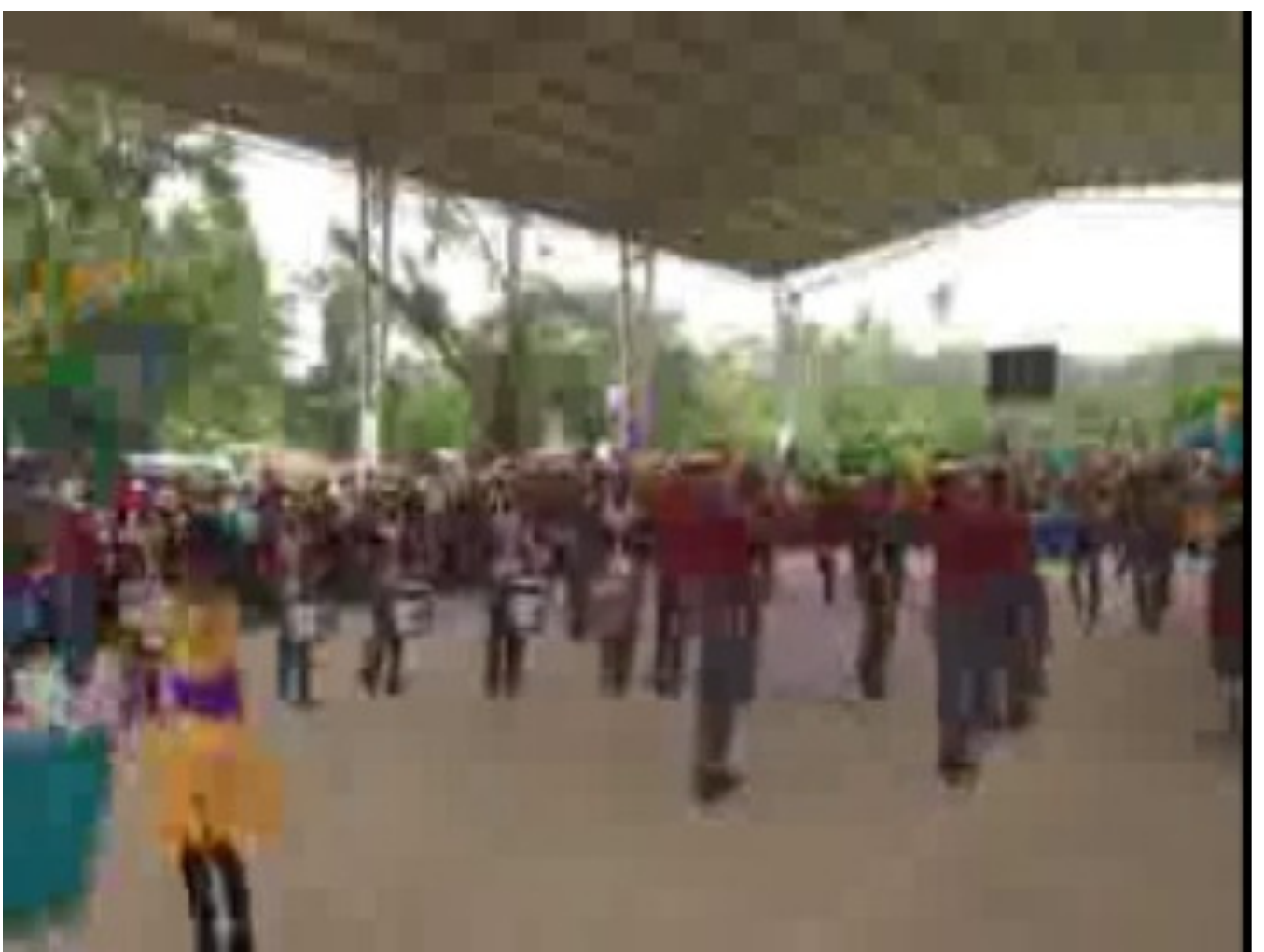}
  \hspace{-1.5mm}
  \includegraphics[width=0.121\linewidth]{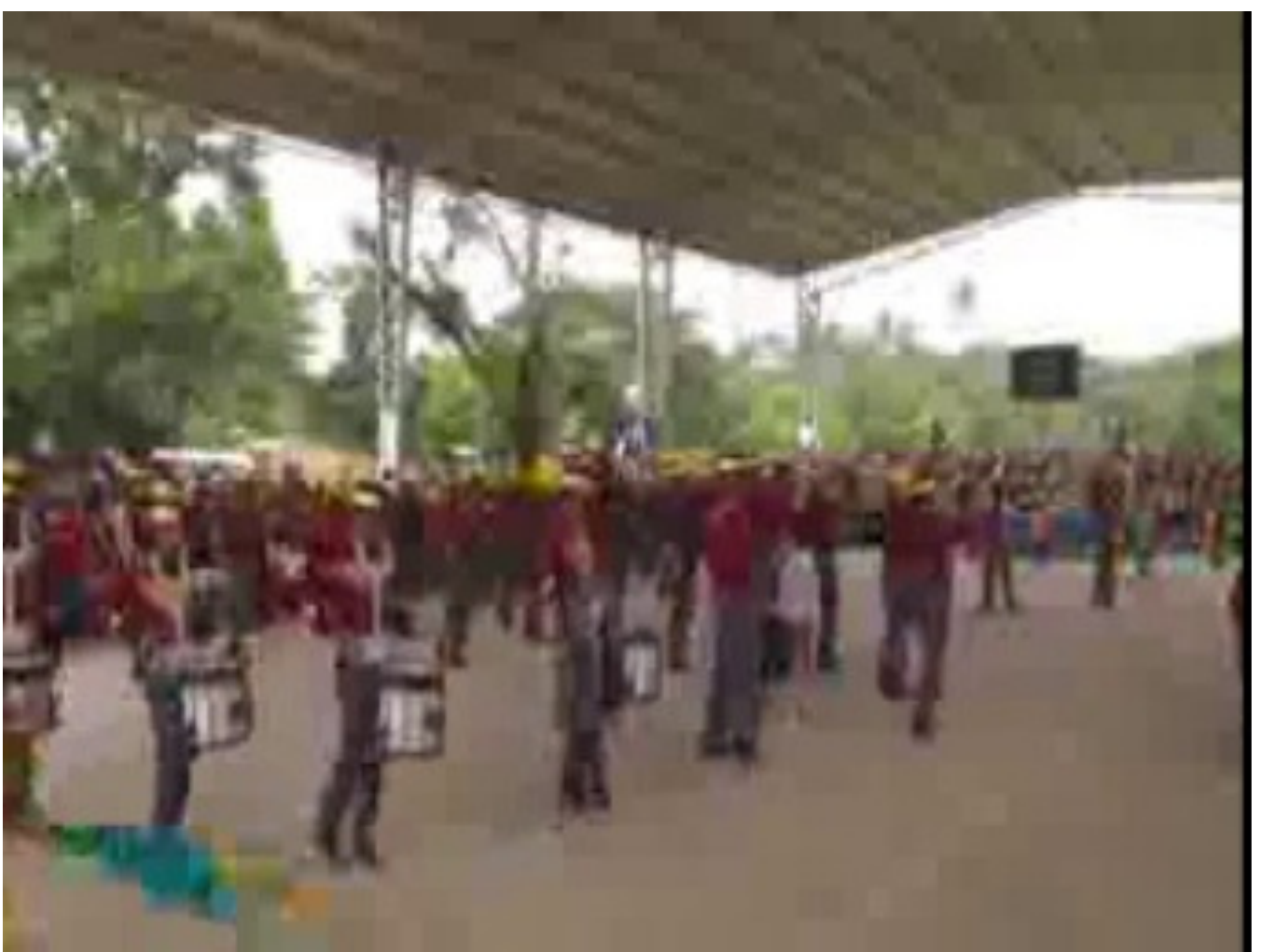}
  \hspace{-1.5mm}
  \includegraphics[width=0.121\linewidth]{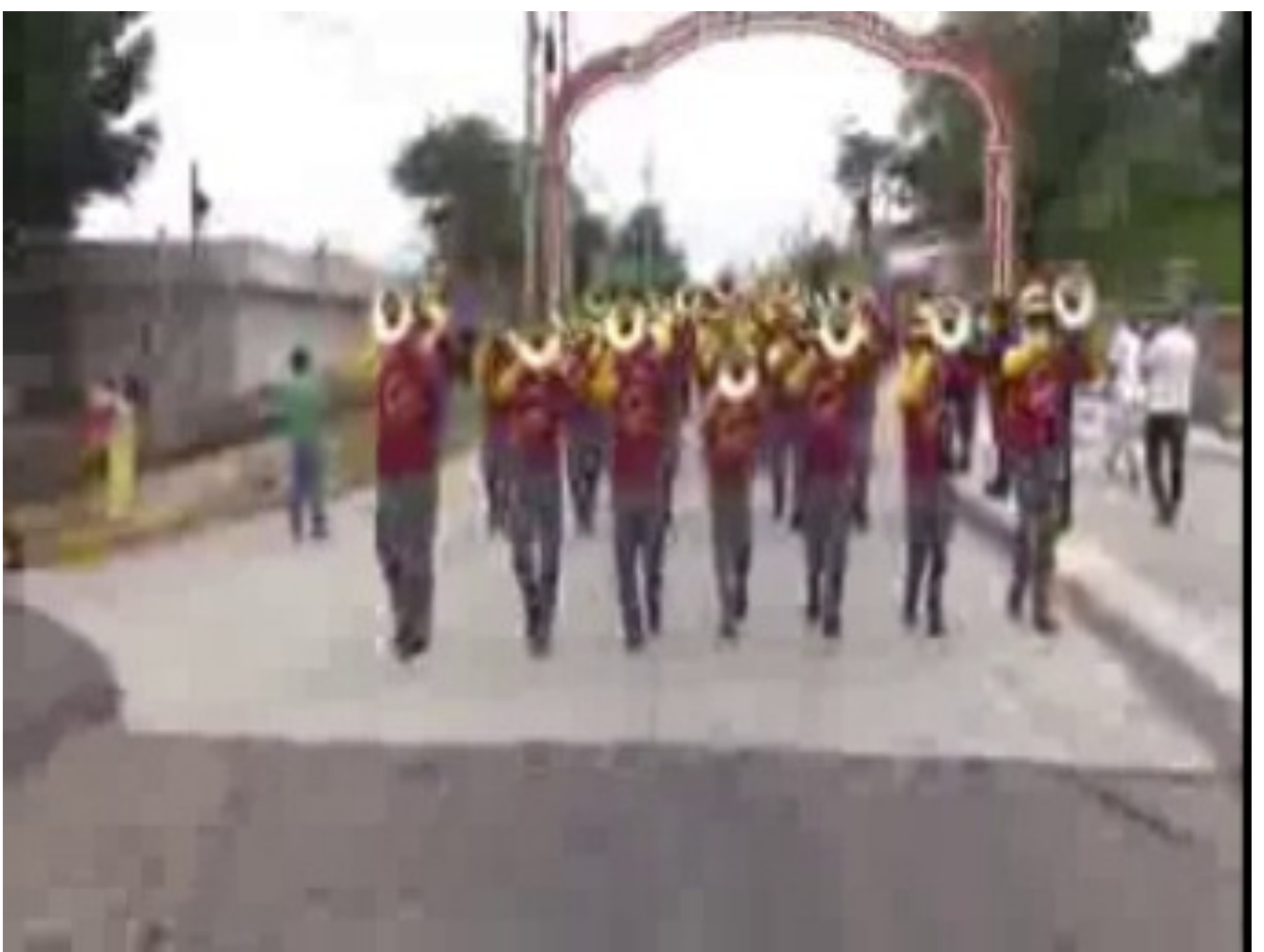}
  \hspace{-1.5mm}
  \includegraphics[width=0.121\linewidth]{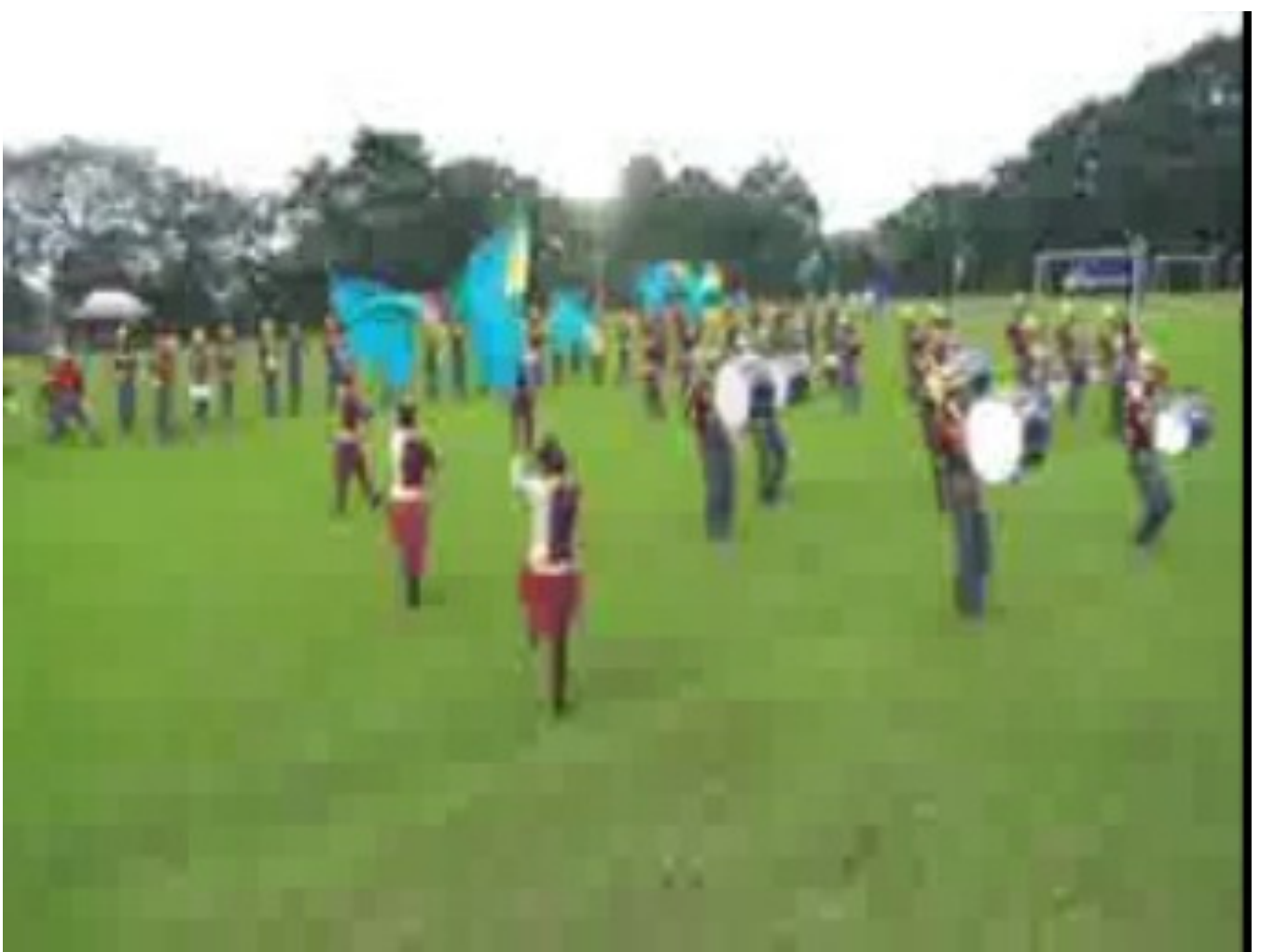}
  \hspace{-1.5mm}
  \includegraphics[width=0.121\linewidth]{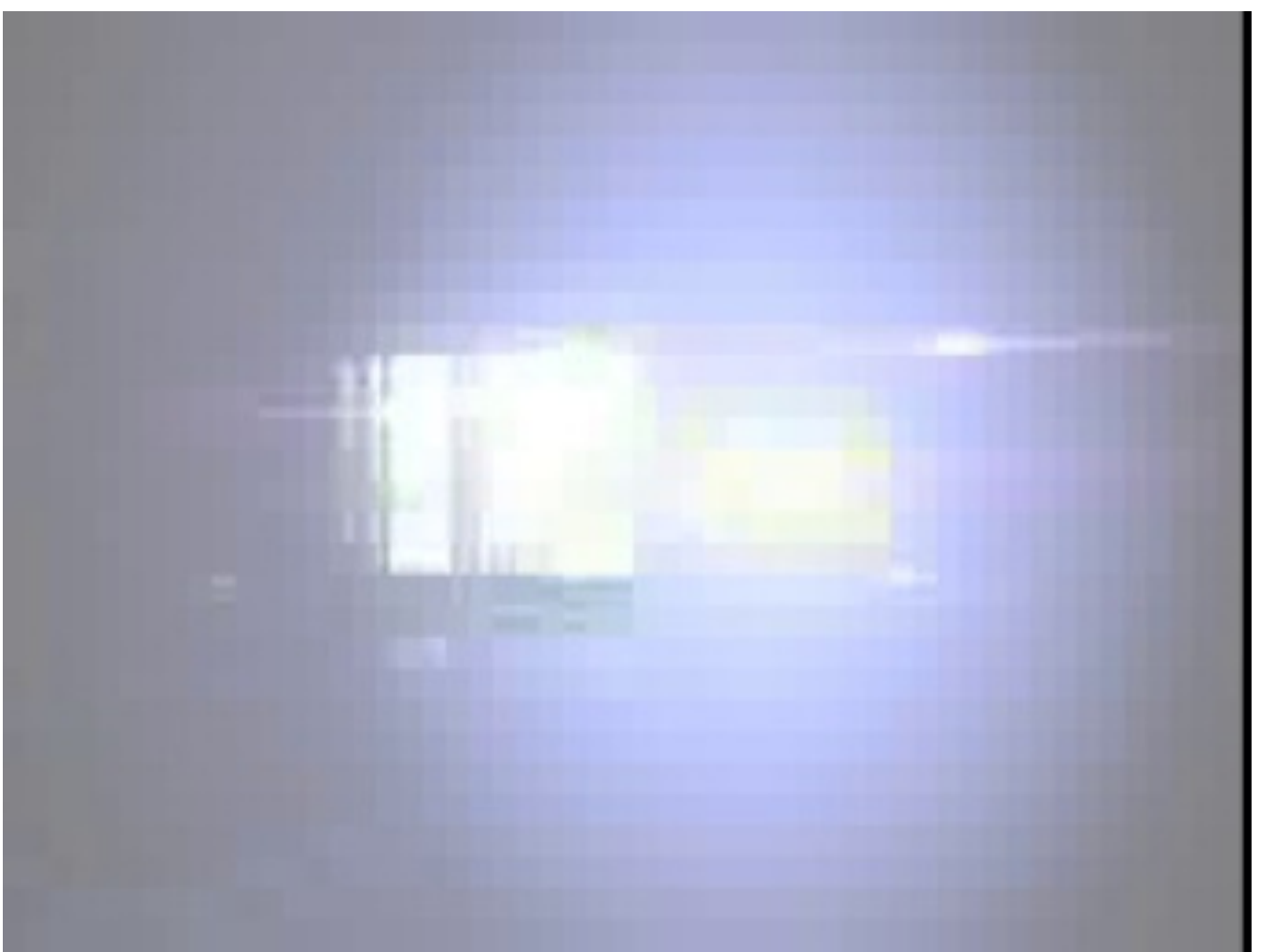}
  \hspace{-1.5mm}
  \includegraphics[width=0.121\linewidth]{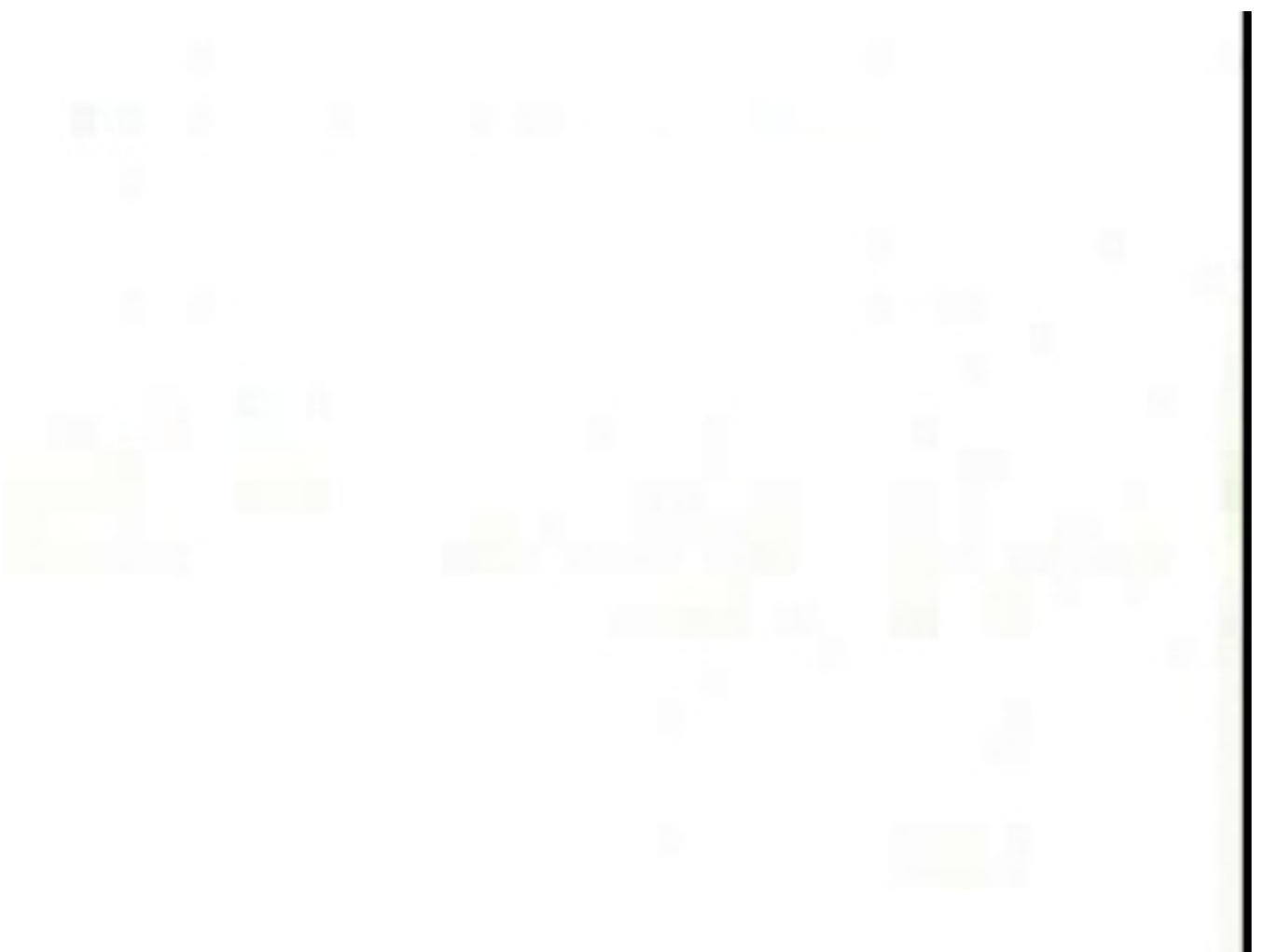}
  \hspace{-1.5mm}
  \includegraphics[width=0.121\linewidth]{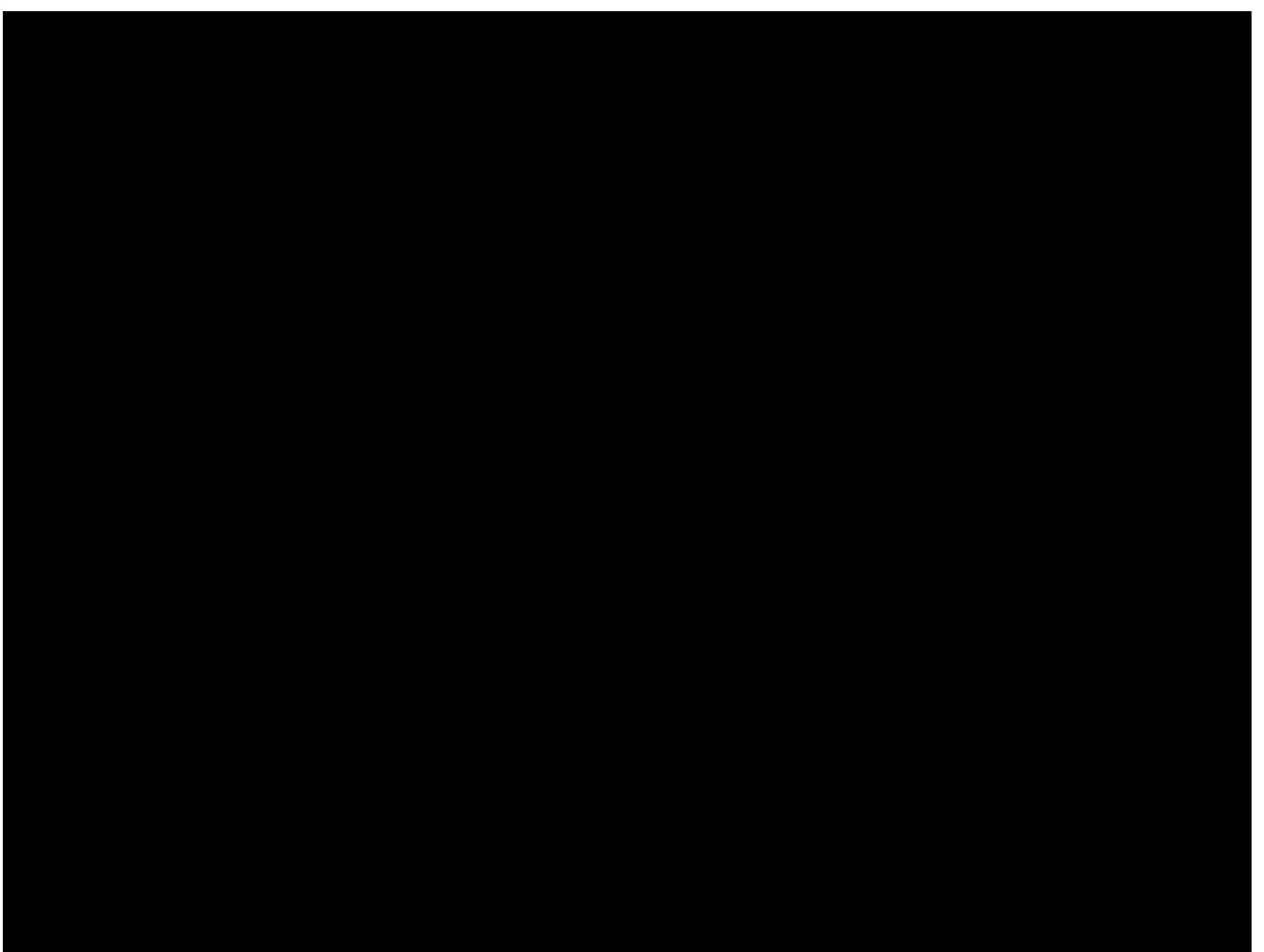}
  \hspace{-1.5mm}
  \includegraphics[width=0.121\linewidth]{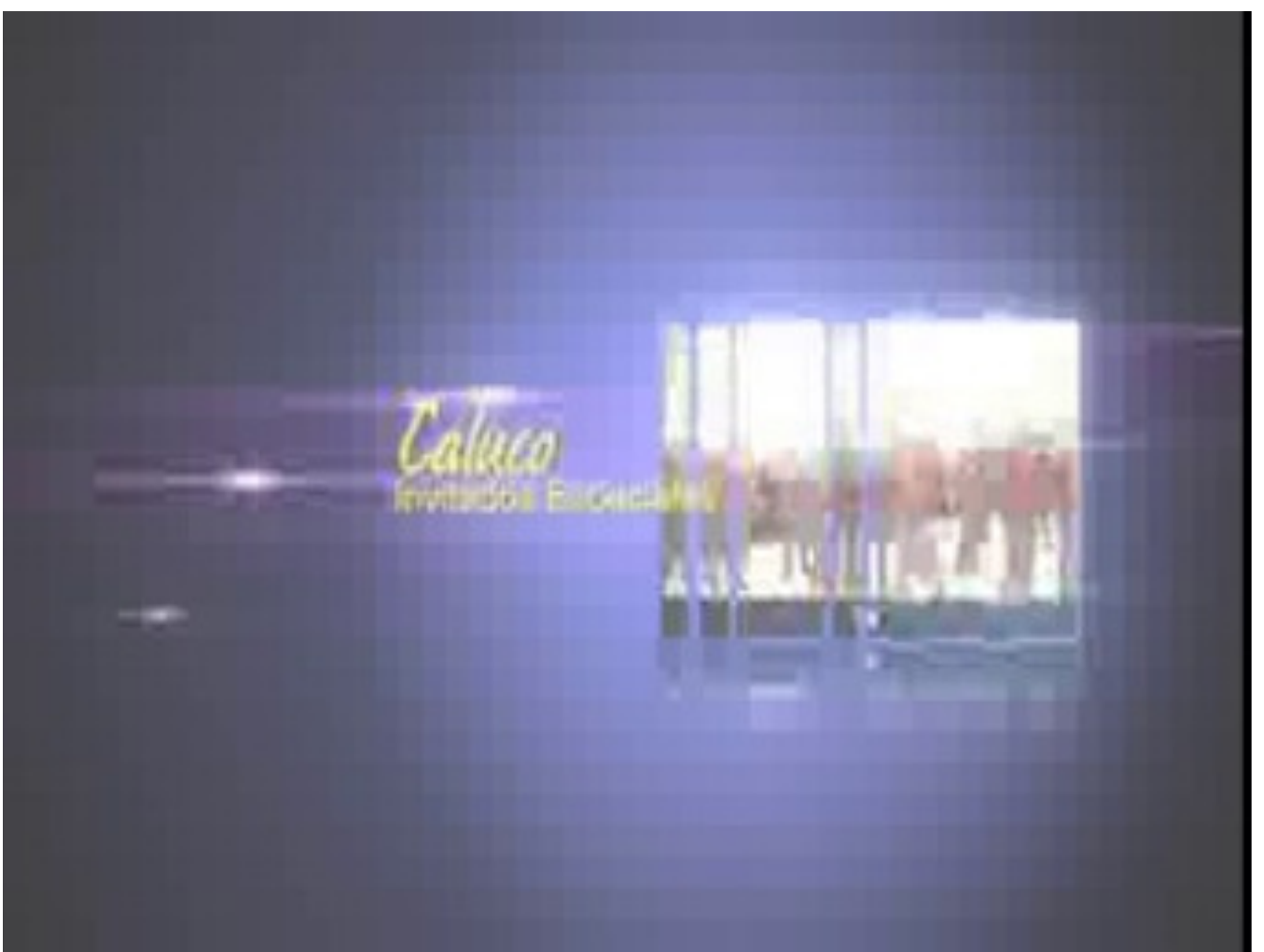}

  \includegraphics[width=0.121\linewidth]{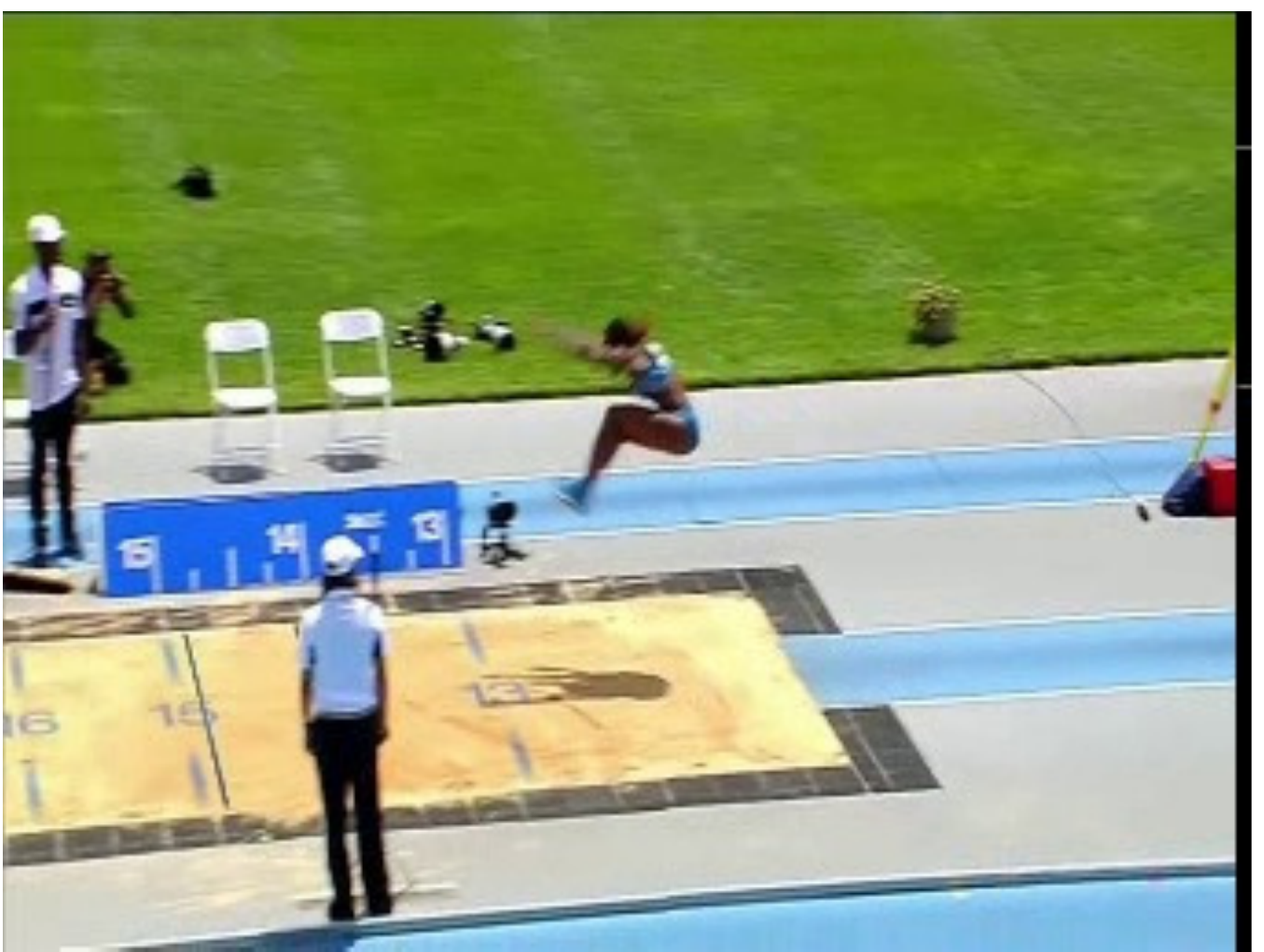}
  \hspace{-1.5mm}
  \includegraphics[width=0.121\linewidth]{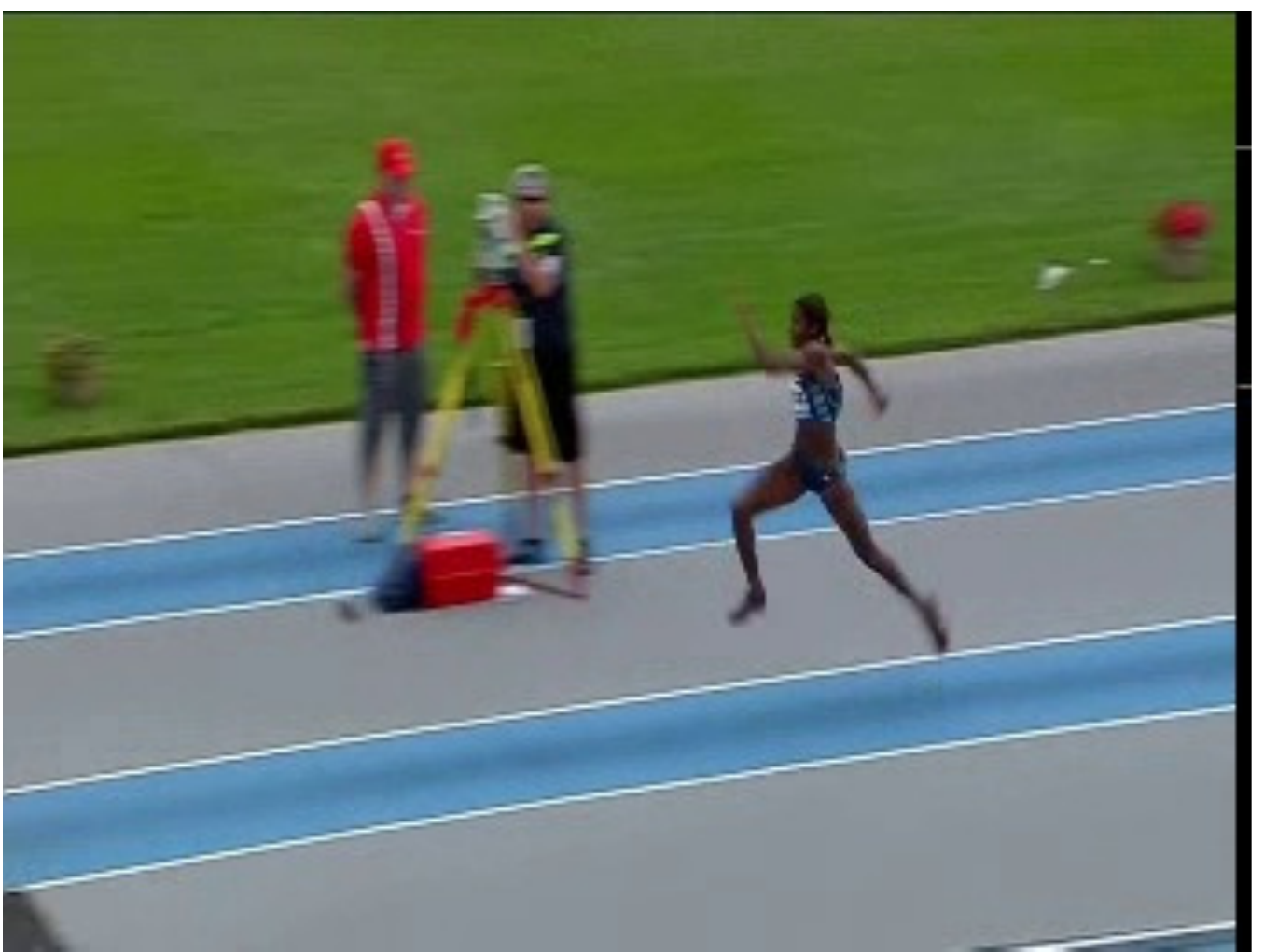}
  \hspace{-1.5mm}
  \includegraphics[width=0.121\linewidth]{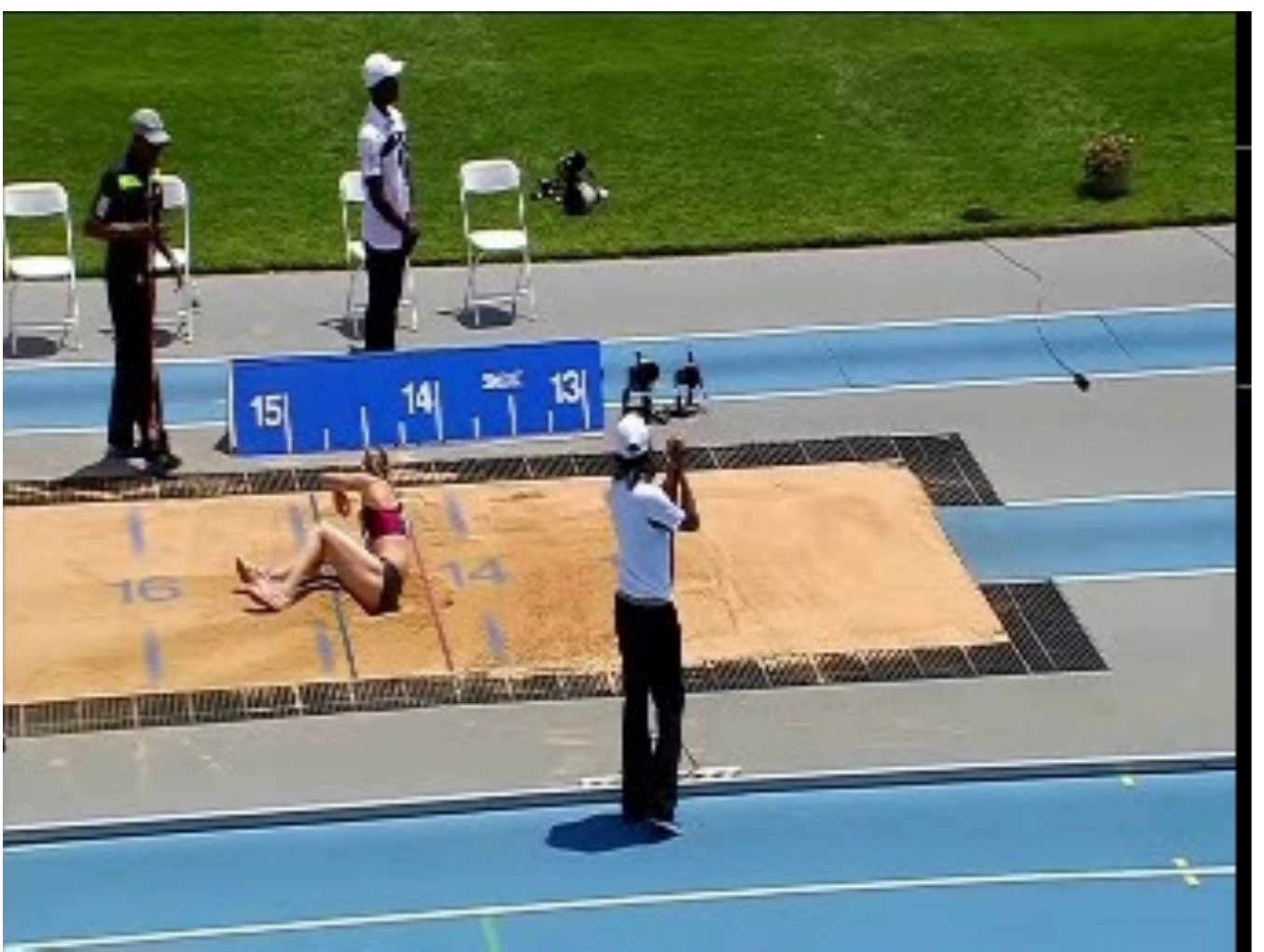}
  \hspace{-1.5mm}
  \includegraphics[width=0.121\linewidth]{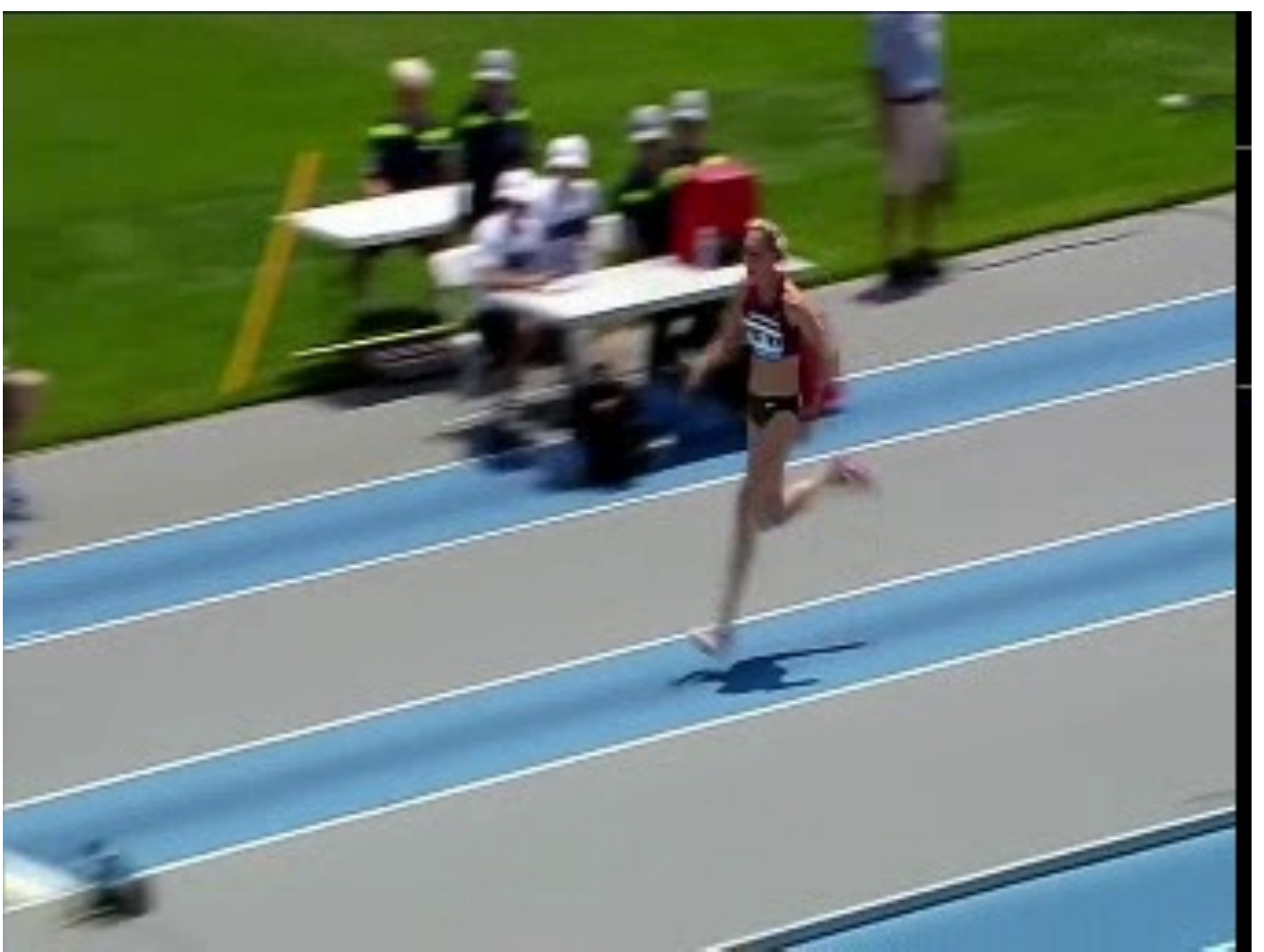}
  \hspace{-1.5mm}
  \includegraphics[width=0.121\linewidth]{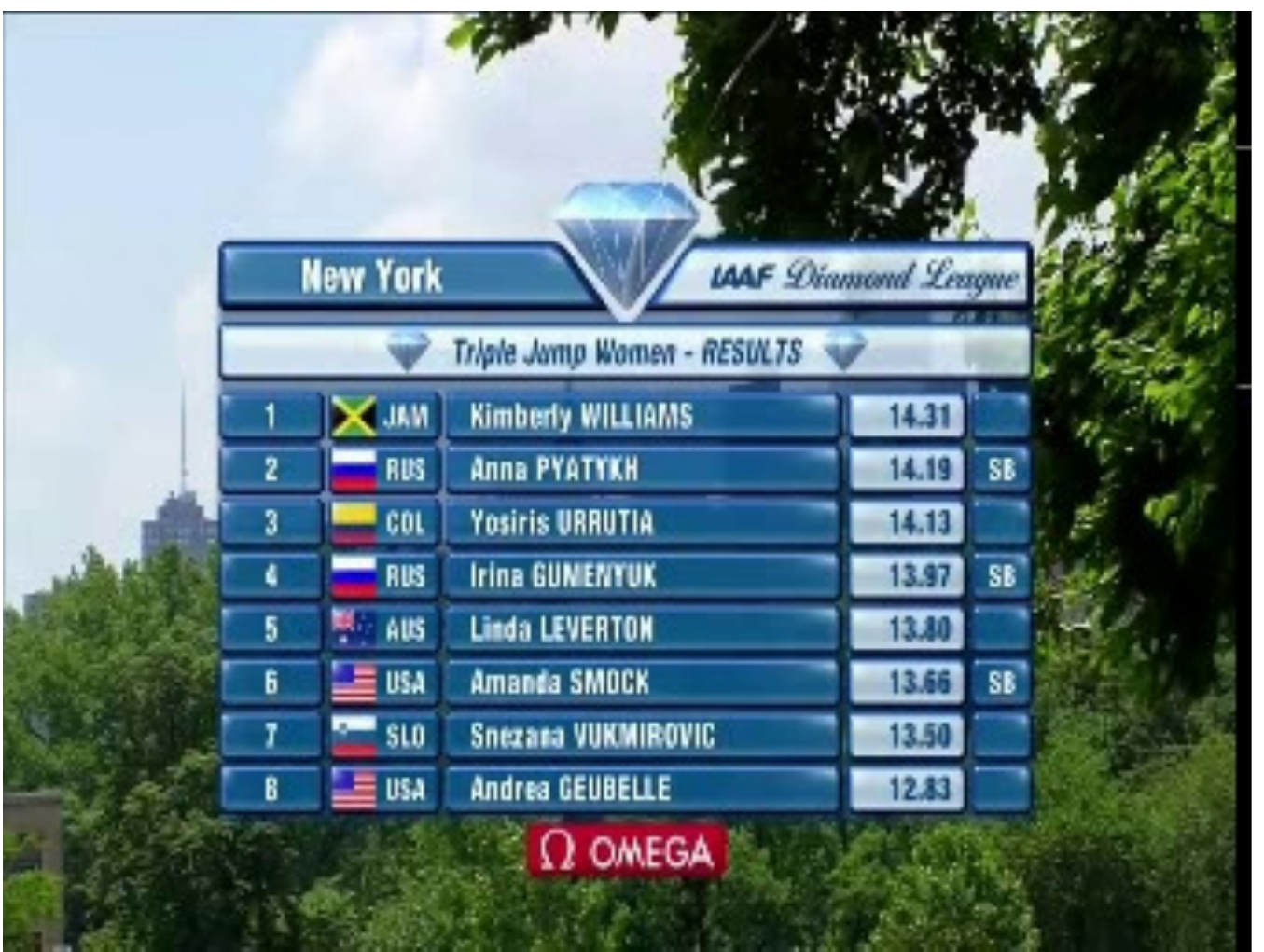}
  \hspace{-1.5mm}
  \includegraphics[width=0.121\linewidth]{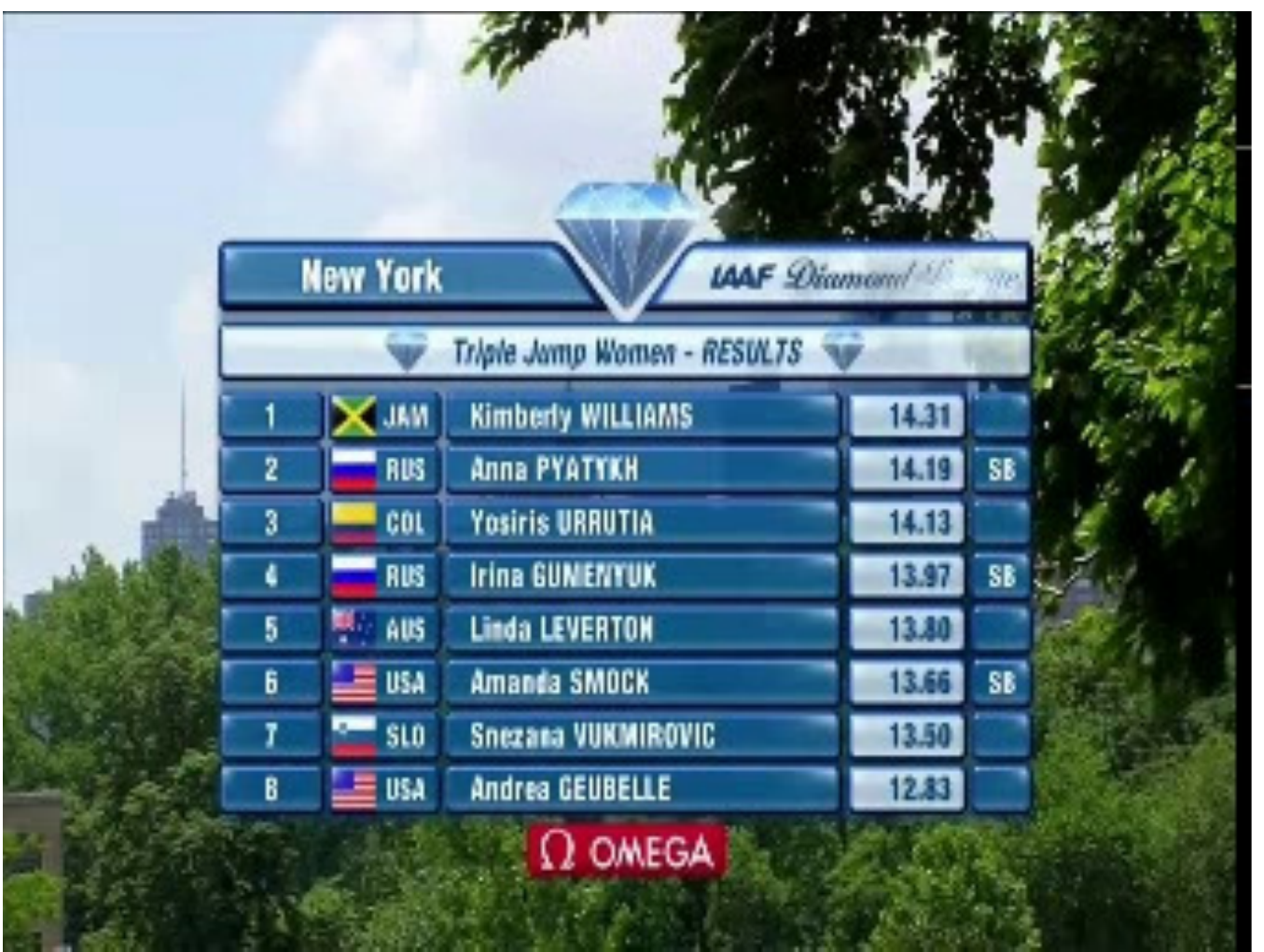}
  \hspace{-1.5mm}
  \includegraphics[width=0.121\linewidth]{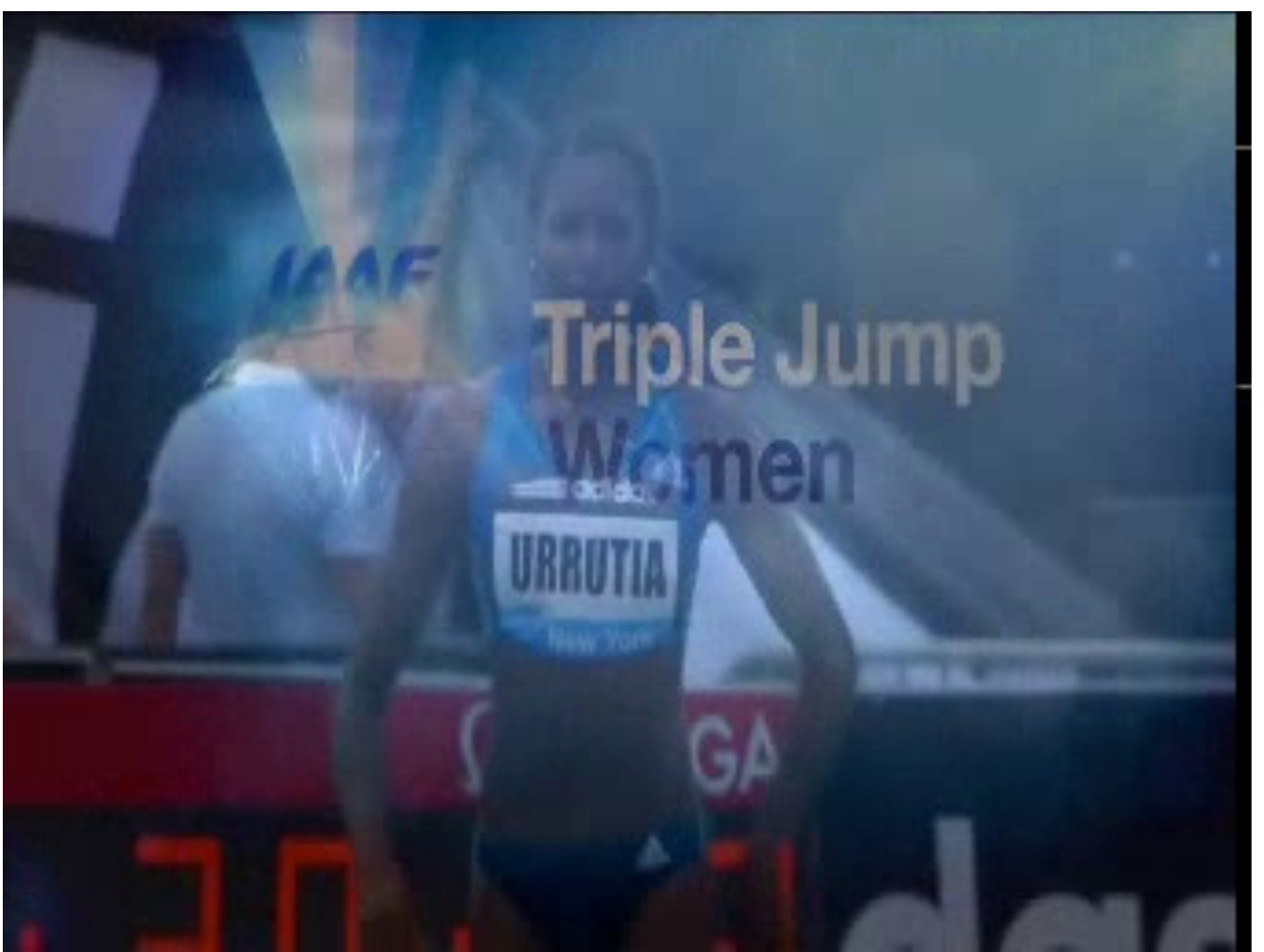}
  \hspace{-1.5mm}
  \includegraphics[width=0.121\linewidth]{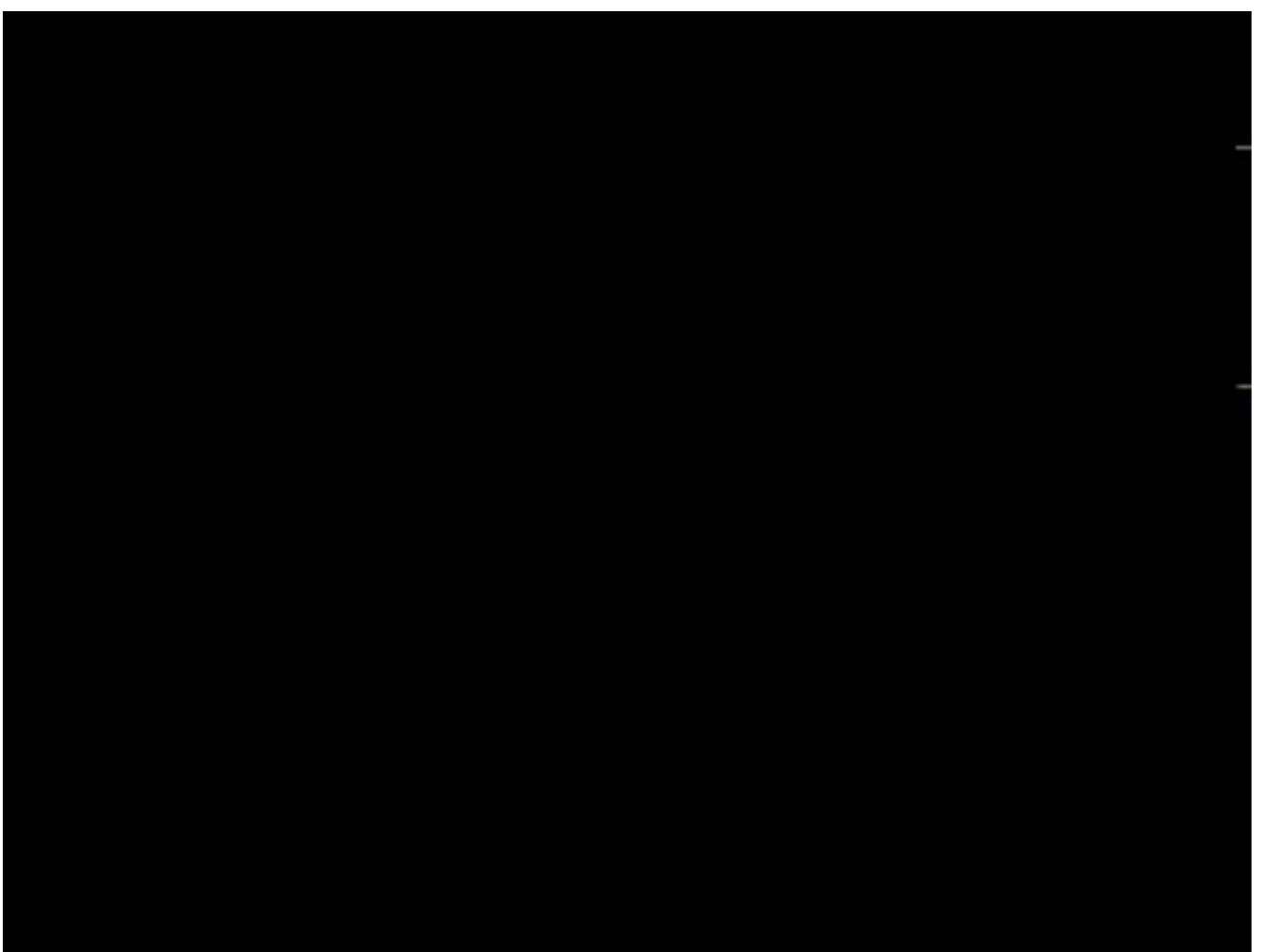}

  \includegraphics[width=0.121\linewidth]{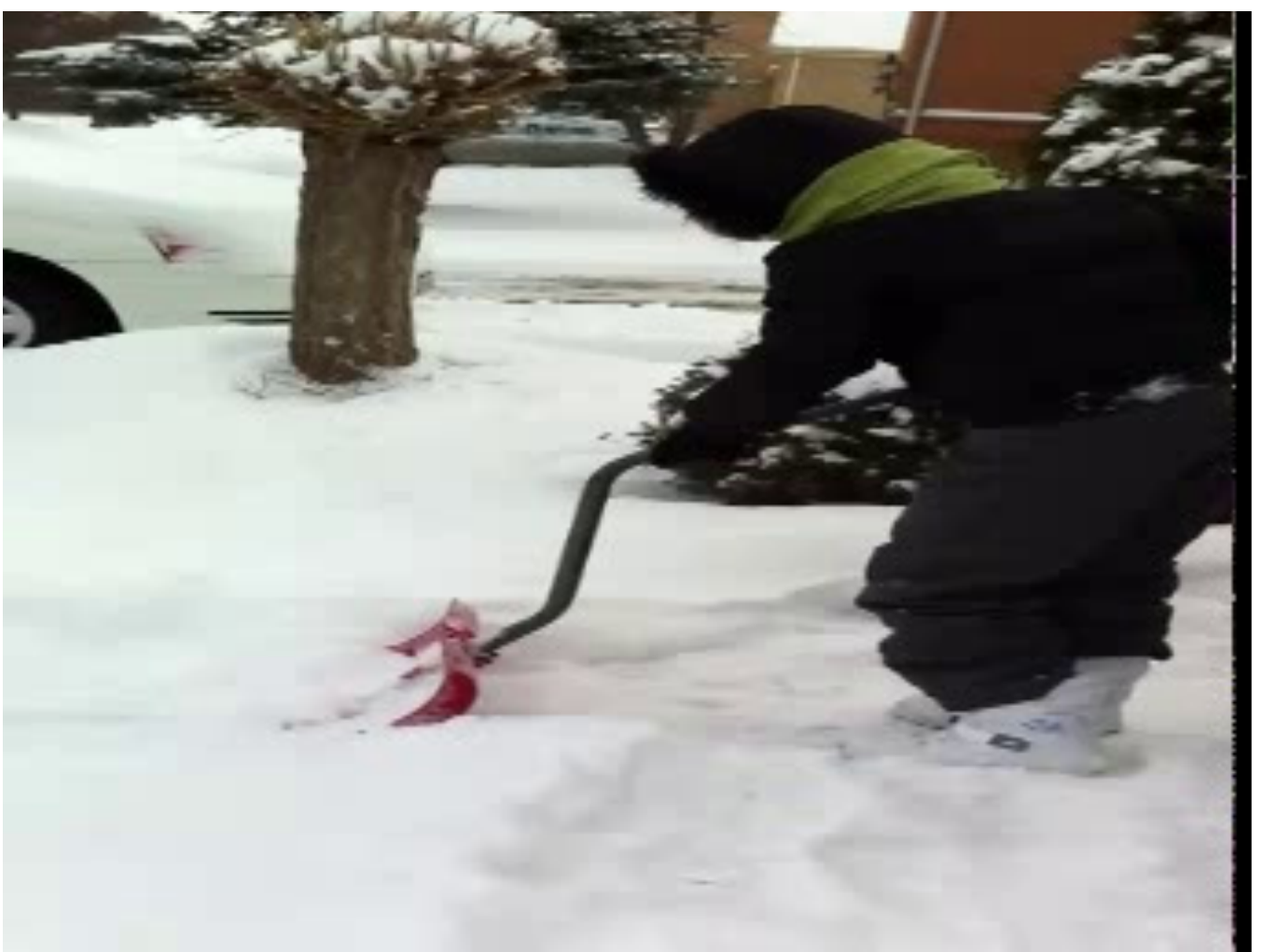}
  \hspace{-1.5mm}
  \includegraphics[width=0.121\linewidth]{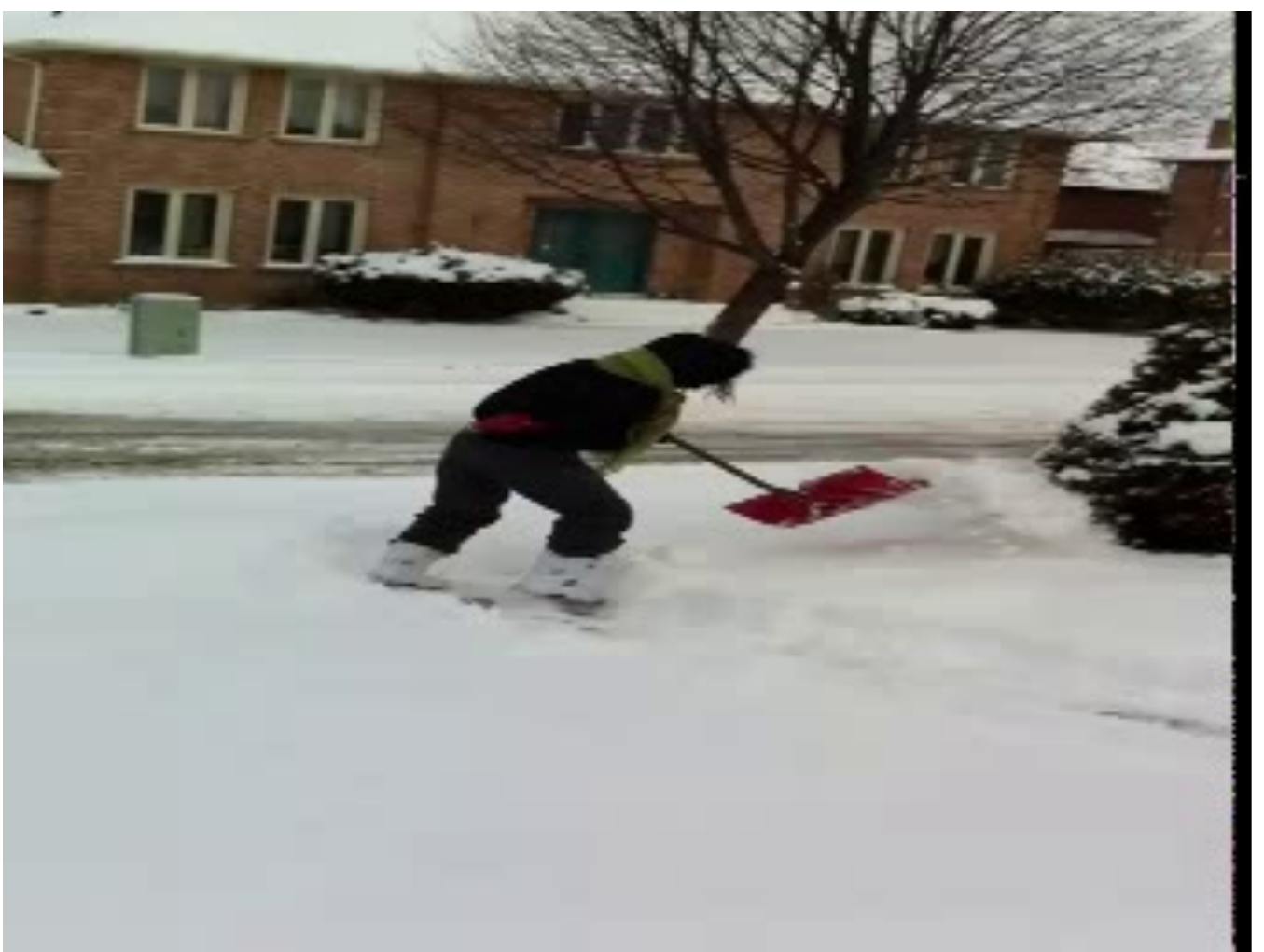}
  \hspace{-1.5mm}
  \includegraphics[width=0.121\linewidth]{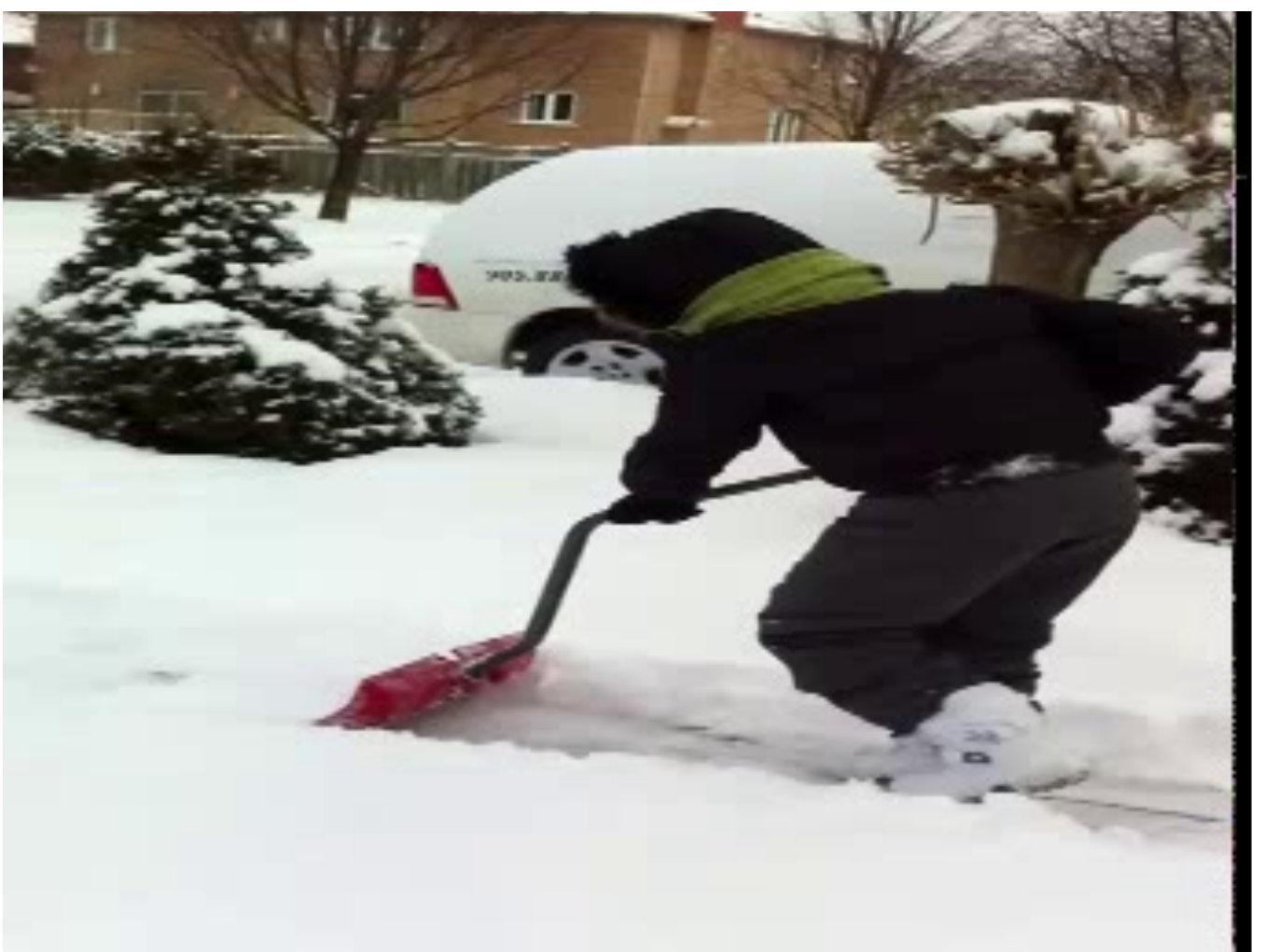}
  \hspace{-1.5mm}
  \includegraphics[width=0.121\linewidth]{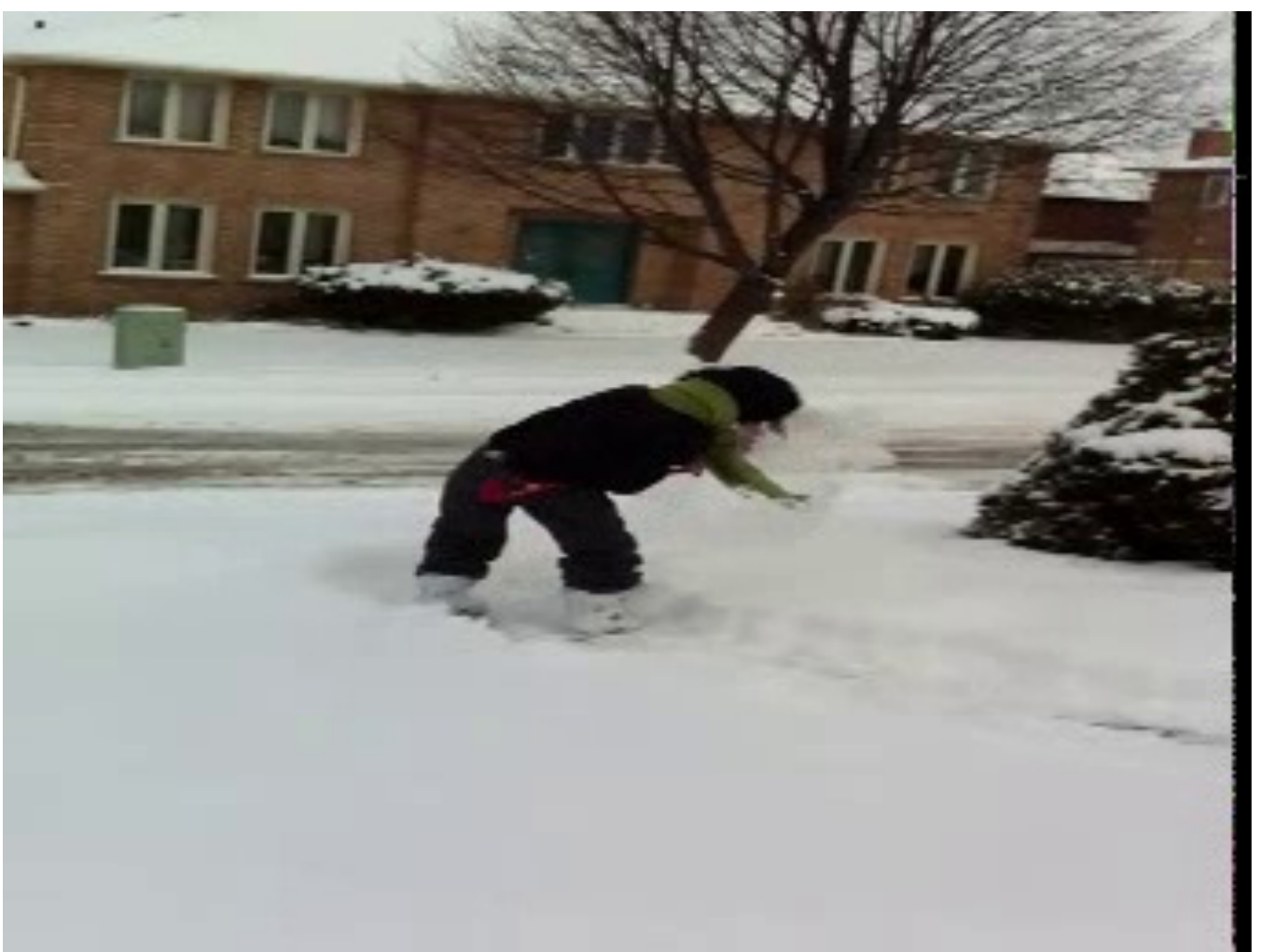}
  \hspace{-1.5mm}
  \includegraphics[width=0.121\linewidth]{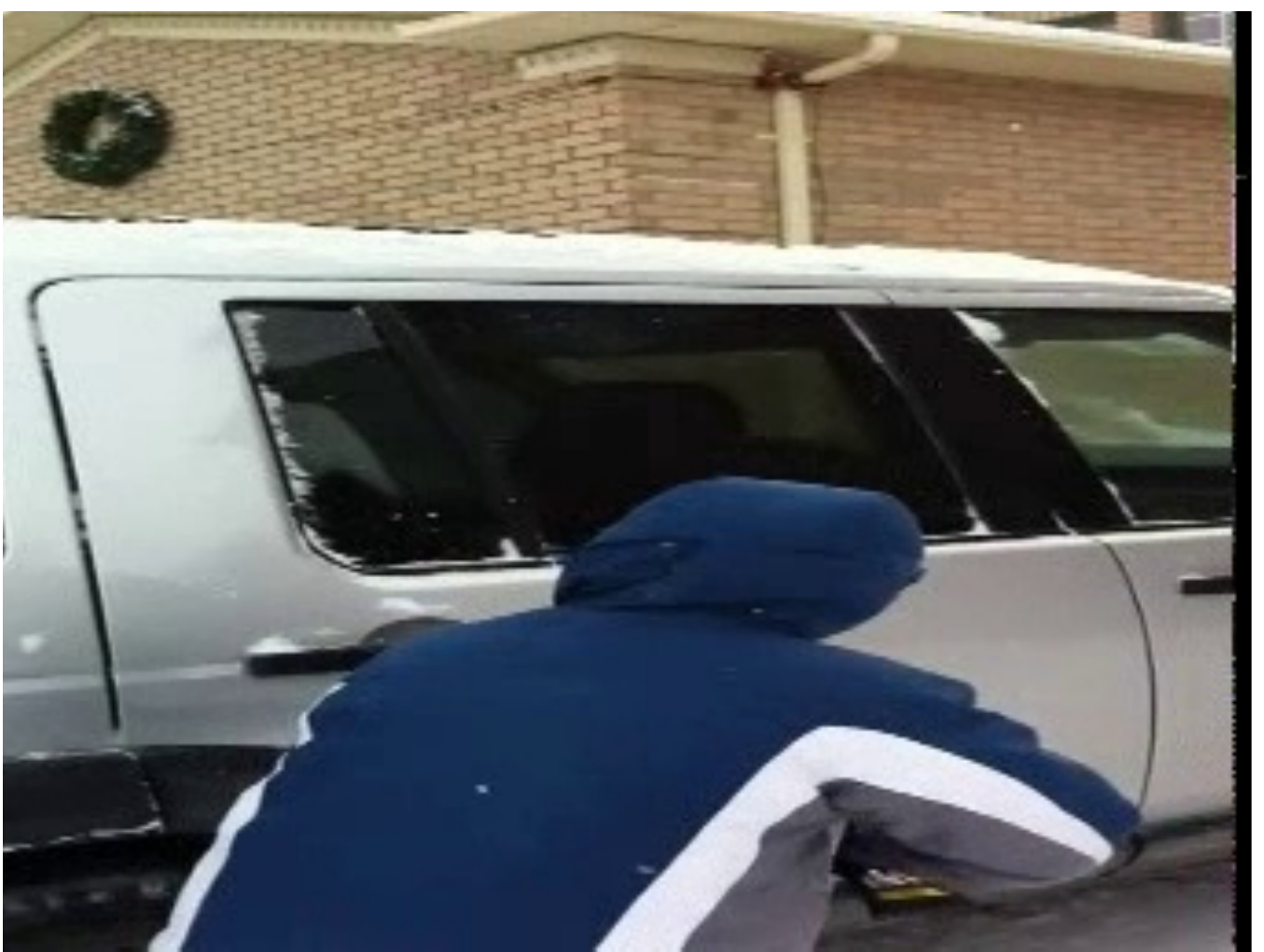}
  \hspace{-1.5mm}
  \includegraphics[width=0.121\linewidth]{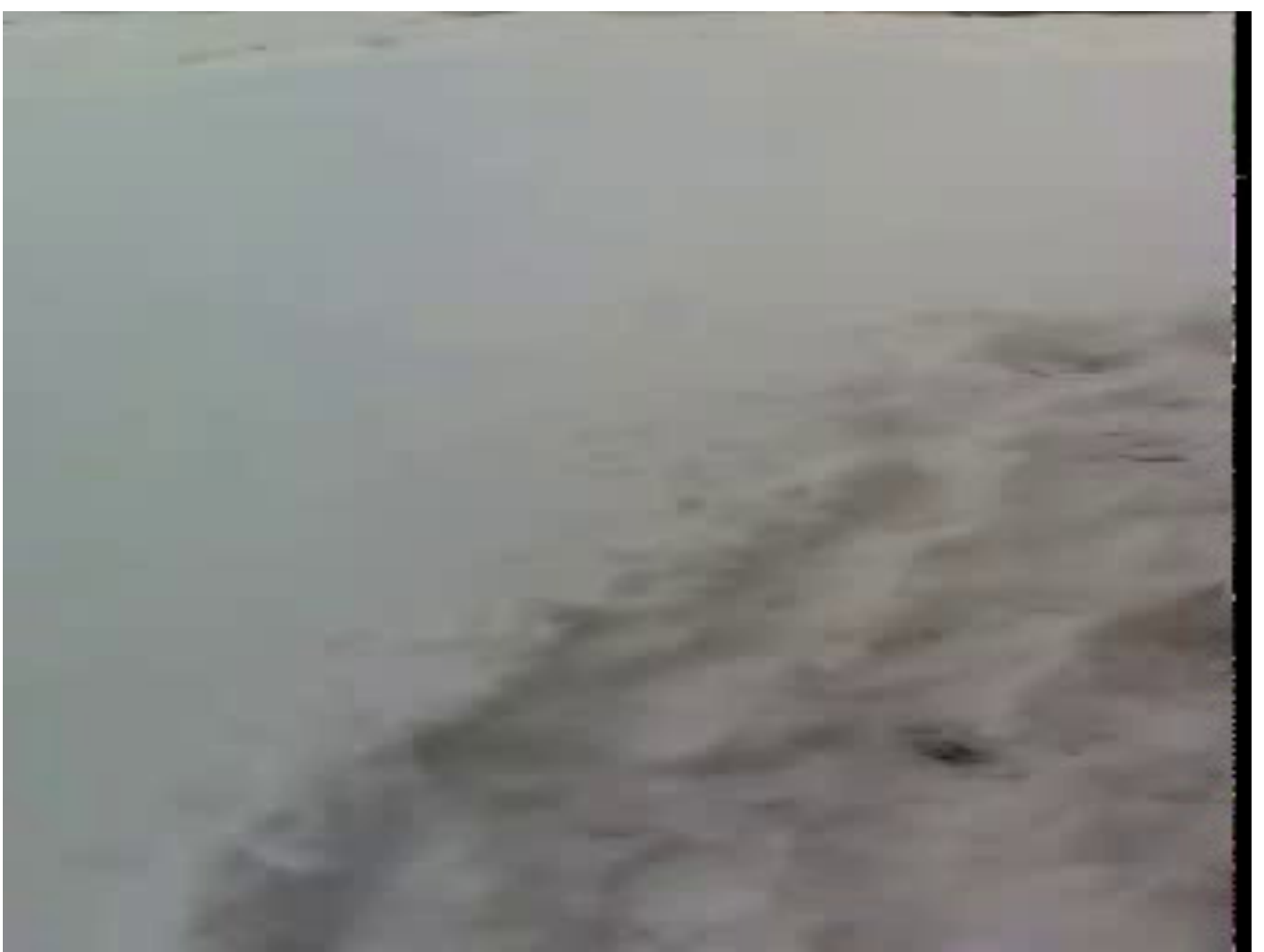}
  \hspace{-1.5mm}
  \includegraphics[width=0.121\linewidth]{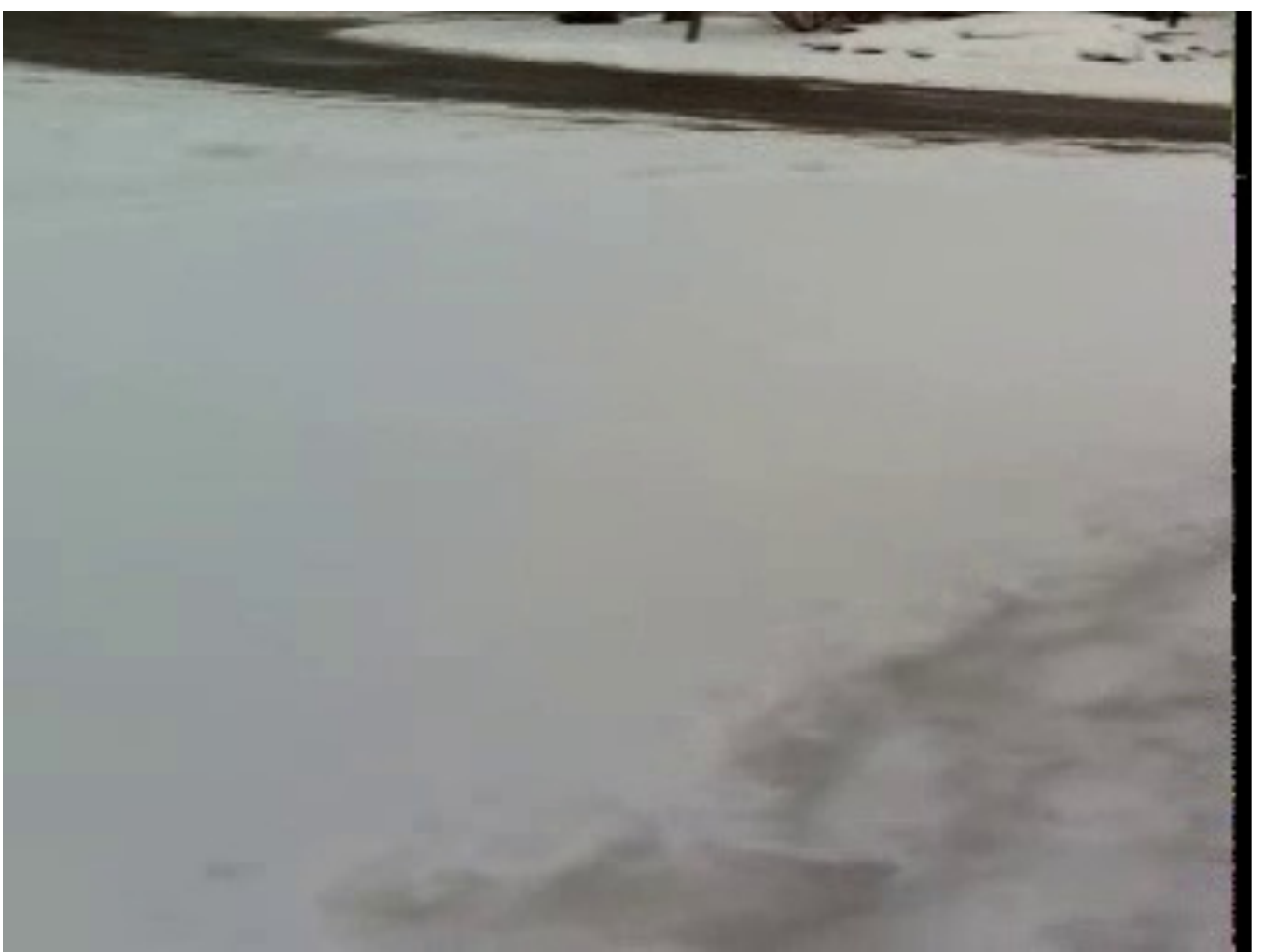}
  \hspace{-1.5mm}
  \includegraphics[width=0.121\linewidth]{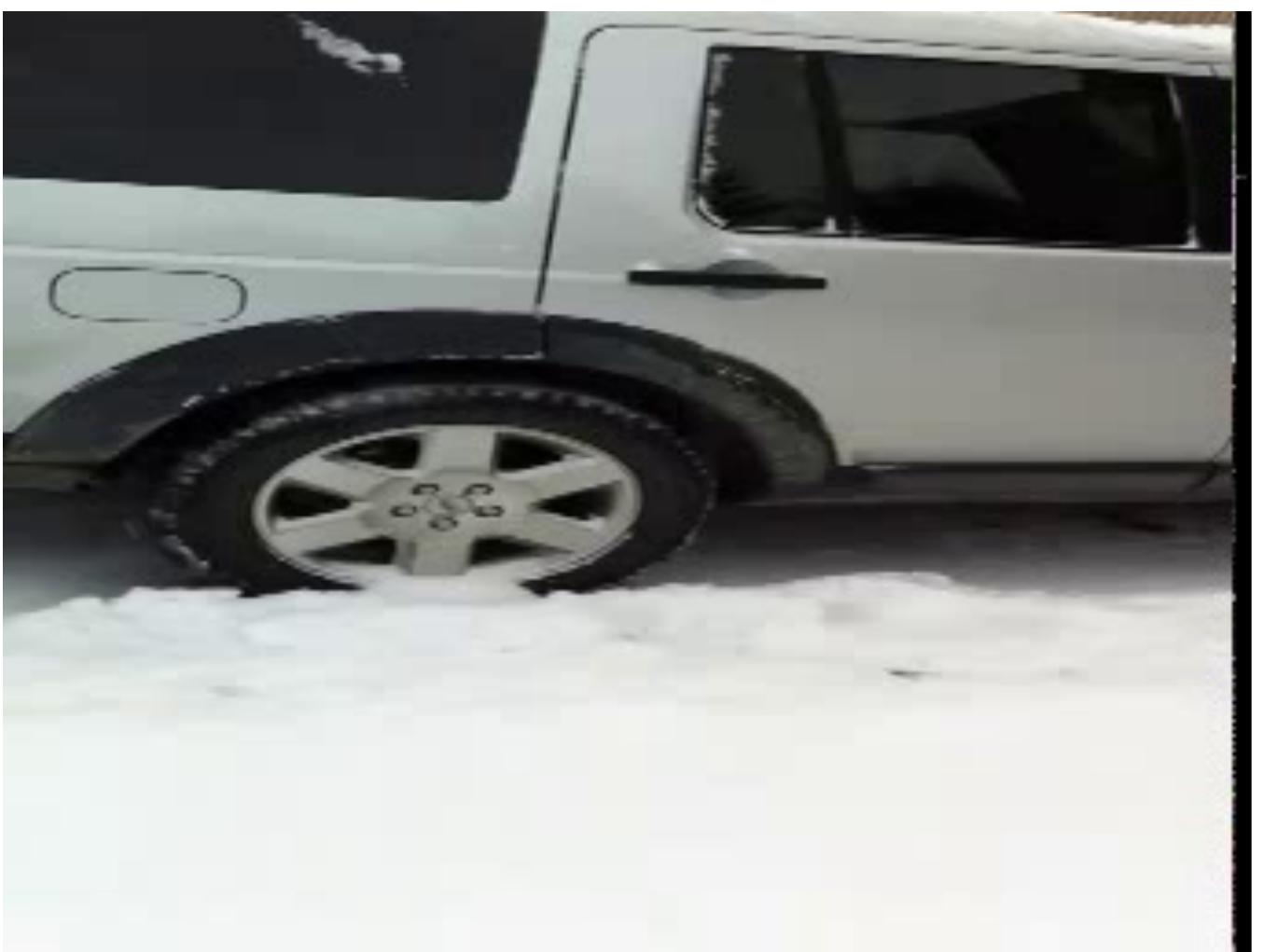}

  \includegraphics[width=0.121\linewidth]{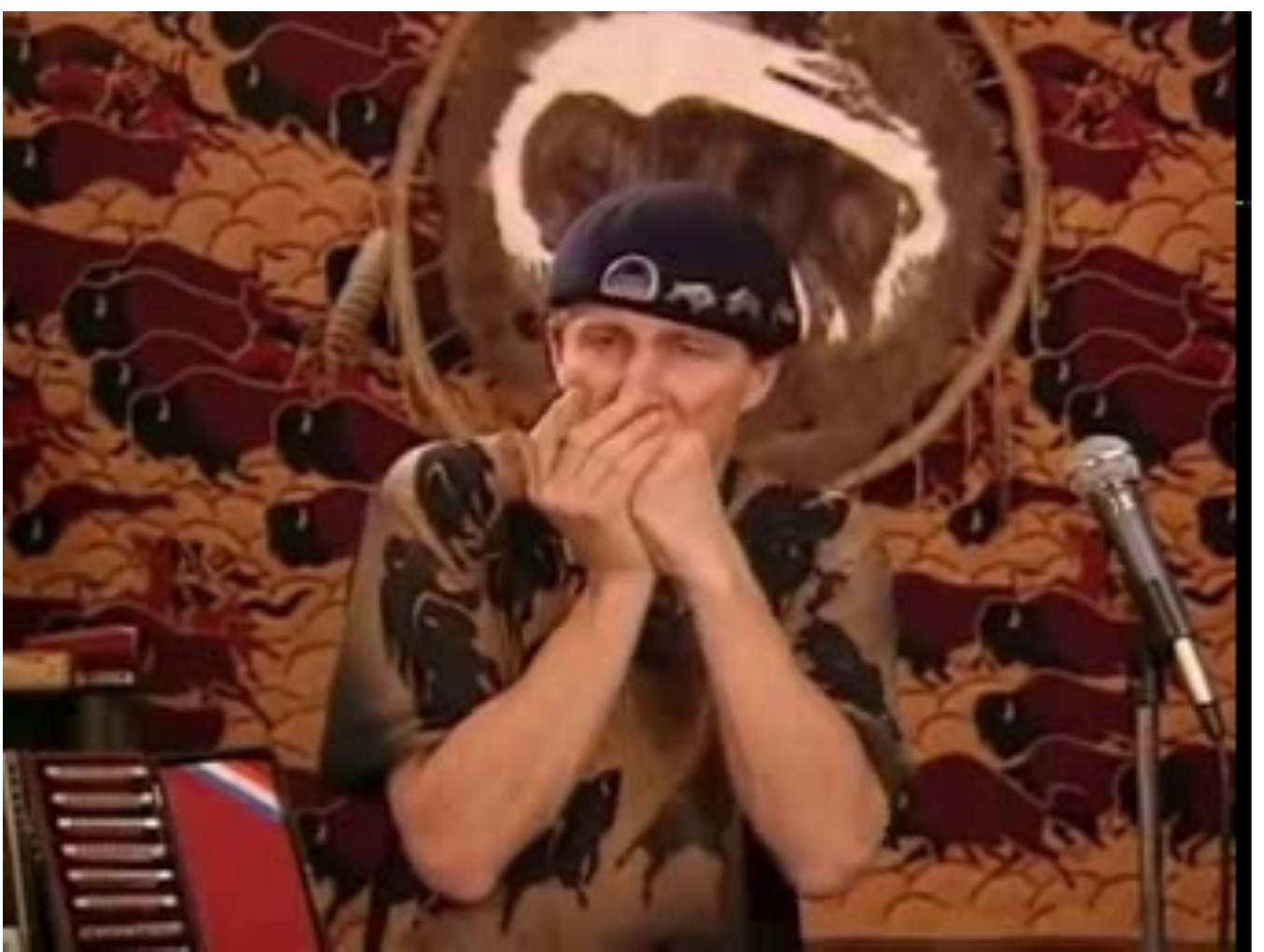}
  \hspace{-1.5mm}
  \includegraphics[width=0.121\linewidth]{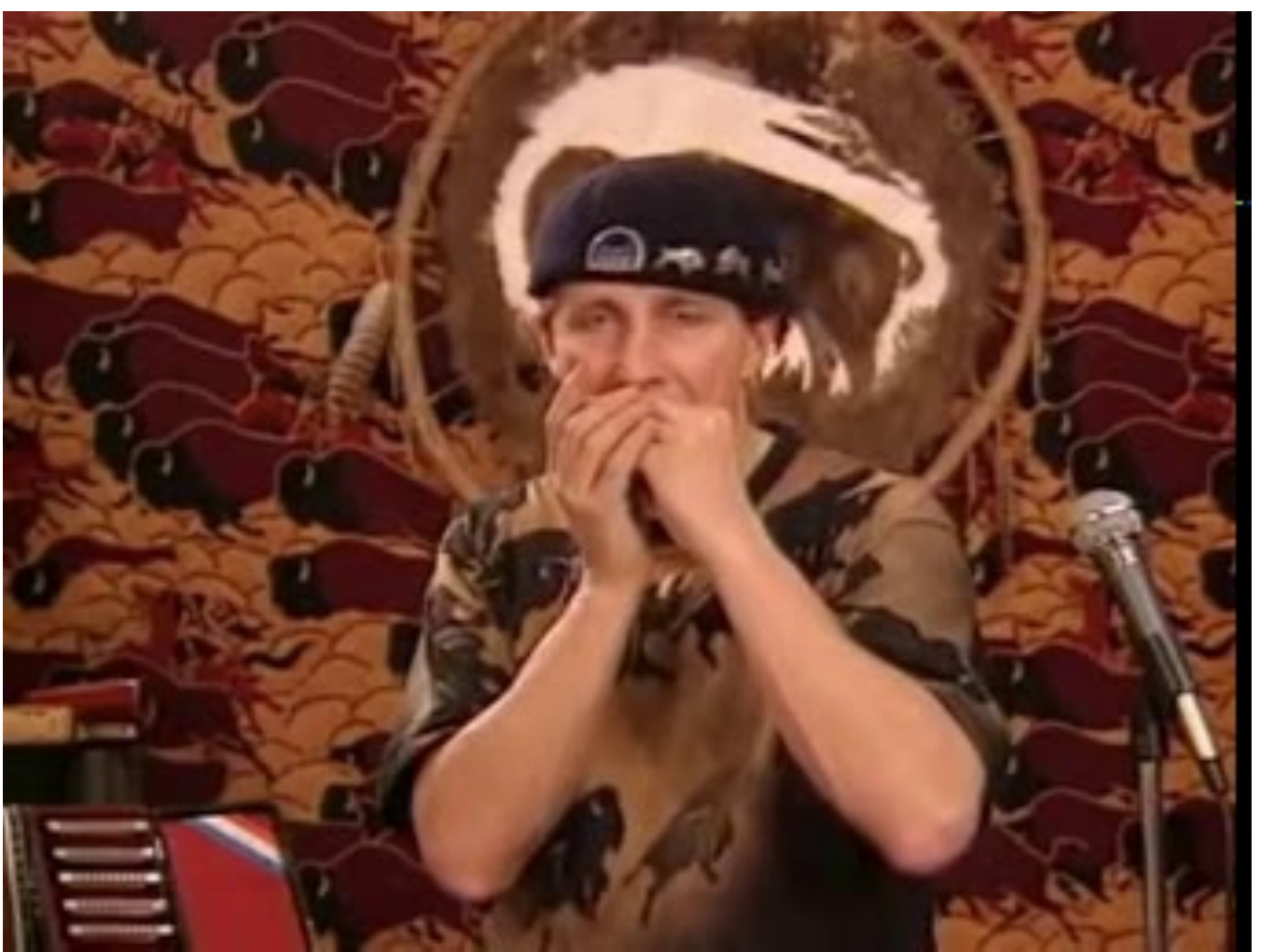}
  \hspace{-1.5mm}
  \includegraphics[width=0.121\linewidth]{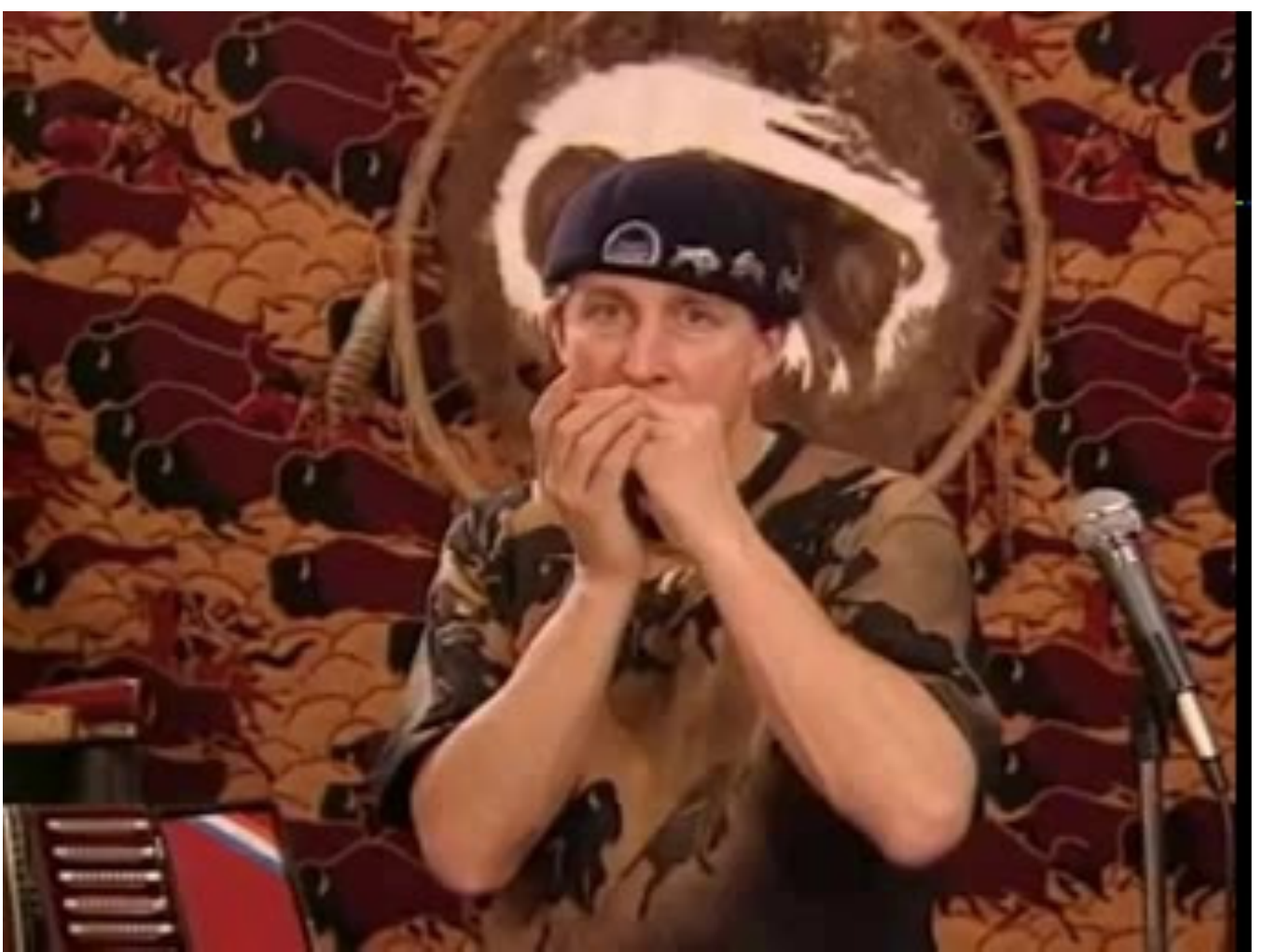}
  \hspace{-1.5mm}
  \includegraphics[width=0.121\linewidth]{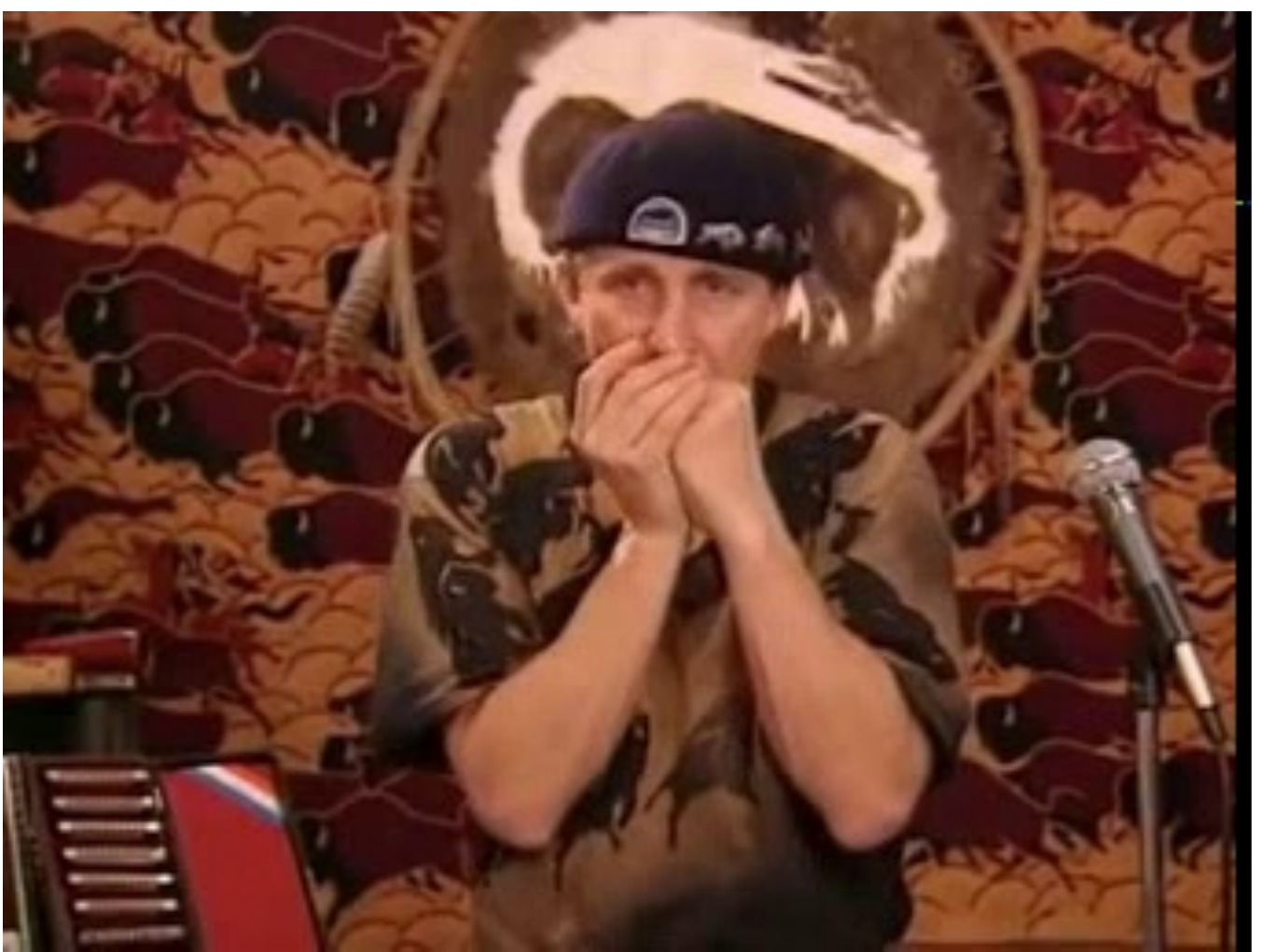}
  \hspace{-1.5mm}
  \includegraphics[width=0.121\linewidth]{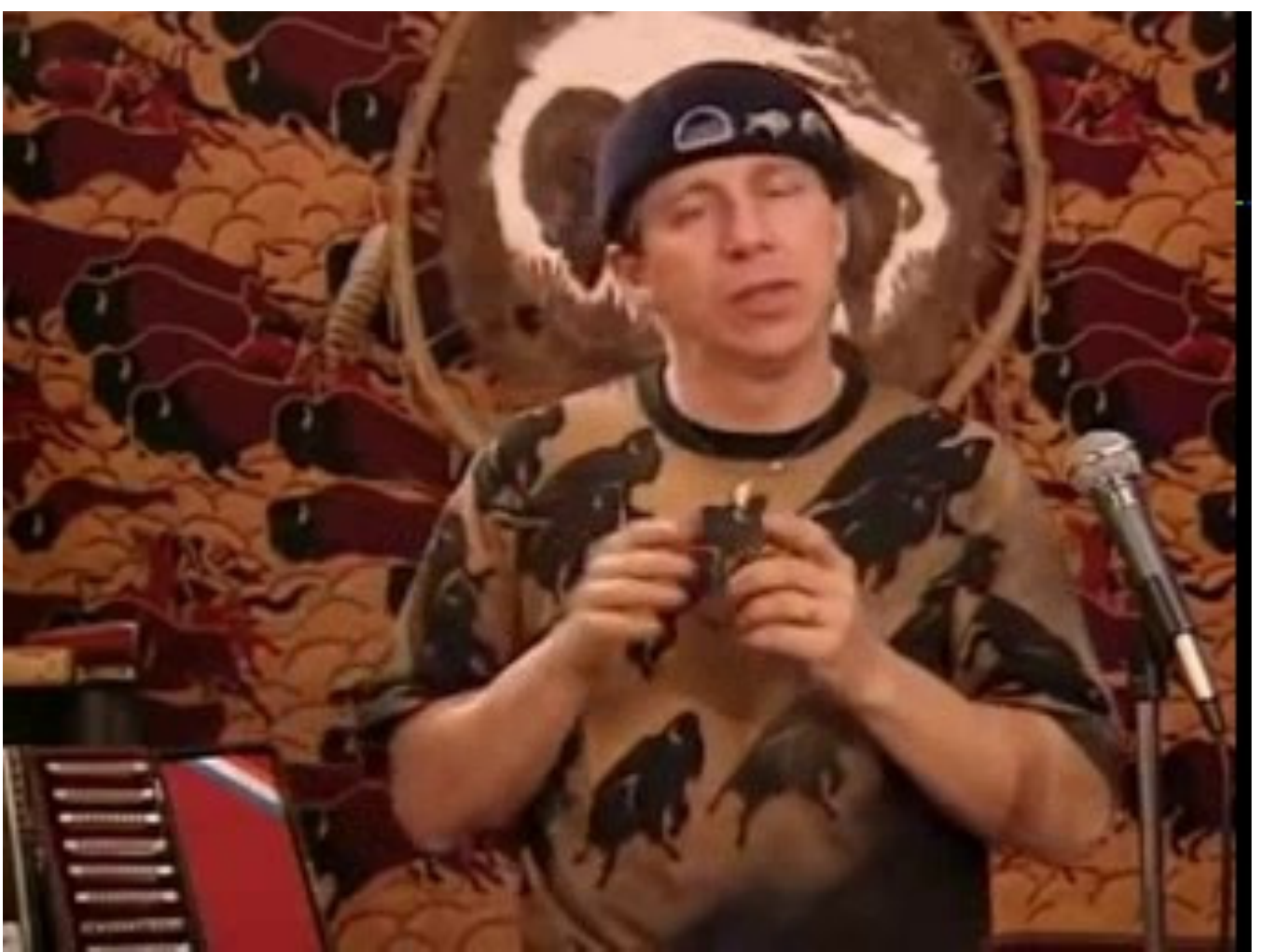}
  \hspace{-1.5mm}
  \includegraphics[width=0.121\linewidth]{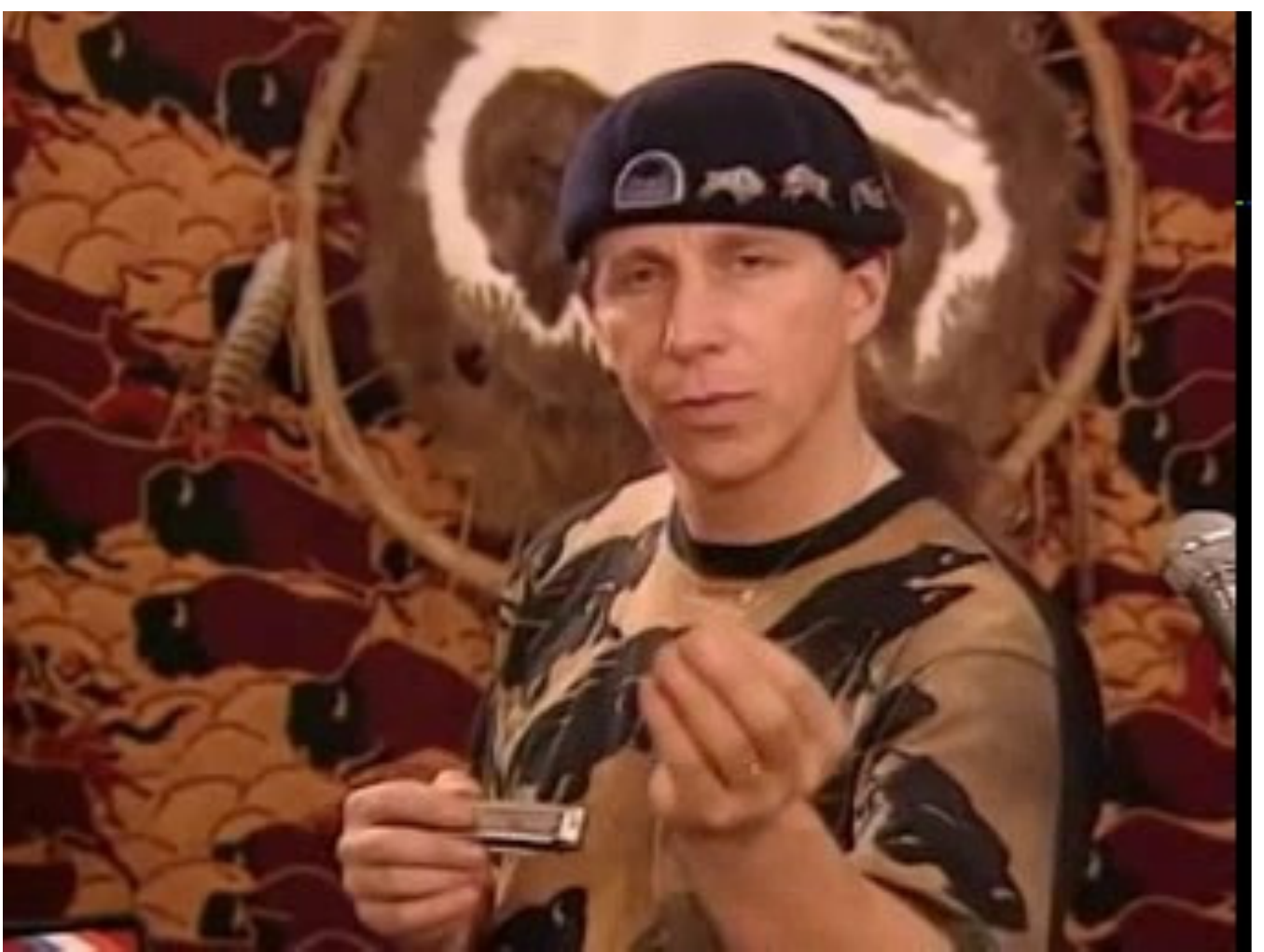}
  \hspace{-1.5mm}
  \includegraphics[width=0.121\linewidth]{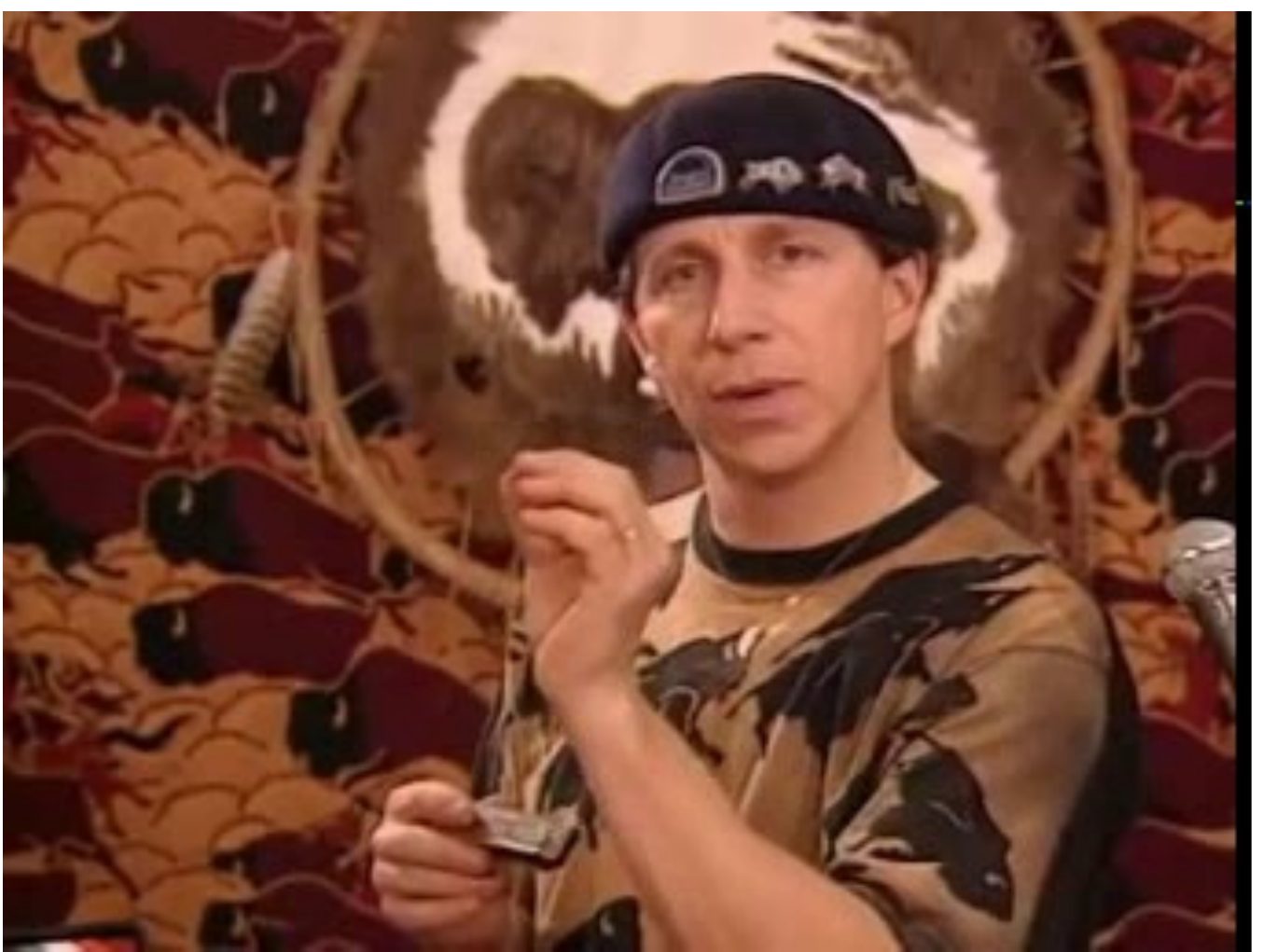}
  \hspace{-1.5mm}
  \includegraphics[width=0.121\linewidth]{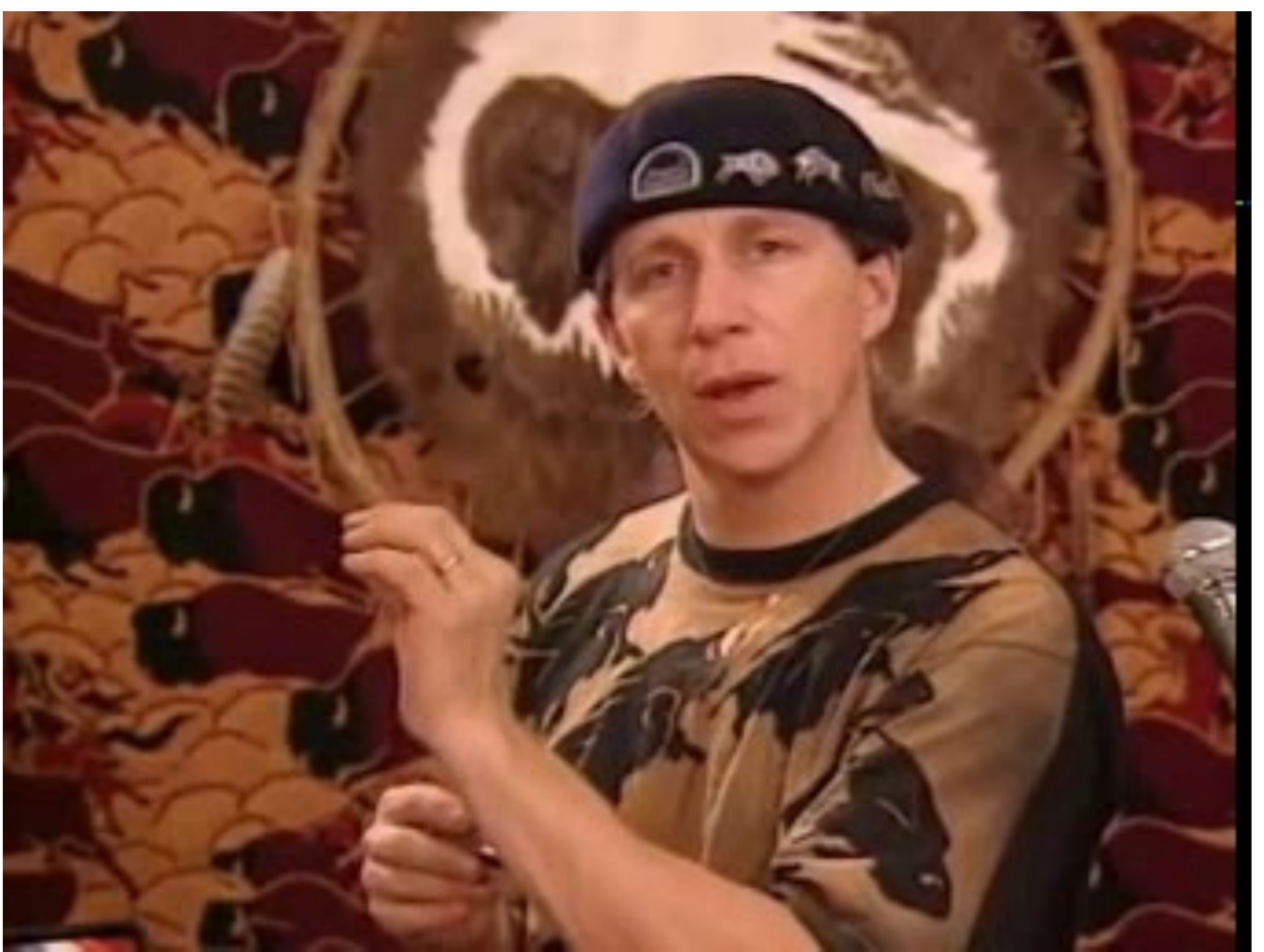}
 \caption{Visualization of attention weights on the test data of THUMOS14 and AcitivtyNet. {\bf The left four frames are with the highest attention weights and the right four frames are with the lowest attention weights.} The above three videos are from THUMOS14 test data with action categories of Rafting, FrontCrawl, BandMarching, and the below three videos are from ActivityNet test data with action classes of TripleJump, ShovelingSnow and PlayingHarmonica.}
 \label{fig:det1}
 \vspace{-15pt}
\end{figure*}

\subsection{Evaluation on WSD}

After evaluation on the problem of weakly supervised action recognition (WSR), we turn to the problem of weakly supervised action detection (WSD) in this subsection. Specifically, we explore the performance of our UntrimmedNet with soft selection module on this problem.

{\bf Qualitative results.} We first visualize the some examples of learned attention weights on the test data of THUMOS14 and ActivityNet. These examples are presented in Figure~\ref{fig:det1}. In this illustration, each row describes one video, where the first 4 images show frames with highest attention weights while the last 4 images are frames with lowest weights. We see that our selection module is able to automatically highlight important frames and to avoid irrelevant frames corresponding to static background or non-action poses.

{\bf Quantitative results.} We also report the performance of action detection on the THUMOS14 dataset, based on the standard intersection over union (IoU) criteria~\cite{THUMOS14}. We simply try a simple detection strategy by thresholding on the attention weights and detection scores as described in Section~\ref{sec:rd}, and aim to illustrate that the learned models with UntrimmedNets could also be applied to action detection. In the future, we may try more advanced detection methods and post-processing techniques. We compare our detection results with other state-of-the-art methods in Table~\ref{tbl:det2}. We notice although our UntrimmedNets simply employ the weak supervision of video-level labels, we can still achieve comparable performance to that of strongly supervised methods, which demonstrates the effectiveness of UntrimmedNets on learning from untrimmed videos.

\section{Conclusions}
\label{sec:con}

In this paper we have presented a novel architecture, called {\em UntrimmedNet}, for weakly supervised action recognition and detection, by directly learning action models from untrimmed videos. As demonstrated on two challenging datasets of untrimmed videos, our UntrimmedNet achieves better or comparable performance for action recognition and detection, when compared with those strongly supervised methods. The superior performance of UntrimmedNet may be ascribed to its advantages of the joint design of classification and selection modules, and optimizing these model parameters in an end-to-end manner.

\section*{Acknowledgement}
This work is partially supported by the ERC Advanced Grant VarCity, the Toyota Research Project TRACE-Zurich, the Big Data Collaboration Research grant from SenseTime Group (CUHK Agreement No. TS1610626), and Early Career Scheme (ECS) grant (No. 24204215).

{\small
\bibliographystyle{ieee}
\bibliography{reference}
}

\end{document}